\documentclass[12pt]{ociamthesis}
\pdfoutput=1

\usepackage{etex}
\usepackage{times}

\usepackage{booktabs}
\usepackage{mathtools}
\usepackage{amsfonts}
\usepackage{amssymb}
\usepackage{stmaryrd}
\SetSymbolFont{stmry}{bold}{U}{stmry}{m}{n}

\usepackage{multirow}
\usepackage{subcaption}
\usepackage{array}
\usepackage[outercaption]{sidecap}
\usepackage{dcolumn}

\usepackage{url}
\usepackage{ccg}
\usepackage{graphicx}
\usepackage[export]{adjustbox}
\usepackage{xspace}
\usepackage{color}

\usepackage{tikz}
\usepackage{pgfplots}
\usepackage{tikz-qtree}
\usetikzlibrary{arrows,positioning,calc,backgrounds,shapes,fit,mindmap,shadows}
\usetikzlibrary{matrix,decorations.pathreplacing}
\usetikzlibrary{fadings}

\usepackage{thesis}

\makeatletter
\newcommand{\@BIBLABEL}{\@emptybiblabel}
\newcommand{\@emptybiblabel}[1]{}
\makeatother
\usepackage{hyperref}

\hypersetup{
    colorlinks=true,
    linkcolor=black,
    citecolor=black,
    filecolor=black,
    urlcolor=black,
}

\title{Distributed Representations for\\[1ex] Compositional Semantics}

\author{Karl Moritz Hermann}
\college{New College}

\degree{Doctor of Philosophy}
\degreedate{Hilary 2014}

\begin{document}

\setcounter{secnumdepth}{3}
\setcounter{tocdepth}{2}

\maketitle

\begin{acknowledgements}

  Many thanks are due at this point. I am deeply grateful to my supervisors,
  Stephen Pulman and Phil Blunsom, for their guidance and advice throughout my
  studies.
  Stephen's encouragement was vital to getting me started again in research
  after my long detour away from computer science.
  Over the course of the past four years my research focus gradually shifted
  towards more statistical and machine learning related approaches. Phil was a
  key driver behind this shift and he has taught me most of what I know about
  these fields today. This thesis is a direct consequence of the many
  discussions I have had with him. It was a great pleasure working with him,
  even if up until this day I walk away from most of our conversations feeling
  enlightened and ignorant at the same time.
  Chris Dyer also deserves thanks: during his stay at Oxford he convinced me to
  look at distributed representations and this is what I ended up doing.

  During my studies I had the opportunity to collaborate with a large number of
  people. I am grateful to Kevin Knight and David Chiang for inviting me to
  spend the summer of 2012 at the ISI/USC.
  In 2013 I spent some time at Google Research in New York, working
  with Kuzman Ganchev as well as with Dipanjan Das and Jason Weston. I have very
  fond memories of that internship, and the work I did at Google ended up
  featuring in this thesis, making for a productive summer.

  I want to thank my colleagues at the CLG group at Oxford and in particular Ed
  Grefenstette who became a good friend and collaborator. Having moved to London
  in 2012 I am very thankful to Sebastian Riedel for hosting me at his MR group
  at UCL, as well as for the stimulating conversations with him and his
  students. Ed and his various flatmates deserve additional thanks for putting
  up with me during my many overnight stays in Oxford in those years.

  On a personal note, thanks for my friends in Oxford and London, and in
  particular the Trinity Arms pub quiz team for helping me maintain a certain
  degree of sanity throughout the whole experience.

  I am grateful to my parents for being a constant source of support and
  encouragement throughout all my (sometimes seemingly random) education and career choices, past and
  present.
  Finally and most of all, I would like to thank my wife
  Cl\'{e}mence for being there.
\end{acknowledgements}

\begin{abstract}

The mathematical representation of semantics is a key issue for Natural Language
Processing (NLP). A lot of research has been devoted to finding ways of representing
the semantics of individual words in vector spaces. \textit{Distributional}
approaches---meaning \textit{distributed} representations that exploit
co-occurrence statistics of large corpora---have proved popular and successful
across a number of tasks. However, natural language usually comes in structures
beyond the word level, with meaning arising not only from the individual words
but also the structure they are contained in at the phrasal or sentential level.
Modelling the compositional process by which the meaning of an utterance arises
from the meaning of its parts is an equally fundamental task of NLP.

This dissertation explores methods for learning distributed semantic
representations and models for composing these into representations for larger
linguistic units. Our underlying hypothesis is that neural models are a
suitable vehicle for learning semantically rich representations and that such
representations in turn are suitable vehicles for solving important tasks in
natural language processing. The contribution of this thesis is a thorough
evaluation of our hypothesis, as part of which we introduce several new
approaches to representation learning and compositional semantics, as well as
multiple state-of-the-art models which apply distributed semantic
representations to various tasks in NLP.

Part \ref{part:dist} focuses on distributed representations and their
application. In particular, in Chapter \ref{chapter:frame-semantic} we explore
the semantic usefulness of distributed representations by evaluating their use
in the task of semantic frame identification.

Part \ref{part:comp} describes the transition from semantic representations for
words to compositional semantics. Chapter \ref{chapter:compositional} covers the
relevant literature in this field. Following this,
Chapter \ref{chapter:syntax}
investigates the role of syntax in semantic composition. For this, we discuss a
series of neural network-based models and learning mechanisms, and demonstrate
how syntactic information can be incorporated into semantic composition. This
study allows us to establish the effectiveness of syntactic information as a
guiding parameter for semantic composition, and answer questions about the link
between syntax and semantics.
Following these discoveries regarding the role of syntax, Chapter
\ref{chapter:multilingual} investigates whether it is possible to further reduce
the impact of monolingual surface forms and syntax when attempting to capture
semantics. Asking how machines can best approximate human
signals of semantics, we propose multilingual information as one method for
grounding semantics, and develop an extension to the distributional hypothesis
for multilingual representations.

Finally, Part \ref{part:concl} summarizes our findings and discusses future work.
\end{abstract}

\begin{romanpages}
\tableofcontents
\listoffigures
\listoftables
\end{romanpages}

\baselineskip=20pt plus1pt
\chapter{Introduction}\label{chapter:intro}

This thesis investigates the application of distributed representations to
semantic models in natural language processing (NLP).  NLP is the discipline
concerned with the interpretation and manipulation of human (natural) language
with computational means. This includes all forms of interaction between
computers and natural language, as well as the development of tools and
resources for working with natural language text.
Tasks within NLP include the annotation of (large-scale) corpora for subsequent
linguistic analysis, algorithms for extracting information from text, models
for translating text across languages, and models for generating text
based on structured data. A lot of recent progress on these problems stems from
the development of statistical approaches to NLP, which deploy machine learning
algorithms that attempt to solve such problems by exploiting patterns found in
large corpora.

Machine learning and statistical NLP have mostly focused on tasks related to
syntax such as part-of-speech (POS) tagging and parsing, as well as larger
tasks such as statistical machine translation, which largely rely on
syntactic and frequency-based effects, too.
More recently, semantics---that is the study of \textit{meaning}---has again
become a focus of research in NLP. While semantics has enjoyed considerable
attention in linguistics and Computational Linguistics, this was primarily from
the perspective of symbolic-reasoning, with the exception of early NLP pioneers
such as Karen Sp\"{a}rck Jones and Margaret Masterman \cite[\textit{inter
    alia}]{SparckJones:1988,Masterman:2005}. This thesis is part of
this line of work, which investigates semantics within the realm of statistical
NLP and machine learning. Precisely, we focus on the study of representing
meaning with continuous, distributed objects and explore how to learn and
manipulate these objects in such a fashion that the information contained
therein can be exploited for various NLP-related tasks.

\section{Aims of this thesis}\label{sec:intro:thesis}

The primary aim of this thesis is to investigate the use of distributed
representations for capturing semantics, and to evaluate their efficacy in
solving tasks in NLP for which a degree of semantic understanding would be
beneficial. Our hypothesis is that distributed representations are a highly
suitable mechanism for capturing and manipulating semantics, and further, that
meaning both at the word level and beyond can be encoded distributionally.

Throughout this thesis we evaluate this hypothesis in a number of ways. In order
to establish the suitability and efficacy of distributed representations for
capturing semantics we apply such representations to a number of popular and
important tasks in NLP. We evaluate the performance of models supported by
distributed semantic representations relative to the performance of alternative,
state-of-the-art solutions. As we end up outperforming the prior state of the
art on a number of such problems, using relatively simple models in conjunction
with distributed representations, these experiments strongly support the first
part of this thesis' hypothesis.
The second aspect of our hypothesis concerns the question whether distributed
representations can be used to represent semantics beyond the word level. This
question is investigated throughout Part \ref{part:comp} of this thesis, which
focuses on distributed representations for compositional semantics.
We attempt to verify this hypothesis two-fold. First, we again develop systems
for semantic vector composition that learn to represent a sentence or a document
in a distributed fashion, and then pit these representations against other
approaches on several tasks. Second, we analyse a number of popular methods for
learning and composing distributed representations and evaluate to what extend
these methods are capable of learning to encode actual semantics.

In the following section we discuss the main contributions of this thesis.
Subsequently, \S\ref{sec:intro:struc} explains the structure of the remainder of
this thesis and summarises the content of each chapter.

\section{Contributions}

Here, we summarise the major contributions of this thesis.

The task considered in Chapter \ref{chapter:frame-semantic}---frame-semantic
parsing---is a very popular and important task within NLP. The chapter
contributes to the thesis two-fold. First, by using distributed semantic
representations to solve the task, we determine the feasibility of using
distributed representations for capturing semantics and furthermore discover a
number of important factors to be considered when using distributed
representations. Second, we present a full frame-semantic parsing pipeline as
part of our experiments, and contribute to the field by setting a new state of
the art on this task. Thus, we have not only validated our thesis about the use
of distributed semantic representations, but further have demonstrated the
superior performance of this approach over all prior work on the semantic
frame-identification and frame-parsing tasks.

Following a background chapter on compositional semantics (Chapter
  \ref{chapter:compositional}), this thesis continues by exploring the effect of
syntax in guiding semantic composition. Here (Chapter \ref{chapter:syntax}), we
present a novel method for composing and learning distributed semantic
representations given syntactic information. We focus on combinatory categorial
grammars in this chapter, and show that our system, which integrates syntactic
information with semantics, outperforms comparable work that does not exploit
syntactic information. We fulfil the aim of this chapter by thus establishing a
link between syntax and semantics. As an additional contribution, we provide a
novel model for sentiment analysis, which outperformed the state-of-the-art
system at that point in time.

Most prior work on semantic representation learning focuses on task-specific
learning, which inevitably will lead to representations exhibiting only certain
aspects of semantics as required by the objective function of a given task. In
Chapter \ref{chapter:multilingual} we explore the use of multilingual data to
learn representations further abstracted away from monolingual surface forms and
task-specific biases, thereby extending the distributional hypothesis for
multilingual joint-space representations. We demonstrate how multilingual data
can be used to learn semantic distributed representations and develop a novel
algorithm for doing so efficiently. We apply representations learned under this
framework to a document classification task to verify their efficacy. The
success of our models in the document classification experiments give further
support to the initial hypothesis of this thesis concerning the usefulness of
distributed semantic representations. As the models learn through semantic
transfer at the sentence level or beyond, we further gain additional insight
into the second part of our hypothesis, namely that such distributed
representation are capable of encoding semantics beyond the word level.

\section{Thesis Structure}\label{sec:intro:struc}

This thesis is organised into two distinct parts. Part \ref{part:dist} focuses
on distributed representations and their application, with Chapter
\ref{chapter:distrib} introducing distributed representations and their
application to natural language semantics. Chapter \ref{chapter:frame-semantic}
explores the efficacy of semantic distributed representations by evaluating
their use in the task of semantic frame identification.

Part \ref{part:comp} describes the transition from semantic representations for
words to compositional semantics. Chapter \ref{chapter:compositional} covers the
relevant literature in this field. Following this, we continue our investigation
of semantics in distributed representations. Chapter \ref{chapter:syntax}
investigates the role of syntax in semantic composition. For this, we discuss a
series of neural network-based models and learning mechanisms, and demonstrate
how syntactic information can be incorporated into semantic composition. This
study allows us to establish the effectiveness of syntactic information as a
guiding parameter for semantic composition, and answer questions about the link
between syntax and semantics.
Following the discoveries made regarding the role of syntax, Chapter
\ref{chapter:multilingual} investigates whether is it possible to further reduce
the impact of monolingual surface forms and syntax when attempting to capture
semantics. Asking the question of how machines can best approximate human
signals of semantics, we propose multilingual information as one proxy for
machines' lack of shared embodiment and bodily experience and describe
mechanisms for extracting semantic representations from parallel corpora.
We conclude with Part \ref{part:concl}, which summarises our findings and
discusses future work.

This thesis contains material that has been previously
published. The bulk of the work in this thesis is based on three papers
presented at the Annual Meeting of the Association for Computational Linguistics
(ACL), with smaller aspects of the thesis being based on further publications as
follows.
The work contained in these publications and presented in this thesis is
principally mine, except when stated otherwise in the relevant chapters. Where
co-authors have contributed significantly or where aspects of the work published
are solely the responsibility of a co-author, this is accredited accordingly in
the respective chapters.

Below we summarise each chapter of this thesis and elaborate on the material
contained therein.

\begin{description}
  \item[Chapter \ref{chapter:distrib}: Distributed Semantic Representations]
    \hfill \\
    We discuss semantics in the context of NLP, and provide an overview of
    popular attempts to capture, express, and reason with semantics in the
    literature. As part of this we motivate distributed semantic
    representations. We then go on to discuss how such representations can be
    learned and cover a number of underlying principles necessary for
    understanding the remainder of this thesis.

  \item[Chapter \ref{chapter:frame-semantic}: Frame Semantic Parsing with
      Distributed Representations] \hfill \\
    Having motivated the use of distributed representations for semantics, we
    underline this argument with an extensive empirical evaluation. We focus on
    the task of semantic frame identification, for which we propose a new
    solution relying on distributed semantic representations. We describe our
    novel approach as well as relevant work in the literature, and subsequently
    evaluate our new model. For a full comparison we also make use of a
    semantic role-labelling system, which allows us to test the semantic frame
    identification model within a full frame-semantic parsing pipeline, where
    our model sets a new state of the art. The work presented in this chapter is
    based on the following publication:
    \begin{list}{bibrandom1}{}
    \bibitem[x1]{x1}
      Karl~Moritz Hermann, Dipanjan Das, Jason Weston and Kuzman Ganchev.
      \newblock 2014.
      \newblock {Semantic Frame Identification with Distributed Word Representations}.
      \newblock In {\em Proceedings of ACL}.
    \end{list}
    \nocite{Hermann:2014:ACLgoogle}

  \item[Chapter \ref{chapter:compositional}: Compositional Distributed
      Representations] \hfill \\
    Having established the efficacy of distributed representations in conveying
    semantic information in the previous chapter, we now go on to focus on
    compositional semantics, that is the representation of meaning of larger,
    composed linguistic units such as phrases or sentences. Here, we survey
    prior work in that field, as well as the theoretical foundations on which
    this thesis builds. Further, we attempt to structure prior efforts on tasks
    in this area by discriminating between lexical-function and algebraic, as
    well as between distributional and distributed approaches to compositional
    semantics.

  \item[Chapter \ref{chapter:syntax}: The Role of Syntax in Compositional
      Semantics] \hfill \\
    Having already made extensive use of syntactic information in Chapter
    \ref{chapter:frame-semantic}, we now investigate the role of syntax in
    compositional (distributional) semantics in more detail. We do this by
    extending existing work on compositional semantics with various types of
    syntactic information based on combinatory categorial grammar, and evaluate
    the effects derived from this additional information. This chapter is based
    on work first published in:
    \begin{list}{bibrandom2}{}
    \bibitem[x2]{x2}
      Karl~Moritz Hermann and Phil Blunsom.
      \newblock 2013.
      \newblock {The Role of Syntax in Vector Space Models of Compositional Semantics}.
      \newblock In {\em Proceedings of ACL}.
    \end{list}
    \nocite{Hermann:2013:ACL}

  \item[Chapter \ref{chapter:multilingual}: Multilingual Approaches for Learning
      Semantics] \hfill \\
    Having so far in this thesis focused on task-specific problems which we
    enhanced with semantic information, we now attempt to learn more general
    semantic representations by reducing task-specific bias when learning
    representations, and further, by abstracting away from monolingual surface
    forms through the use of multilingual data. We develop a novel objective
    function for word representation learning that can be applied to
    multilingual data and that---as a further novelty---does not rely on word
    alignment across languages. Multiple evaluations validate our approach, with
    our model setting a new state of the art on a crosslingual document
    classification task. The work presented in this chapter is based on the
    following two publications:
    \begin{list}{bibrandom3}{}
    \bibitem[x3]{x3}
      Karl~Moritz Hermann and Phil Blunsom.
      \newblock 2014.
      \newblock {Multilingual Distributed Representations without Word Alignment}.
      \newblock In {\em Proceedings of ICLR}.
    \bibitem[x4]{x4}
      Karl~Moritz Hermann and Phil Blunsom.
      \newblock 2014.
      \newblock {Multilingual Models for Compositional Distributed Semantics}.
      \newblock In {\em Proceedings of ACL}.
    \end{list}
    \nocite{Hermann:2014:ACLphil,Hermann:2014:ICLR}

  \item[Chapter \ref{chapter:conclusions}: Conclusions] \hfill \\
    The final chapter of this thesis summarises our findings and proposes
    future work based on the work presented here.

\end{description}

\part{Distributed Semantics}\label{part:dist}
\chapter{Distributed Semantic Representations}\label{chapter:distrib}

\begin{chapterabstract}
  This chapter presents an overview of key concepts, formalisms and background
  literature related to distributed semantics. This review begins by introducing the
  distributional account of semantics according to \newcite{Firth:1957} and various
  dimensionality reducing techniques typically combined with extracting
  distributional representations. Subsequently, we will explore alternative
  methods for learning \textit{distributed} representations for words and their
  applications.
\end{chapterabstract}

\section{Introduction}
\label{sec:introduction}

In this chapter we provide an overview of popular methods for learning
distributed word representations. We begin in \S\ref{sec:ts:semantics} by
discussing the role of semantics in natural language processing.
\S\ref{sec:ts:distributional} describes the distributional account of semantics
and how it can be exploited for learning distributed representations in an
unsupervised setting. Subsequently, \S\ref{sec:ts:neurallm} covers alternative
methods for learning distributed representations, going beyond a purely
distributional approach. Finally, we survey prior work on, and applications of,
distributed representations for words in \S\ref{sec:ts:applications}.
In the literature distributed word representations are frequently referred to as
\textit{word embeddings}; please note that we will use these two terms
interchangeably throughout this thesis.

Words can be represented as discrete units by mapping a string of characters
to integers by looking up words in a dictionary. Frequently, however, it is
better to represent words by going beyond their surface form and attempting to
capture syntactic and semantic aspects in their representation. This would be
useful for establishing similarities and relationships among different words.
Within language modelling for instance, part of speech (POS) tags have proved a
useful method for clustering words and determining likely word sequences in a
given language.  Related ideas include augmenting word representations with
grammatical information such as their conjugated or declined form, their
infinitive or stem and other morpho-syntactic information. Such grammatical
information can be used to learn relationships between morphemes of the same
base word.

While syntactic information can be useful for a number of tasks such as language
modelling or word reordering in generative models, these problems, as well as a
large number of others, would also benefit from semantic information included in
a word's representation. In the case of language modelling it is easy to see how
a measure for semantic similarity between words would allow such a model to
better generalise for rare words, as the semantic similarity score could be used
to make predictions based on the statistics of semantically similar, more
frequent terms.

\section{Semantics}\label{sec:ts:semantics}

While there is little doubt concerning the usefulness of semantic information,
the question of how such knowledge can be ``acquired, organized and ultimately
used in language processing and understanding has been a topic for great debate
in cognitive science'' \cite{Mitchell:2010}.
Semantics have been represented in a number of ways throughout the literature.
Broadly, such accounts of semantics can be categorised into feature-based models
and semantics spaces. The related concept of semantic networks also deserves a
mention, and will also briefly be discussed together with the other two accounts
below.

\subsection{Feature-Based Representations}\label{sec:ts:sem:feature}

Feature-based models attempt to capture specific aspects of semantics, either
through a list of pre-defined features or by learning attributes that are
considered relevant to the meaning of a word by human annotators
\cite[\textit{inter alia}]{Andrews:2009,McRae:1997}.

Thesauri and other lexicographical resources such as the \wordnet project
\cite{Fellbaum:1998} can provide some such semantic features by providing
relational information for words such as hypernomy and hyponymy, meronymy or
synonymy and antonymy.

Related lines of work include super-tagging \cite{Bangalore:1999} and
subsequently supersense-tagging \cite{Ciaramita:2003,Curran:2005}. Super-tagging
provides richer syntactic information about words by capturing the localised
syntactic context in which they appear. The similarly named supersense-tagging,
on the other hand, attempts to learn ``supersenses'' as used by the \wordnet
lexicographers for words outside of the \wordnet lexicon.

All of these approaches, however, are limiting in that they can only capture
specific aspects of syntactic or semantic information, and further, in that they
typically rely on syntactic and semantic categories as defined by hand.
Unsupervised clustering methods can partially overcome the second issue, but
the first remains.

\subsection{Semantic Networks}\label{sec:ts:sem:network}

Semantic networks describe semantic relations between entities or concepts.
Conventionally such networks are represented as directed or undirected graphs,
with nodes representing concepts and vertices (edges) between nodes representing
relations. The idea was first proposed by \newcite{Peirce:1931}, with the
application to semantics proposed in \newcite{Richens:1956} and
\newcite{Richens:1958} and developed by \newcite{Collins:1969}.

\wordnet, introduced in \S\ref{sec:ts:sem:feature} is an example for such a
semantic network, where words---concepts---are linked by relations such as
synonymy or meronymy. Alternative networks use more explicit relations, such as
\textsc{IS-A} and \textsc{SIBLING-OF} relations. Semantic similarity tends to be
measured by the path length between two concepts.

Semantic networks are popular for certain tasks. For instance, in joint work
prior to this thesis, we studied the use of such semantic networks as a form of
interlingua for machine translation \cite{ISI:2012,ISI:2013}. Similarly, they
are popular for tasks in relation extraction and identification
\egcite{Riedel:2013}.

However, as semantic networks are typically hand crafted with a predetermined
set of features, their application is limited to domains with the necessary
resources or availability of annotators. Further, path length as a similarity
measure is vague and cannot be applied globally: For instance, for
\textit{cat}, one could envisage relations ``\textit{cat} IS-A \textit{mammal}''
and ``\textit{cat} HAS \textit{whiskers}'', which would insinuate an equal
degree of similarity between these terms.

\subsection{Semantic Space Representations}\label{sec:ts:sem:space}

An alternative approach for representing words, which we explore in this
thesis, are distributed representations or semantic space representations. Here,
words are represented by mathematical objects, frequently vectors.

Conventional dictionary-based methods for representing words as
indices can be used to represent words as vectors. In that case, word vectors
would have the size of the dictionary and each word would be captured by a
vector containing zeros in all positions except for a one in the position of
their index. This is known as a \textit{one-hot} representation.
Obvious shortcomings of one-hot representations include their high
dimensionality, their inability to deal with out of vocabulary (OOV) words, and
furthermore their lack of robustness with regard to sparsity, as no information
is shared across words.

\begin{figure}[t]
  \centering
\begin{tikzpicture}
  \begin{axis}[ xlabel=Speed, ylabel=Taste, scale only axis, xmin=0,xmax=9,
      ymin=0,ymax=9, width=0.4\textwidth, ]
    \node[inner sep=0.5pt,pin=88:Pig] at (axis cs:7,5) {};
    \addplot[color=black,mark=*] coordinates { (0, 0) (7, 5) };
    \node[inner sep=0.5pt,pin=178:Cow] at (axis cs:5,7) {};
    \addplot[color=black,mark=*] coordinates { (0, 0) (5, 7) };
    \node[inner sep=0.5pt,pin=88:Car] at (axis cs:8,0.5) {};
    \addplot[color=black,mark=*] coordinates { (0, 0) (8, 0.5) };
    \addplot[color=blue,style=dashed] coordinates { (7, 5) (5, 7) };
    \addplot[color=red,style=dashed] coordinates { (7, 5) (8, 0.5) };
    \addplot[color=red,style=dashed] coordinates { (5, 7) (8, 0.5) };
    \draw (axis cs:1.5,1.07)arc[radius=0.48cm,start angle=15,end angle=63];
    \node at (axis cs:1.1,1.15) {$\theta$};
  \end{axis}
\end{tikzpicture}
\caption[A hypothetical distributed semantic space]{A hypothetical distributed
  semantic space for \textit{cow}, \textit{pig} and \textit{car}. $\theta$
  denotes the cosine angle between \textit{cow} and \textit{pig}, the dashed
  lines the Euclidean distances between the three words.}\label{fig:ts:cows}
\end{figure}
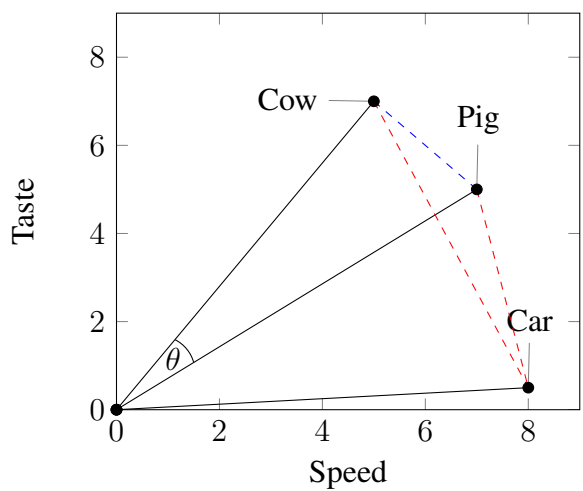

Better results can be achieved by representing words as continuous vectors,
where each dimension represents some latent category (e.g. a semantic or
syntactic feature). See Figure \ref{fig:ts:cows} for an example. Key benefits of
such a representation are that it does not require hand crafted features, and
that distance measures can be applied to evaluate semantic proximity between
words given their distributed representation.

Such distributed representations stem from the idea that the meaning of a word
can be captured from its linguistic environment. While not all models of
distributed semantics make explicit use of this distributional hypothesis, it
directly or indirectly forms the basis of most work in this field. Later on in
this thesis, when introducing multilingual models in Chapter
\ref{chapter:multilingual}, we generalise this concept by using a different form
of context informing the semantic learning process.
In the next
section \S\ref{sec:ts:distributional} we introduce the distributional hypothesis
in greater detail before surveying other popular methods for learning continuous
distributed representations of word-level semantics from \S\ref{sec:ts:neurallm}
onwards.

\section{Distributional Representations}\label{sec:ts:distributional}

Distributional representations encode an expression by its environment, assuming
the context-dependent nature of meaning according to which one ``shall know a
word by the company it keeps'' \cite{Firth:1957}.  The underlying idea of this
distributional account of semantics---concisely captured by the quote above---is
that the meaning of words can be inferred from their usage and the context they
appear in. By implication, this also means that words with similar distributions
over the contexts they appear in have similar meaning. For instance, we assume
that the words \textit{bicycle} and \textit{bike} would occur in similar
contexts, whereas the contexts of \textit{bicycle} and \textit{oranges} would be
rather different.

This distributional hypothesis provides the basis for statistical semantics,
allowing the inference of semantics from distributional information extracted
from sufficiently large corpora. Distributional models of semantics thus
characterize the meanings of words as a function of the words they co-occur
with.

Effectively this is usually achieved by considering the co-occurrence with other
words in large corpora and mapping this co-occurrence information onto a matrix.
Thus, distributional representations are a special form of distributed
representations, where the distributed information equals distributional
information.
Distributional representations can be learned through a number of approaches and
are not limited to using words as the basis of their co-occurrence matrix.
Examples for other bases include larger linguistic units such as $n$-grams
\cite{Jones:2007}, documents \cite{Landauer:1997} or predicate-argument slots
\cite{Grefenstette:1994,Pado:2007}.
In their simplest form, statistical information from large corpora can be used
to learn distributed word representations.
Usually, the information used to compute word
embeddings are words occurring very close to the target word, typically in a
five
word window.  This is related to topic-modelling techniques such as LSA
\cite{Dumais:1988}, LSI, and LDA \cite{Blei:2003} (see \S\ref{sec:ts:dr:lsi}),
but these methods use a document-level context, and tend to capture the topics a
word is used in rather than its more immediate syntactic context.

As already stated in \S\ref{sec:ts:sem:space}, an advantage of this statistical
approach to semantics is that word meaning can now be quantified. The semantic
similarity between two words can be measured by the distance between their
representation in such a space (or the cosine of the angle between them). See
again Figure \ref{fig:ts:cows} for an illustration of this.

These models, mathematically instantiated as sets of vectors in high dimensional
vector spaces, have been applied to tasks such as thesaurus extraction
\cite{Grefenstette:1994,Curran:2004}, word-sense discrimination
\cite{Schutze:1998}, automated essay marking \cite{Landauer:1997}, word-word
similarity \cite{Mitchell:2008} and so on. We provide an overview of such
applications and methods in \S\ref{sec:ts:applications}.

We describe the collocational approach for learning distributional
representations in \S\ref{sec:ts:dr:learning}, followed by expanding on a
number of strategies for improving these representations using dimensionality
reduction and smoothing techniques.

\subsection{Learning distributional representations}\label{sec:ts:dr:learning}

The vectors for distributional semantic models are generally produced from
corpus data via the following procedure:
\begin{enumerate}
  \item For each word in a lexicon, its contexts of appearance are collected
    from a corpus, based on some context-selection criterion (e.g.~tokens within
    $k$ words of the target word, or being linked by a dependency or other
    syntactic relation).
  \item These contexts are processed to reshape or filter the information they
    contain (e.g.~only considering context words from the most frequent $n$
    words in a corpus, or those of specific syntactic classes).
  \item These contexts of occurrence are encoded in a vector where each vector
    component corresponds to a possible context, and every component weight
    corresponds to how frequently the target word occurs in that context.
  \item Optionally, the vector component weights are reweighted by some function
    (e.g.~term-frequency-inverse-document-frequency, ratio of probabilities,
      pointwise mutual information).
  \item Optionally, the vectors are subsequently projected onto a lower
    dimensional space using some dimensionality reduction technique (see
    \S\ref{sec:ts:methods:redux}).
\end{enumerate}

The semantic similarity of words is then determined by computing the distance
between their thus-constructed vector representations, using geometric
similarity metrics such as the cosine of the angle between the vectors (see
\S\ref{sec:ts:methods:simil}).

While typically only the co-occurrence with some $n$ most frequent words is
considered, this can still lead to fairly high-dimensional vector
representations. As this dimensionality will influence the size of the many
other model parameters, it can be useful to reduce word representations learned
through distributional means. Such dimensionality reduction further allows words
to have similar representations regardless of the particular documents from
which their co-occurrence statistics have been extracted. In
\S\ref{sec:ts:methods:redux} we provide an overview of commonly used methods for
this purpose.

For a comprehensive overview of the different parametric options typically used
in the production of such semantic vectors and of their comparison, we refer to
the surveys found in \newcite{Curran:2004} and \newcite{Mitchell:2011}.

\subsection{Weighting Techniques}\label{sec:ts:dr:norm}

One frequently employed mechanism for improving the quality of the extracted
distributional vectors is to apply some form of normalisation. The purpose of
normalising vectors is to maximise their information content and/or to
pre-process such vectors for subsequent use in composition models or other
functions where inputs are expected to be bound by a certain range or matching a
certain distribution.
Similarly, vectors are frequently normalized to form a probability distribution.

\paragraph{TF-IDF}
The term frequency-inverse document frequency (TF-IDF) is a statistical measure
frequently employed for this purpose \cite{SparckJones:1988}. TF-IDF stems from
information retrieval and text mining, and is used to weight words by their
semantic content. In its simplest form, TF-IDF provides a weight for each word,
computed by considering two statistics.

First, the term frequency is a measure of how frequently a word appears in a given
document. Second, the inverse document frequency is the total number of
documents in a corpus, divided by the number of documents containing the word in
question at least once. Thus, common words may obtain a high term frequency but
a low inverse document frequency. On the other hand, words appearing in a
particular document but rarely throughout the overall corpus may offset their
low term frequency by a high \textit{idf}-score.

For distributional representations, TF-IDF can be used in several ways. It
provides a simple heuristic for identifying and removing \textit{stop}-words.
Similarly, TF-IDF weights can be used to scale distributional counts. It was
shown that this technique can alleviate sparsity-induced problems in distributed
representation learning. Conventionally, this is achieved by treating the
context words of each word type as a document from which TF-IDF weights can be
learned. Thereby, greater weight is given ``to words with more idiosyncratic
distributions and may improve the informativeness of a distributional
representation'' \cite{Huang:2009}.

\section{Neural Language Models}\label{sec:ts:neurallm}

Neural language models are another popular approach for inducing distributed
word representations, first developed by Y. Bengio and coauthors \cite{Bengio:2003}.

They have subsequently been explored by others
\cite{Collobert:2008,Mnih:2009,Mikolov:2010} and have achieved good performance
across various tasks.
The neural language model described by \newcite{Mikolov:2010} for instance
learns word embeddings together with additional transformation matrices which
are used to predict the next word given a context vector created by the previous
words. \newcite{Collobert:2011} further popularized neural network architectures
for learning word embeddings from large amounts of largely unlabeled data by
showing the embeddings can then be used to improve standard supervised tasks,
i.e. in a semi-supervised setup.  Unsupervised word representations can easily
be plugged into a variety of NLP tasks.

\section{Methods}\label{sec:ts:methods}

\subsection{LSA, LSI and LDA}\label{sec:ts:dr:lsi}

Latent Semantic Analysis (LSA, henceforth) \cite{Dumais:1988} describes a
mechanism for extracting latent semantic information from words in context.
In the context of information retrieval, LSA is also known as Latent Semantic
Indexing (LSI).

LSA uses a term-document matrix which describes the occurrences of terms in
documents. For this matrix $X$ a lower rank approximation is found, using the
$k$ largest singular values from $X = U\Sigma V^T$ where $U$ and $V$ are
orthogonal matrices and $\Sigma$ a diagonal matrix containing the singular
values in question. Using the decomposition $U\Sigma V$ allows one to find the
best $k$ rank approximation for $X$.

While LSA is typically used in connection with bag of word models focussed on
learning topic representations for documents, it can also be applied to
distributional representation learning. Similar to TF-IDF
(\S\ref{sec:ts:dr:norm} above), LSA can be applied to context vectors of a given
word \cite[\textit{inter alia}]{Huang:2009}.

\begin{figure}[t]\centering
\includegraphics[scale=0.7]{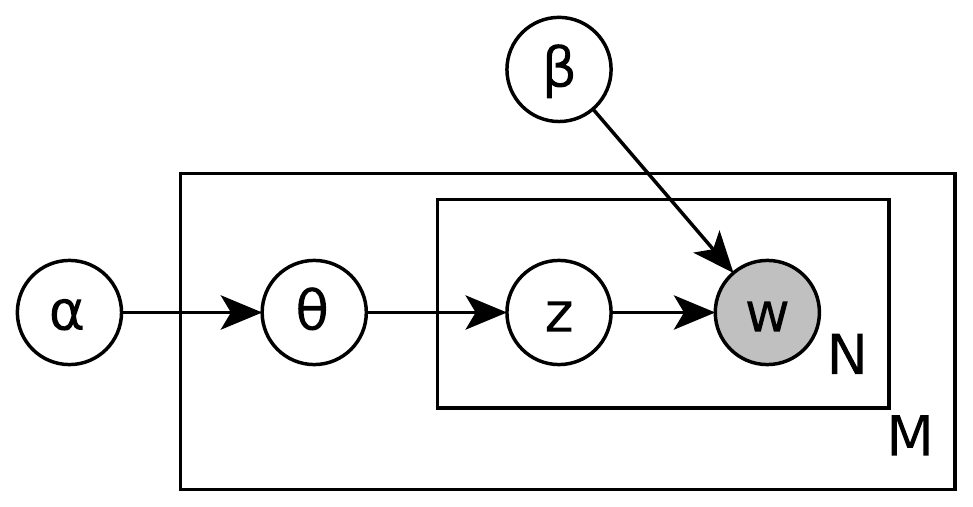}
\hspace{3em}
\includegraphics[scale=0.7]{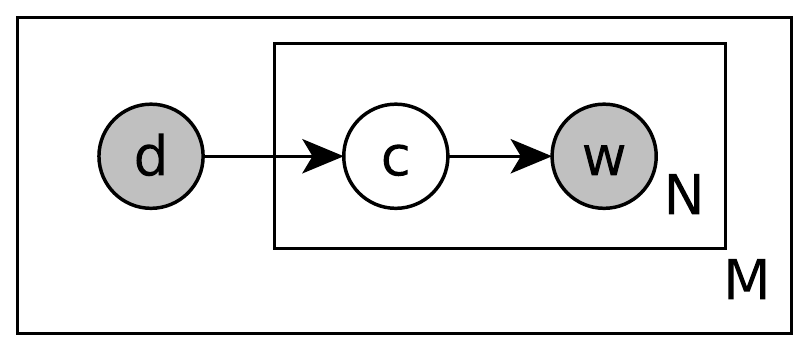}
\caption[Comparison of LDA and pLSA models]{Comparison of LDA (left) and pLSA
  (right) models as plate diagrams. In both diagrams, $M$ depicts the number of
  documents and $N$ the number of words within a document. The LDA model uses
  two Dirichlet priors, parametrized with $\alpha$ and $\beta$ for its
  document-topic and topic-word distributions.  $\theta$ denotes the topic
  distribution for a given document, and $z$ the drawn topic for a particular
  word $w$. In the case of pLSA, $d$ is the document index which informs the
  topic choice $c$ per word $w$.
}\label{fig:ts:ldaplsa}
\end{figure}

\paragraph{Latent Dirichlet Allocation} (LDA, henceforth) also deserves mention
here. Similar to LSA, LDA was initially developed with a focus on document-level
analysis and topic modelling in particular \cite{Blei:2003}.
LDA uses a generative latent variable model that models documents as mixtures
over topics with each of these latent topics using a probability
distribution over a vocabulary to generate words. This is comparable to a probabilistic
variant of LSA (pLSA), with the topics of LDA being equivalent to the latent
class structure of pLSA. See Figure \ref{fig:ts:ldaplsa} for a comparison of the
two models.

A crucial difference between topic models such as LDA and other models presented
in this chapter is that LDA represents words as a probability distribution
rather than as points in a high-dimensional (semantic) space. This probability
distribution, however, is equivalent to points on a simplex in a
high-dimensional space. As generative
models, they learn probabilities for words given a topic. Thus, both LDA and
pLSA can easily be used to learn low dimensional representations of observed
variables by expressing their distribution over the latent topic variables as a
probabilistic distributed representation. In prior work, not included in this
thesis, we applied variations of LDA to learn semantic representations for noun
compounds and adjective noun pairs
\cite{Hermann:2012:Ranking,Hermann:2012:SelPref}.

\subsection{Dimensionality Reduction Techniques}\label{sec:ts:methods:redux}

Beyond specific methods such as LSA there are a number of general, statistical
methods for reducing the rank of vectors, which can easily be applied to both
distributional and distributed word representations.
As some of the methods introduced in this chapter, particularly those extracting
distributional representations, can lead to very large vectors, these methods
are very useful both to alleviate sparsity via smoothing as well as to improve
the efficiency of subsequent models making use of such representations.

\paragraph{Principal Component Analysis} (PCA, henceforth) is similar to LSA
(above) in that it performs rank reduction on a matrix using decomposition and
orthogonal vectors to take correlation between individual vector elements
(matrix rows) into account. The key difference is that instead of using a
term-document matrix, PCA uses a term covariance matrix, processed to have a
zero mean.
Both LSA and PCA rely on singular value decomposition (SVD) for the actual rank
reduction operation.

\paragraph{Factor Analysis} is another statistical method for discovering latent
variables (\textit{factors}) that can represent higher-dimensional data through
a lower-rank approximation. Here, variables are first shifted to have zero mean,
and subsequently a factor matrix $L$ is learned, such that $X - \mu = LF +
\epsilon$, where $F$ denotes the low rank approximation of $X - \mu$ and
$\epsilon$ denotes some error term.
On the surface, this approach is comparable to PCA. The main difference between
the two methods is that PCA is a descriptive technique, while factor analysis uses
a latent modelling technique to learn its factors.

\subsection{Similarity Metrics}\label{sec:ts:methods:simil}

For many tasks it is necessary to evaluate the similarity between several
distributed representations. Examples for this include word-word similarity
tasks (similarity between two representations), unsupervised clustering (cluster
a number of entities with distributed representations) or annotation tasks (find
the closest label to a representation in a given space).

Depending on the model settings and normalization options, \textit{cosine
  similarities} (Eq. \ref{eqn:ts:cos}) or \textit{Euclidean distances} (Eq.
  \ref{eqn:ts:euc}) can be used to evaluate such tasks.
\begin{align}
  Cos(\vec{x},\vec{y}) &= cos(\theta) = \frac{\vec{x} \cdot \vec{y}}{\|\vec{x}\| \|\vec{y}\|} \label{eqn:ts:cos}\\
  Eucl(\vec{x},\vec{y}) &= |\vec{x} - \vec{y}| = \sqrt{\sum_{i=0}^{d} (x_i - y_i)^2} \label{eqn:ts:euc}
\end{align}
The key difference between the two measures is that the cosine distance
(or similarity) takes into account the difference between two vectors in terms
of their angle, while the Euclidean distance accounts for the metric distance
between two points. Space is treated as an inner product space for the
cosine distance, and the cosine distance can be derived from the \textit{inner
  product} (or \textit{dot product}):
\begin{equation}
  \vec{x}\cdot \vec{y} = \|\vec{x}\| \|\vec{y}\| cos(\theta),
\end{equation}
which can also be used as a similarity metric in its own right.

Another metric that is interesting to consider for distributed representations
is the Mahalanobis distance. The Mahalanobis distance can be viewed as a
scale-invariant extension of the Euclidean distance that further accounts for
correlations within a dataset. In its general formulation (Eq.
\ref{eqn:ts:maha}), it uses a covariance matrix $S$ to effectively scale and
account for interactions among the different elements within the vectors under
comparison.
\begin{equation}
d(\vec{x},\vec{y}) = \sqrt{(\vec{x}-\vec{y})^TS^{-1}(\vec{x}-\vec{y})}
\label{eqn:ts:maha}
\end{equation}

Note that this describes the quadratic form of a Gaussian distribution, as
used in the exponentiated part of a Gaussian mixture model (Equations
  \ref{eqn:ts:gmm1} and \ref{eqn:ts:gmm2}, where $\vec{\mu}_i$ is the mean and
  $S_i$ the covariance in a $d$-variate mixture model).
\begin{align}
  p(\vec{x}|\theta) &= \sum^{M}_{i=1}w_i g(\vec{x}|\vec{\mu}_i,S_i)
  \label{eqn:ts:gmm1}\\
  g(\vec{x}|\vec{\mu}_i,S_i) &=
  \frac{1}{\sqrt{(2\pi)^d|S_i|}}e^{-\frac{1}{2}(\vec{x}-\vec{\mu}_i)^T
    S_i^{-1}(\vec{x}-\vec{\mu}_i)} \label{eqn:ts:gmm2}
\end{align}

\newcite{Weinberger:2009} propose a pseudometric based on the
squared Mahalanobis distance $\mathcal{D}_{\mathbf{M}}$ (confusingly termed
Mahalanobis metric), depicted in Eq. \ref{eqn:ts:sqmaha}. The difference here is
that rather than using the covariance matrix, a custom positive semidefinite
matrix $M$ can be used.
\begin{equation}
\mathcal{D}_{\mathbf{M}}(\vec{x},\vec{y}) =
(\vec{x}-\vec{y})^T\mathbf{M}(\vec{x}-\vec{y}) \label{eqn:ts:sqmaha}
\end{equation}
This metric has been shown to be useful for tasks such as $k$NN classification
and related annotation tasks due to the its ability to adjust for variance in
scale in multidimensional data. The key difference between the general form
(Equation \ref{eqn:ts:maha}) and the pseudometric in Equation
\ref{eqn:ts:sqmaha} is that the latter can learn these scaling factors on
supervised data by adjusting matrix $\mathbf{M}$, whereas the former uses a
default scale adjustment based on the covariance matrix $S$.
In Chapter \ref{chapter:frame-semantic} we make
use of a derivation of this metric for our frame semantic parsing task.

\newcite{Cha:2007} provides a comprehensive survey of these distance metrics and
others. Also, we refer the interested reader to \newcite{Curran:2004} and
\newcite{Mitchell:2011} who provide further insight into this subject.

\section{Applications}\label{sec:ts:applications}

Semantic space representations can easily be plugged into a variety of NLP
related tasks. By providing richer representations of meaning than what can be
encompassed in a discrete model, such distributed representations have
demonstrated improvements on a wide range of tasks. The list below is by no
means exhaustive, but is intended to demonstrate the usefulness of such
representations across a broad range of areas in NLP.

Topic modelling has been explored using distributed representations,
particularly in the context of LDA \cite{Blei:2003,Steyvers:2005}. Other fields
include thesaurus extraction \cite{Grefenstette:1994,Curran:2004}, word-sense
discrimination \cite{Schutze:1998}, automated essay marking
\cite{Landauer:1997}, synonymy or word-word similarity
\cite{McDonald:2000,Griffiths:2007,Mitchell:2008}, named entity recognition
\cite{Collobert:2011,Turian:2010}, cross-lingual document classification
\cite{Klementiev:2012}, bilingual lexicon induction \cite{Haghighi:2008},
semantic priming \cite{Steyvers:2005,Landauer:1997}, discourse analysis
\cite{Foltz:1998,Kalchbrenner:2013} or selectional preference acquisition
\cite{Pereira:1993,Lin:1999}.

When considering representations of larger syntactic units this list will expand
even further. In this chapter and the next, we focus on word-level
representations only. From Chapter \ref{chapter:compositional} onwards we
discuss models for learning representations for higher linguistic units such as
phrases, sentences or documents.

\section{Summary}

In this chapter, we have surveyed the field of semantics within the context of
natural language processing. Following a brief exposition of the various strands
of semantic frameworks popular in the field, this chapter has subsequently
focused on distributed semantic representations in particular. Having explained
how such distributed representations can be learned, we are now able to begin
evaluating the hypothesis that this thesis sets out to solve.

The following chapter (Chapter \ref{chapter:frame-semantic}) begins this
evaluation, by describing a relatively simple model which employs distributed
representations to solve the frame-identification step of the semantic
frame-parsing task. As this is both an important and a popular task within the
NLP community, outperforming a series of prior models on this task with this
simple approach supports our hypothesis concerning the efficacy of distributed
representations and their use in solving challenging tasks in NLP.

\newcommand{\R}{\mathbb{R}}

\newcommand{\wsabie}{{\sc \small{Wsabie}}\xspace}
\newcommand{\wsMod}{{\sc \small{Wsabie Embedding}}\xspace}
\newcommand{\logWord}{{\sc \small{Log-Linear Words}}\xspace}
\newcommand{\logEmb}{{\sc \small{Log-Linear Embedding}}\xspace}

\newcommand{\FNtype}[1]{\textsf{\small{#1}}}
\newcommand{\fname}[1]{\FNtype{\textsc{#1}}}
\newcommand{\rname}[1]{\FNtype{#1}}

\renewcommand{\Theta}{\boldsymbol{\theta}}

\newcolumntype{C}{>{\lower.2ex\hbox to 9ex\bgroup\hss}c<{\hss\egroup}}
\newcolumntype{N}{>{\lower.2ex\hbox to 2.4ex\bgroup\hss}c<{\hss\egroup}}
\newcommand{\eat}[1]{\ignorespaces}

\chapter{Frame Semantic Parsing with Distributed
  Representations}\label{chapter:frame-semantic}

\emptyfootnote{The material in this chapter was originally presented in
  \newcite{Hermann:2014:ACLgoogle}. The aspects of the model related to
  distributed representations in the context of frame identification are
  primarily the first author's own work. The argument identification system used
  in the experimental part of this chapter is a standard system with some
  modifications by my co-authors and should not be counted towards the original
  work presented in this thesis.}

\begin{chapterabstract}
  This chapter investigates the use of distributed semantic representations for
  semantically complex tasks. We focus on the task of semantic frame
  identification and present a novel technique using distributed representations
  of predicates and their syntactic context for identifying semantic frames.
  This technique leverages automatic syntactic parses and a generic set of word
  embeddings.
  In order to evaluate this approach against the state-of-the-art, we combine
  our method with a standard argument identification system. In this
  combination, we outperform the former best model on FrameNet-style
  frame-semantic analysis, while reporting competitive results on the PropBank
  related task.
  These results strongly indicate the value of distributed representations for
  capturing semantics and for tackling tasks that benefit from semantic
  representations.
\end{chapterabstract}

\section{Introduction}\label{sec:fs:intro}

Having introduced distributed representations as a way to encode semantic
information in NLP, we put this concept to the test. This chapter investigates
the use of distributed representations for semantic frame identification---a
task that would clearly benefit from a degree of semantic understanding.

As pointed out in Chapter \ref{chapter:distrib}, there exists a large body of
literature on learning distributional representations.
However, much less work has been done on establishing whether such representations
truly capture semantics and whether they can be used for explicitly semantic
tasks.
Here, we address this question, which directly relates to the hypothesis stated
at the outset of this thesis. For this purpose, we develop a novel technique for
semantic frame identification that leverages distributed word representations,
and compare its performance against similar models that do not rely on
distributed representations in various experimental settings.
A benefit of the semantic frame identification task is that the
usefulness of semantic information in solving this task is almost self-evident.
Further, two popular formalisms and related test sets exist for this task
together with a suitably large body of prior work which allows for fair
comparison of our approach.
The empirical results in this chapter and our analysis of the various
experimental settings support our initial hypothesis and provide us with further
insight into the use and usefulness of distributed representations.

In our experiments we assume the presence of generic word embeddings constructed
independently of the task at hand, such as e.g. the distributions learned and
provided by \newcite{Collobert:2011}, \newcite{Turian:2010} or
\newcite{Al-Rfou:2013}. Given a predicate in a sentence, we extract
its syntactic context via an automatic parse; we learn to project the collection
of the word embeddings for each context word into a low-dimensional space.
Simultaneously, we learn an embedding for the semantic frames in our domain.
Both projections are learnt jointly to perform well on the supervised task.  At
prediction time, the context representation of a predicate is projected to the
low dimensional space and the nearest frame is chosen as our prediction. We
perform the learning within \wsabie \cite{Weston:2011}, a generic framework for
embedding instances and their labels in a shared low-dimensional space.

We apply our approach to frame-semantic parsing tasks on two frame-semantic
formalisms. First, we evaluate on the FrameNet corpus
\cite{Baker:1998,Fillmore:2003}, and show that we outperform the prior
state-of-the-art system \cite{Das:2014} on semantic frame identification. When
combined with a standard argument identification method (Appendix
  \ref{appendix:fs:argid}), we also report the best results on this task to
date.  Second, we present results on PropBank-style data
\cite{Palmer:2005,Meyers:2004,Marquez:2008}, where we achieve results on a par
with the prior state of the art \cite{Punyakanok:2008}.

This remainder of this chapter is structured as follows.
\S\ref{sec:fs:background}
provides the necessary background on semantic-frame parsing and the various
corpora and formalisms used in this chapter. \S\ref{sec:fs:overview} provides a
high-level overview of our model and \S\ref{sec:fs:model} then describes our
frame identification model and in particular the context-extraction method
developed to incorporate distributed representations into the frame
identification process.
Finally, we describe and discuss the empirical
evaluation of our approach (\S\ref{sec:fs:experiments} and
  \S\ref{sec:fs:conclusion}), as well as their implications for our further
study into the nature and application of distributed representations for
semantics.

\section{Frame-Semantic Parsing}\label{sec:fs:background}

According to the theory of frame semantics \cite{Fillmore:1982}, a semantic
frame represents an event or scenario, and possesses frame elements (or semantic
  \textbf{roles}) that participate in the event. From the perspective of Natural
Language Processing, frame semantics provide the formal basis for a parsing
task---not unlike phrase structure grammars for constituency-based parse trees
and dependency grammars for dependency-based parsing.

Frame-semantic analysis as a task in NLP was pioneered by \newcite{Gildea:2002},
who proposed a system for identifying semantic roles given a sentence and
frame-annotated frame-inducing word. \newcite{Gildea:2002} based their
work on the FrameNet formalism, with subsequent work in this area focusing on
either the FrameNet or PropBank framework. Among the two frameworks PropBank has
proved somewhat more popular since. Supervised approaches typically use the
corpora developed by the two projects of the same name.

Most work on frame-semantic \textit{parsing} has divided the task into
two major subtasks: \textit{frame identification}, namely the disambiguation of
a given predicate to a frame, and \textit{argument identification} (or semantic
  role labeling), the analysis of words and phrases in the sentential context
that satisfy the frame's semantic roles \cite{Das:2010,Das:2014}. Naturally,
there are some exceptions, wherein the task has been modelled using a pipeline
of three classifiers that perform frame identification, a binary stage that
classifies candidate arguments, and argument identification on the filtered
candidates \cite{Baker:2007,Johansson:2007}.

We focus on the first subtask, frame identification for given predicates, which
we tackle using distributed representations as a key input.  Subsequently, we
combine our distributed approach to this problem with a standard argument
identification method (Appendix \ref{appendix:fs:argid}). This allows us to
compare our approach with the state-of-the-art on the full frame-semantic
parsing task.

Frame-semantic analysis has received a significant boost in attention owing to
the CoNLL 2004 and 2005 shared tasks \cite{Carreras:2004,Carreras:2005} on
PropBank semantic role labeling (SRL). At least since then, it has been treated
as an important problem in NLP. That said, research has mostly focused on
argument identification, the second of the two subtasks described above,
entirely skipping the frame disambiguation step and its potential interaction
with argument analysis.

Frame-semantic parsing is closely related to SRL and describes the full process
of resolving a predicate sense into a frame and the subsequent analysis of the
frame's arguments. Therefore it could be viewed as a strict extension of SRL for
situations where sentences may contain multiple frames and moreover where those
frames require labeling on top of argument identification. Due to the
differences between PropBank and FrameNet, as detailed below, work in this area
focuses on the FrameNet full text annotations of the SemEval'07 data
\cite{Baker:2007}. The original FrameNet corpus is unsuitable for this task as
it consists of exemplar sentences with a single annotated frame each.

Notable work on frame-semantic parsing includes \newcite{Johansson:2007}, the
best performing system at SemEval'07 and \newcite{Das:2010} who significantly
improved performance before presenting the current state-of-the-art system in
\newcite{Das:2014}. \newcite{Matsubayashi:2009} provide an overview of the
efficacy of various argument identification features exploiting different types
of taxonomic relations to generalize over roles.

\begin{figure}[t]
\centering
\includegraphics[width=\columnwidth]{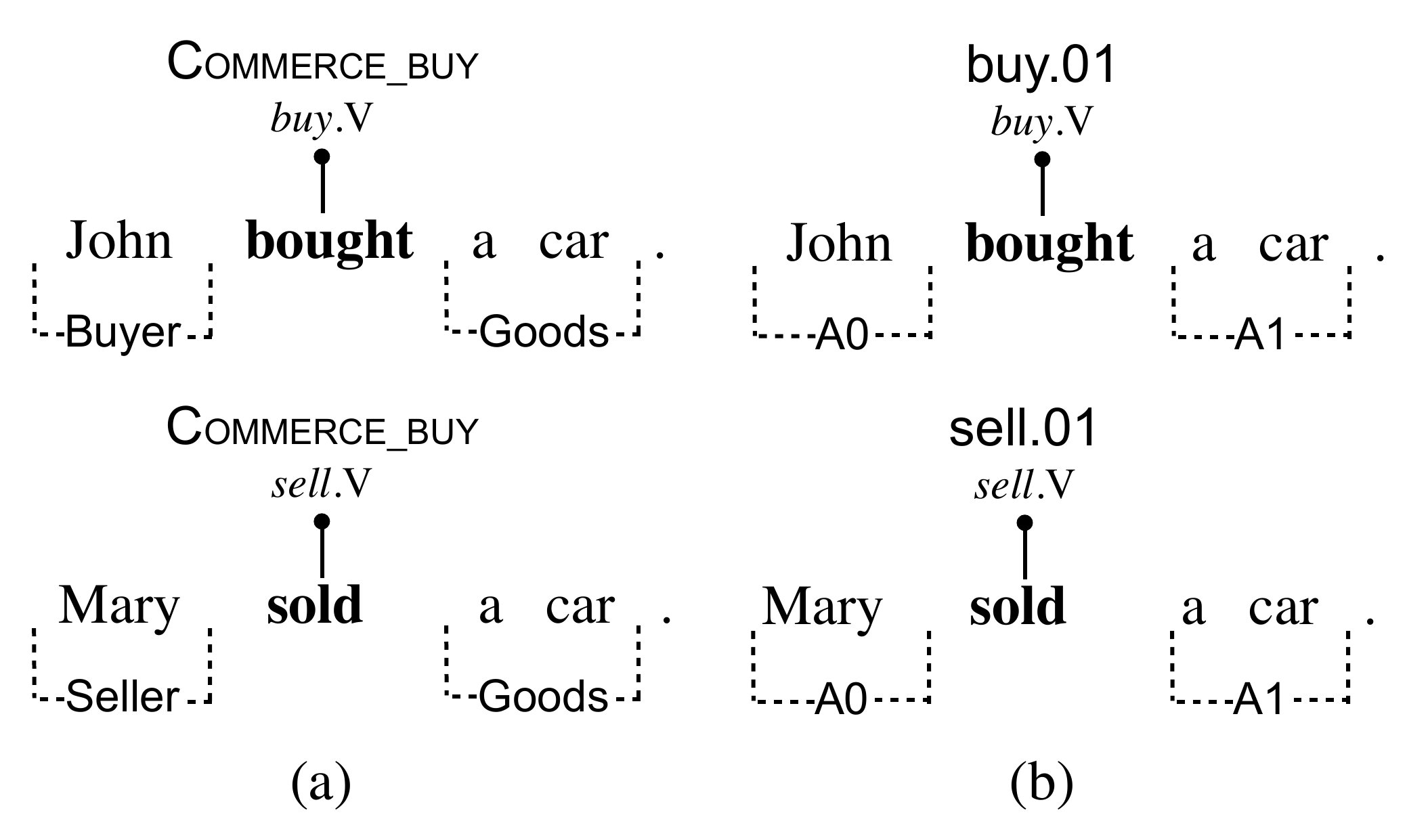}
\caption[Example sentences with frame-semantic analyses]{Example sentences with frame-semantic analyses. FrameNet annotation
conventions are used in (a) while (b) denotes PropBank conventions.\label{fig:framenet-propbank}
}
\vspace{-0.2in}
\end{figure}

\subsection{FrameNet}

The FrameNet project \cite{Baker:1998} is a lexical database that contains
information about words and phrases (represented as lemmas conjoined with a
  coarse part-of-speech tag) termed as lexical units, with a set of semantic
frames that they could evoke. For each frame, there is a list of associated
frame elements (or roles, henceforth), that are also distinguished as core or
non-core.\footnote{Additional information such as finer distinction of the
  coreness properties of roles, the relationship between frames, and that of
  roles are also present, but we do not leverage that information in this work.}
Sentences are annotated using this universal frame inventory.  For example,
consider the pair of sentences in Figure~\ref{fig:framenet-propbank}(a).
\fname{Commerce\_buy} is a frame that can be evoked by morphological variants of
the two example lexical units \textit{buy}.\textsc{V} and
\textit{sell}.\textsc{V}.  \rname{Buyer}, \rname{Seller} and \rname{Goods} are
some example roles for this frame.

\subsection{PropBank}

The PropBank project \cite{Palmer:2005} is another popular resource related to
semantic role labeling. The PropBank corpus has verbs annotated with sense
frames and their arguments. Like FrameNet, it also has a lexical database that
stores type information about verbs, in the form of sense frames and the
possible semantic roles each frame could take. There are modifier roles that are
shared across verb frames, somewhat similar to the non-core roles in FrameNet.
Figure~\ref{fig:framenet-propbank}(b) shows annotations for two verbs ``bought''
and ``sold'', with their lemmas (akin to the lexical units in FrameNet) and
their verb frames \FNtype{buy.01} and \FNtype{sell.01}. Generic core role labels
(of which there are seven, namely \rname{A0}-\rname{A5} and \rname{AA}) for the
verb frames are marked in the figure.\footnote{NomBank \cite{Meyers:2004} is a
  similar resource for nominal predicates, but we do not consider it in our
  experiments.} A key difference between the two annotation systems is that
PropBank uses a local frame inventory, where frames are predicate-specific.
Moreover, role labels, although few in number, take specific meaning for each
verb frame.  Figure~\ref{fig:framenet-propbank} highlights this difference:
while both \textit{sell}.\textsc{v} and \textit{buy}.\textsc{v} are members of
the same frame in FrameNet, they evoke different frames in PropBank.  In spite
of this difference, nearly identical statistical models could be employed for
both frameworks.

\section{Model Overview}\label{sec:fs:overview}

We model the frame-semantic parsing problem in two stages: \textbf{frame
  identification} and \textbf{argument identification}. As mentioned in
\S\ref{sec:fs:intro}, these correspond to a frame disambiguation stage and a
stage that finds the various arguments that fulfil the frame's semantic roles
within the sentence, respectively.
We are particularly interested in the frame disambiguation or identification
stage. To exemplify this stage, consider PropBank, where the predicate
\textit{buy} (or the lexical unit\footnote{PropBank never formally uses the
    term lexical unit, whose usage we adopt from the frame semantics
    literature.} \textit{buy}.\textsc{V}) has three verb frames. We want to
learn to disambiguate between these frames given a sentential context.

This framework is similar to that of \newcite{Das:2014}, with the difference
that \newcite{Das:2014} solely focus on FrameNet corpora. The main novelty of
our approach lies in the frame identification stage (\S\ref{sec:fs:model}). Note
that this two-stage approach is unusual for the PropBank corpora when compared
to prior work, where the vast majority of published papers have not focused on
the verb frame disambiguation problem at all, only focusing on the role
labelling stage. We refer the interested reader to the overview paper of
\newcite{Marquez:2008} for more information on this.

We approach the frame identification stage from a distributed perspective and
present a model that takes word embeddings (distributed representations in
  $\mathbb{R}^n$) as input and learns to identify semantic frames given these
embeddings. Specifically, we use word representations to capture the syntactic
context of a given predicate instance, and to represent this context as a
vector.  This can be exemplified using a short sentence such as ``He runs the
company''.  Here, the predicate \emph{runs} has two syntactic dependents---a
subject and direct object (but no prepositional phrases or clausal complements).

Making use of these syntactic dependencies, we could represent the context of
\emph{runs} as a structured vector with slots for all possible types of
dependents warranted by a syntactic parser. With a structured vector, we refer
to a vector where certain spans over indices are reserved for certain content.
So, in this case, this hypothetical structured vector might contain a slot for
the subject dependency at $0\ldots n$, a second slot corresponding to the direct
object complement in $n{+}1\ldots 2n$, a third slot corresponding to a clausal
complement dependent in $2n{+}1\ldots 3n$, and so forth.  Overall, this results
in a vector of $\R^{kn}$, where $k$ is the number of dependency types available
in a given parsing formalism.

Given our example sentence ``He runs the company.'', this would result in a
vector representation where the subject dependent slot contains the
embedding of \emph{he} and the direct object dependent slot contains the
embedding for \emph{company}, with all other slots empty. For the purposes of
our model, we define empty slots as being equal to zero.

Now, given such input vectors for our training data, we learn a mapping from
this high-dimensional space $\R^{kn}$ into a lower dimensional space $\R^m$.
Simultaneously, the model learns an embedding for all the possible labels (i.e.
  the frames in a given lexicon). At inference time, the predicate-context is
mapped to the low dimensional space, and we choose the nearest frame label as
our classification.
There are multiple reasons for taking this approach. First, learning a mapping
into a lower dimensional space $\R^m$ makes the model more efficient at run
time, as the nearest neighbour classification takes place in a smaller search
space. More importantly, the two mappings of training data and labels into the
joint space describe the actual model parameter space, with our update function
being able to learn which dimensions of the larger input space $\R^{kn}$ are
relevant; to what extent; and in which combination. Third, by placing the labels
in this joint space---rather than learning a multi-class classifier from that
space to the set of labels---joint inference of the model is simplified. Related
experiments within the context of image annotation have highlighted the
effectiveness of this approach \cite{Weston:2011}.

\section{Frame Identification with Embeddings}
\label{sec:fs:model}\label{sec:fs:model:overview}

\begin{figure*}
 \captionsetup{font=small}
\centering
\includegraphics[scale=0.5]{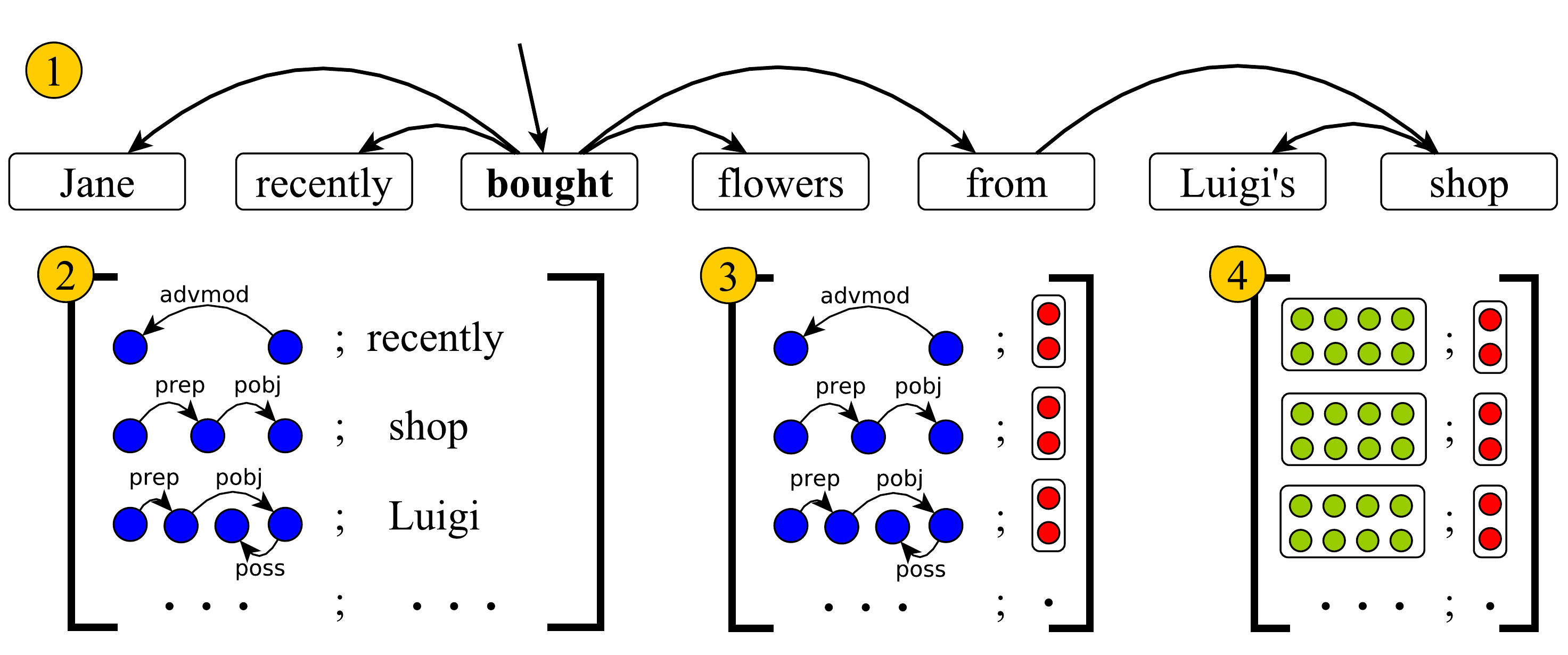}
\caption[Context representation extraction for the embedding model]{
Context representation extraction for the embedding model. Given a dependency
parse (1) the model extracts all words matching a set of paths from the frame
evoking predicate and its direct dependents (2). The model computes a composed
representation of the predicate instance by using distributed vector
representations for words (3) -- the (red) vertical embedding vectors for each
word are concatenated into a long vector. Finally, we learn a
linear transformation function parametrised by the context blocks (4).
}          \label{fig:fs:embeddings}
\end{figure*}

We continue using the example sentence from \S\ref{sec:fs:overview}: ``He
runs the company'', for which we want to disambiguate the frame of \emph{runs}
in context. First, we extract the words in the syntactic context of \emph{runs}
and concatenate their word embeddings as described in
\S\ref{sec:fs:overview} to create an initial vector space representation.
Using this vector representation, which will be high dimensional, we learn a
mapping into a lower dimensional space. In this low-dimensional space we also
learn representations for each possible frame label. This enables us to posit
the task of frame identification as a distance measure, where each frame
instance is resolved to the closest suitable label in the low-dimensional space.
In order to accomplish this, we use an objective function that ensures that the
correct frame label is as close as possible to the mapped context
representations, while competing frame labels are farther away.

Formally, let $x$ represent the actual sentence with a marked predicate, along
with the associated syntactic parse tree; let our initial representation of the
predicate context be $g(x)$. Suppose that the word embeddings we start with are
of dimension $n$. Then $g$ is a function from a parsed sentence $x$ to
$\R^{kn}$, where $k$ is the number of syntactic context types considered by $g$
and will vary depending on $g$.
For instance, assume $g_z$ to only consider clausal complements and direct
objects. Then $g_z:X\rightarrow \R^{2n}$, with $0\ldots n$ reserved for the
clausal complement and $n{+}1\ldots 2n$ reserved for direct objects.
In the case of our example sentence, the resultant vector would have zeros in
positions $0\ldots n$ and the embedding of the word \emph{company} in positions
$n{+}1\ldots 2n$:
\begin{equation}
g_z(x) = [0,\ldots,0,\text{embedding of \emph{company}}]\nonumber
\end{equation}
The actual context representation extraction function $g$ we use in our
experiments is somewhat more complex than this. We describe this function in
\S\ref{sec:fs:model:contex}.

Next, we describe the mapping from $\R^{kn}$ to the low dimensional joint space.
Let the low dimensional space we map to be $\R^m$ and the learned
mapping be $M: \R^{kn} \rightarrow \R^m$. The mapping $M$ is a linear
transformation, and we learn it using the \wsabie algorithm~\cite{Weston:2011}.
\wsabie also learns an embedding for each frame label ($y$, henceforth).  In our
setting, this means that each frame corresponds to a point in $\R^m$. Given $F$
possible frames we can store those parameters in an $F \times m$ matrix, one
$m$-dimensional point for each frame, which we will refer to as the linear
mapping $Y$. Thus, we have two mappings into $\R^m$:
\begin{equation}
  \begin{aligned}
    &M: \R^{kn} \rightarrow \R^m \\
    &Y: \{1\ldots F\} \rightarrow \R^m
  \end{aligned}
\end{equation}

Assume a frame lexicon which stores all available frames, corresponding semantic
roles and the lexical units associated with the frame. Let the lexical unit (the
  lemma conjoined with a coarse POS tag) for the marked predicate be $\ell$. We
denote the frames that associate with $\ell$ in the frame lexicon and our
training corpus as $F_{\ell}$.

As discussed in \S\ref{sec:fs:overview}, we follow the learning method for joint
embedding spaces as introduced in \newcite{Weston:2011}, following their success
in using this model to learn distributed representations for annotating images.
\wsabie performs gradient-based updates on an objective that tries to minimize
the distance between $M(g(x))$ and the embedding of the correct label $Y(y)$,
while maintaining a large distance between $M(g(x))$ and the other possible
labels $Y(\bar{y})$ in the confusion set $F_\ell$. At disambiguation time, we
use the dot product as our distance metric, meaning that the
model chooses a label by computing the $\mbox{argmax}_y s(x,y)$ where $s(x,y) =
M(g(x)) \cdot Y(y)$, where the $\mbox{argmax}$ iterates over the possible frames
$y \in F_{\ell}$ if $\ell$ was seen in the lexicon or the training data, or $y
\in F$, if it was unseen.  This disambiguation scheme is similar to the one
adopted by \newcite{Das:2014}, but they use unlemmatised words to define their
confusion set.  Model learning is performed using the margin ranking loss
function as described in \newcite{Weston:2011}, and in more detail in
\S\ref{sec:fs:learning}.

Since \wsabie learns a single mapping $M$ from $g(x)$ to $\R^m$, parameters are
shared between different words and different frames. So for example ``He
\emph{runs} the company'' could help the model disambiguate ``He \emph{owns} the
company.''  Moreover, since $g(x)$ relies on word embeddings rather than word
identities, information is shared between words. For example ``\emph{He} runs
\emph{the company}'' could help us to learn about ``\emph{She} runs \emph{a
  corporation}''.
It is due to this kind of information sharing that we believe that distributed
representations are a suitable vehicle for encoding semantics and by extension
the continuous and multi-dimensional degrees of semantic similarity that could
be useful for learning from examples such as above. These types of
semantically motivated inference problems are also one of the reasons why we
selected semantic frame-identification as a good problem to evaluate the
efficacy of semantic distributed representations.

\subsection{Context Representation Extraction} \label{sec:fs:model:contex}

In principle $g(x)$ could be any feature function. For the purposes of our
investigation we focus on a particular variant, where our representation is a
block vector where each block corresponds to a syntactic position relative to
the frame inducing token (e.g. the predicate). Each block's value of course
corresponds to the embedding of the word at that syntactic position, as we
described using the example of $g_z(x)$ above.
Therefore we have $g(x) \in \R^{kn}$, where $n$ is the dimension of the input
word embeddings and $k$ is the number of positions we are modelling, and hence
$k$ is the number of blocks. We parse the input sentence using a dependency
parser, and consider positions relative to the frame inducing element in terms
of the parse tree.

We consider two types of relative positions. First, we consider syntactic
dependents in a fashion corresponding to the example provided in
\S\ref{sec:fs:model:overview}. To elaborate, the positions of interest are the
labels of the direct dependents of the predicate, so $k$ is the number of labels
that the dependency parser can produce. For example, if the label on the edge
between \emph{runs} and \emph{He} is \FNtype{nsubj}, we would put the embedding
of \emph{He} in the block corresponding to \FNtype{nsubj}. If a label occurs
multiple times, then the embeddings of the words below this label are averaged.

Unfortunately, using only the direct dependents can miss a lot of useful
information.  For example, topicalisation can place discriminating information
farther from the predicate. Consider ``He \textit{runs} the \textit{company}.''
vs. ``It was the \textit{company} that he \textit{runs}.''  In the second
sentence, the discriminating word, \emph{company}, dominates the predicate
\emph{runs}.

Similarly, predicates in embedded clauses may have a distant agent which cannot
be captured using direct dependents. For example, consider the sentences
``\textit{The athlete} \textit{ran} the marathon." vs. ``\textit{The athlete}
prepared himself for three months to \textit{run} the marathon." In the second
example, for the predicate \emph{run}, the agent \emph{The athlete} is not a
direct dependent, but is connected via a longer dependency path.

For these reasons, and in order to capture more relevant context, we include a
second group of syntactic dependents defined as follows: We scanned the training
data for a given task (either the PropBank or the FrameNet domains) for the
dependency paths that connected the gold predicates to the gold semantic
arguments. This set of dependency paths were deemed as possible positions in
the initial vector space representation.

Thus, the final cardinality of $k$ is the sum of the number of scanned gold
dependency path types plus the number of dependency labels in our parser.
Given a predicate in its sentential context, we therefore extract only those
context words that appear in positions warranted by the accordingly defined set.
See Figure~\ref{fig:fs:embeddings} for an illustration of this process. Note that
in this figure we posit the linear mapping $M: \R^{kn} \rightarrow \R^m$ as a
fully weighted additive function of embeddings, with each input block $i \in
\{1..k\}$ being associated with a submatrix $M_i \in \R^{n\times m}$. Note that
this reformulation results in exactly the same representations as the original
formulation where empty blocks are represented with zeros.

\subsection{Learning} \label{sec:fs:learning}

We model our objective function following \newcite{Weston:2011}, by using a
weighted approximate-rank pairwise loss, learned with stochastic gradient
descent.  The mapping from $g(x)$ to the low dimensional space $\R^m$ is a
linear transformation, so the model parameters to be learnt are the matrix $M
\in \R^{kn \times m}$ as well as the embedding of each possible frame label,
represented as another matrix $Y \in \R^{F \times m}$ when there are $F$ frames
in total. The training objective function minimizes
\begin{equation}
  \sum_{x} \sum_{\bar{y}}
  L\big(rank_y(x)\big)   \left[\gamma + s(x,y) - s(x,\bar{y})\right]_+\nonumber
\end{equation}
where $x, y$ are the training inputs and their corresponding correct frames, and
$\bar{y}$ are negative frames (that do not correspond with $x$), $\gamma$ is the
margin and $s(x,y)$ is the score between an input and a frame. Further, $[x]_+ =
max(0,x)$ denotes the standard hinge loss and $rank_y(x)$ is the rank of the
positive frame $y$ relative to all the negative frames:
\begin{equation}
rank_y(x) = \sum_{\bar{y}} I( s(x,y) \leq \gamma + s(x,\bar{y}) ),\nonumber
\end{equation}
and $L(\eta)$ converts the rank to a weight, e.g. $L(\eta) = \sum_{i=1}^\eta
1/i$.

The purpose of $L$ is to convert the rank to a weighting of the given pairwise
constraint comparing $d$ and $\bar{d}$.  Choosing $L(\eta) = C \eta$ for any
positive constant $C$ optimizes the mean rank, whereas a weighting such as
$L(\eta) = \sum_{i=1}^{\eta} 1/i$ (adopted here) optimizes the top of the ranked
list, as described in \newcite{Usunier:2009}.
To train with such an objective, we can employ stochastic gradient descent.

For speed the computation of $rank_y(x)$ is then replaced with a sampled
approximation: sample $N$ items $\bar{y}$ until a violation is found, i.e.
$\max(0, \gamma + s(x,\bar{y}) - s(x,y)) > 0$ and then approximate the rank
with $(F - 1)/N$; see  \newcite{Weston:2011} for more details on this procedure.
For the choices of the stochastic gradient learning rate, margin ($\gamma$) and
dimensionality ($m$), please refer to
\S\ref{sec:fs:experiments:framenet}-\S\ref{sec:fs:experiments:propbank}.

Note that an alternative approach could learn only the matrix $M$, and then use
a $k$-nearest neighbour classifier in $\R^m$, as in~\newcite{Weinberger:2009}.
The advantage of learning an embedding for the frame labels is that at inference
time we need to consider only the set of labels for classification rather than
all training examples. Additionally, since we use a frame lexicon that gives us
the possible frames for a given predicate, we usually only consider a handful of
candidate labels. If we used all training examples for a given predicate for
finding a nearest-neighbour match at inference time, we would have to consider
many more candidates, making the process very slow.

\section{Experiments} \label{sec:fs:experiments}

In this section, we present our experiments in frame-semantic parsing. As
already motivated in \S\ref{sec:fs:intro}, we first evaluate our
frame identification system in isolation to evaluate whether distributed
representations can be a suitable choice for a problem. Subsequently, we
combine our system with a standard argument identification system (Appendix
  \ref{appendix:fs:argid}), which allows us to compare performance with the
previous state of the art.

While we used a standard approach for the argument identification step, we used
a custom implementation and a number of new features. Therefore, in order to
evaluate the effects of the distributed model fairly, we also combine our
argument identification model with a standard log-linear frame identification
model for comparison (see \ref{sec:fs:experiments:baselines}).

\subsection{Data} \label{sec:fs:experiments:data}

We run experiments on both FrameNet- and PropBank-style structures. For
FrameNet, we use the full-text annotations in the FrameNet 1.5
release\footnote{\url{https://framenet.icsi.berkeley.edu}.} which was
used by \newcite{Das:2014}. We used the same test set as Das~et~al. containing
23 documents with 4,458 predicates. Of the remaining 55 documents, 16
documents were randomly chosen for development (Appendix \ref{appendix:fs:dev}),
resulting in a development set with a simlar number of predicates.

For experiments with PropBank, we used the OntoNotes corpus \cite{Hovy:2006},
version 4.0, and only made use of the Wall Street Journal documents. Following
convention, we used sections 2-21 for training, section 24 for development and
section 23 for testing. This split also resembles the setup used by
\newcite{Punyakanok:2008}. All the verb frame files in OntoNotes were used for
creating our frame lexicon.

\subsection{Frame Identification Baselines}\label{sec:fs:experiments:baselines}

As briefly rationalised above, we also implemented a set of baseline models with
the purpose of isolating the effect of the distributed model within the full
frame semantic parsing setup.
Our baselines are log-linear models with varying feature configurations. At
training time, the baseline models use the following probability:
\begin{equation}
  p(y | x, \ell) =
  \frac{
    e^{ \boldsymbol{\psi} \cdot \mathbf{f}(y, x, \ell)}
  }{
    \sum_{\bar{y} \in F_\ell}
    e^{\boldsymbol{\psi} \cdot \mathbf{f}(\bar{y}, x, \ell)}
  }
\end{equation}
At test time, this model chooses the best frame as $\mbox{argmax}_y
\boldsymbol{\psi} \cdot \mathbf{f}(y, x, \ell)$ where $\mbox{argmax}$ iterates
over the possible frames $y \in F_{\ell}$ if $\ell$ was seen in the lexicon or
the training data, or $y \in F$, if it was unseen, like the disambiguation
scheme of \S\ref{sec:fs:model:overview}. We train this model by maximizing the
$L_2$ regularized log-likelihood. We use L-BFGS for training, with
the regularisation constant set to 0.1 in all experiments.

For comparison with our model from \S\ref{sec:fs:model}, which we call
\textsc{\small{Wsabie Embedding}}, we implemented two baselines with the
log-linear model.
Both of these baseline models rely on a variation of the context extraction
mechanism developed for the distributed model and described
in~\S\ref{sec:fs:model:contex}.

The first baseline reuses the context extraction system, but uses word
identities rather than word embeddings for its representation. The model first
conjoins the word identities with their respective dependency paths, and second
uses the words themselves as backoff features. This model could be viewed as a
standard NLP approach for the frame identification problem. However, despite its
simplicity, we find that it performs competitively with the state of
the art.  We refer to this baseline model as \textsc{\small{Log-Linear Words}}.

Our second baseline decouples the concept of using embeddings from the specific
training method applied here. Specifically, it decouples the \wsabie training
from the embedding input, and trains a log-linear model using the embeddings as
input. Thus, this second baseline model (\textsc{\small{Log-Linear Embedding}})
has exactly the same input representation as the \textsc{\small{Wsabie
    Embedding}} model. Unlike the \textsc{\small{Wsabie Embedding}} model,
however, the \textsc{\small{Log-Linear Embedding}} model does not learn
distributed representations for labels in a joint space but instead simple
probabilities given the high-dimensional distributed input.

\subsection{Common Experimental Setup}\label{sec:fs:experiments:common}

We process our PropBank and FrameNet training, development and test corpora with
a shift-reduce dependency parser that uses the Stanford conventions
\cite{Marneffe:2013} and which relies on an arc-eager transition system with
beam size of 8; the parser and its features are described
by~\newcite{Zhang:2011}.

Before parsing the data, it is tagged with a POS tagger trained using a
conditional random field \cite{Lafferty:2001} with the following emission
features: the word, the word cluster, word suffixes of length 1, 2 and 3,
capitalization, whether it has a hyphen, digit and punctuation. Beyond the bias
transition feature, we have two cluster features for the left and right words in
the transition. We use Brown clusters learned using the algorithm of
\newcite{Uszkoreit:2008} on a large English newswire corpus for cluster
features. We use the same word clusters for the argument identification features
in Table~\ref{tab:argid-features}.

We learn the initial embedding representations for our frame identification
model (\S\ref{sec:fs:model}) using a deep neural language model similar to the
one proposed by \newcite{Bengio:2003}. We use 3 hidden layers each with 1024
neurons and learn a 128-dimensional embedding from a large corpus containing
over 100 billion tokens. In order to speed up learning, we use an unnormalised
output layer and a hinge-loss objective. The objective tries to ensure that the
correct word scores higher than a random incorrect word, and we train with
minibatch stochastic gradient descent.

\subsection{Experimental Setup for FrameNet}\label{sec:fs:experiments:framenet}

\subsubsection{Hyperparameters}

We optimise the hyperparameters for the \wsabie training on our development
data. There are three relevant parameters: the stochastic gradient descent
learning rate, which we set to $0.0001$, the margin $\gamma$ which we set to
$0.01$ and the dimensionality of the joint space $\R^m$, where we set $m$ to
$256$. See Table \ref{tab:fs:hyperparams} for an overview of the parameters
we evaluated in the hyperparameter search.
Our hyperparameter sweep optimises for \textit{ambiguous} frame identification
accuracy, that is the performance of the model with respect to identifying the
frames of lexical units with more than one possible semantic frame.

\begin{table}[t]
  \centering
  \begin{tabular}{@{}llll@{}}
    \toprule
    Parameter & Choices & \multicolumn{2}{c}{Selected Value} \\
    \cmidrule{3-4}
    & & FrameNet & PropBank \\
    \midrule
    Learning Rate & $\{0.01, 0.001, 0.0001\}$ & $0.0001$ & $0.01$\\
    Margin ($\gamma$) & $\{1, 0.1, 0.01, 0.001\}$ & $0.001$ & $0.01$ \\
    Dimensionality ($m$) & $\{256, 512\}$ & $256$ & $512$ \\
    \bottomrule
  \end{tabular}
  \caption[Hyperparameter search space for the FrameNet and PropBank
    experiments]{Hyperparameter search space for the FrameNet and PropBank
    experiments, as well as the chosen hyperparameters for both tasks.}
  \label{tab:fs:hyperparams}
\end{table}

\subsubsection{Argument Candidates}\label{sec:fs:exp:fn:argid}

The candidate argument extraction method used for the FrameNet data (as
  described in Appendix \ref{appendix:fs:argid}) was adapted from the algorithm of
\newcite{Xue:2004} applied to dependency trees. Since the original algorithm was
designed for verbs, we augmented the set of rules to also handle non-verbal
predicates as follows:
\begin{enumerate}
  \item We added the predicate itself as a candidate argument.
  \item We added the span ranging from the sentence position to the right of the
    predicate to the rightmost index of the subtree headed by the predicate's
    head. This helps capture cases like ``a \textit{few} \underline{months}'',
    where \textit{few} is the predicate and \underline{months} is the argument.
\item We added the span ranging from the leftmost index of the subtree headed by
  the predicate's head to the position immediately before the predicate, for
  cases like ``\underline{your gift} \textit{to} Goodwill'', where \textit{to} is
    the predicate and \underline{your gift} is the argument.
\end{enumerate}

Note that \newcite{Das:2014} describe the state of the art in FrameNet-based
analysis, but their argument identification strategy considered all possible
dependency subtrees in a parse, resulting in a much larger search space.

\subsubsection{Frame Lexicon}\label{sec:fs:exp:fn:lex}

In our experimental setup, we scanned the XML files in the ``frames'' directory
of the FrameNet 1.5 release, which lists all the frames, the corresponding roles
and the associated lexical units, and created a frame lexicon to be used in our
frame and argument identification models.

A frame lexicon extracted using such a process encompasses all relevant units
for our test case. Therefore, at frame disambiguation time, we only have to
score the frames in $F_\ell$ for each predicate $\ell$. In essence this means
that for all instances we only had to choose a semantic frame from a small list
of candidates rather than from the global set of available frames in FrameNet.
Please refer back to \S\ref{sec:fs:model:overview} and
\S\ref{sec:fs:experiments:baselines} for more details on this. We refer to this
setup as \textsc{\small{Full Lexicon}}.

While attempting to compare our system with the prior state of the art
\cite{Das:2014}, we noted that they found several unseen predicates at test
time. This is due to their frame lexicon creation method: Instead of scanning
the frame files, \newcite{Das:2014} constructed a frame lexicon by scanning
FrameNet's exemplars and the training corpus. For fair comparison, we replicated
their lexicon and removed all instances containing their \textit{unseen}
predicates from our training data, thereby having to select semantic frames from
$F$ rather than $F_\ell$ for those predicates during test time. We refer to this
setup as \textsc{\small{Semafor Lexicon}}, and further report results on the set
of unseen instances used in that paper.

\subsubsection{ILP Constraints}\label{sec:fs:experiments:framenet:ilp}

For FrameNet, we used three integer linear programming (ILP) constraints during
argument identification (Appendix \ref{appendix:fs:argid}). First, each span
could have only one role; second, each core role could be present only once; and
third, all overt arguments had to be non-overlapping.

\subsection{Experimental Setup for PropBank}\label{sec:fs:experiments:propbank}

From the point of view of frame identification, the PropBank and FrameNet
annotation schemes differ in an important respect.  PropBank frames are specific
to a predicate, so for example \FNtype{run.01} is a sense of run, which means to
operate a machine or company, and is very similar in meaning to
\FNtype{operate.01}, but when the system sees the word \emph{run} in a sentence,
the sense \FNtype{run.01} is available while \FNtype{operate.01} is not
available.  In contrast, FrameNet frames are shared between verbs, and both
\emph{operate} and \emph{run} can evoke the frame \fname{Operating\_a\_system}.
Because of this difference, the set of experiments possible for PropBank and
FrameNet differ in a number of aspects.

\subsubsection{Hyperparameters}

We report the hyperparameters in Table \ref{tab:fs:hyperparams}, which we chose
by searching over the same space as described for the FrameNet case above.
Again, we optimised for ambiguous lexical units in the hyperparameter search.

\subsubsection{Argument Candidates}

For PropBank we use the algorithm of \newcite{Xue:2004} applied to dependency
trees. As PropBank only considers predicates, there was no need to adjust the
algorithm.

\subsubsection{Frame Lexicon}

For the PropBank experiments we scanned the frame files for propositions in
OntoNotes 4.0, and stored possible core roles for each verb frame. The lexical
units were simply the verb associating with the verb frames. There were no
unseen verbs at test time.

\subsubsection{ILP Constraints}

For PropBank, we used five ILP constraints. The first three were the same as the
ones for FrameNet (\S\ref{sec:fs:experiments:framenet:ilp}). In addition, we also
made sure that: 4) continuation arguments of the form \FNtype{C-A*} could appear
only after the corresponding overt \FNtype{A*} argument, and 5) relative
arguments of the form \FNtype{R-A*} could appear only if the corresponding
\FNtype{A*} argument is present. These constraints are identical to the ones
proposed and used in \newcite{Punyakanok:2008}.

\subsection{FrameNet Results}\label{sec:fs:results:framenet}

\begin{table}[tp]
  \centering
  \renewcommand{\arraystretch}{1.1}
  \begin{tabular}{@{}l@{}c@{}c@{}c@{}c@{}c@{}c@{}c@{}c@{}}
    \toprule
    Model & \phantom{ac} & \multicolumn{3}{c}{\textsc{\small{Semafor Lexicon}}}
          & \phantom{ac}
          &\multicolumn{3}{c}{\textsc{\small{Full Lexicon}}}\\
    \cmidrule{3-5}
    \cmidrule{7-9}
    && All & Ambiguous & Rare  && All & Ambiguous & Rare\\
    \midrule
    \textsc{\small{Log-Linear Words}}    && 89.21 & 72.33 & 88.22 && 89.28 & 72.33 & 88.37 \\
    \textsc{\small{Log-Linear Embed.}}   && 88.66 & 72.41 & 87.53 && 88.74 & 72.41 & 87.68 \\
    \textsc{\small{Wsabie Embedding}} && \textbf{90.78} & \textbf{76.43} & \textbf{90.18} && \textbf{90.90} & \textbf{76.83} & \textbf{90.18}\\
    \bottomrule
  \end{tabular}
  \caption[Frame identification results on FrameNet development data]{
    Frame identification results on the FrameNet development data
    (\S\ref{sec:fs:experiments:data}). The \textsc{\small{Semafor Lexicon}}
    allows comparison with the prior state of the art \cite{Das:2014}. See
    \S\ref{sec:fs:exp:fn:lex} for a description of the two lexica. Results are
    reported on all frame instances (\textit{All}), all seen but ambiguous instances
    (\textit{Ambiguous}) and all rare but ambiguous predicates, appearing 10 or
    fewer times in the training data (\textit{Rare}).
  }\label{tab:framenetfidres:dev}
\end{table}

\begin{table}[tp]
\centering
\begin{tabular}{@{}l@{}c@{}c@{}c@{ }c@{ }c@{}c@{}c@{}c@{}c@{}}
  \toprule
  Model & \phantom{ab} & \multicolumn{4}{c}{\textsc{\small{Semafor Lexicon}}} &
  \phantom{ab} & \multicolumn{3}{c}{\textsc{\small{Full Lexicon}}}\\
  \cmidrule{3-6}
  \cmidrule{8-10}
  && All & Ambiguous & Rare & Unseen && All & Ambiguous & Rare \\
  \midrule
  \newcite{Das:2014} supervised && 82.97 & 69.27 & 80.97 & 23.08 &&&& \\
  \newcite{Das:2014} best && 83.60 & 69.19 & 82.31 & 42.67 &&&& \\
  \midrule
  \textsc{\small{Log-Linear Words}} && 84.53 & 70.55 & 81.65 & 27.27 && 87.33 & 70.55 & 87.19 \\
  \textsc{\small{Log-Linear Embed.}} && 83.94 & 70.26 & 81.03 & 27.97 && 86.74 & 70.26 & 86.56 \\
  \textsc{\small{Wsabie Embedding}}  && \textbf{86.49} & \textbf{73.39} & \textbf{85.22} & \textbf{46.15} && \textbf{88.41} & \textbf{73.10} & \textbf{88.93}\\
  \bottomrule
\end{tabular}
\caption[Frame identification results on FrameNet test data]{
  Frame identification results on the FrameNet test data
  (\S\ref{sec:fs:experiments:data}). Results are structured as in Table
  \ref{tab:framenetfidres:dev}. The \textit{Unseen} column compares performance
  of the unseen instances as reported in \cite{Das:2014}.
  }\label{tab:framenetfidres}
\end{table}

\begin{table}[tp]
  \centering
  \begin{tabular}{@{}l@{}c@{}x{0.7in}@{}x{0.7in}@{}x{0.7in}@{}c@{}x{0.7in}@{}x{0.7in}@{}x{0.7in}@{}}
    \toprule
    Model & \phantom{ab} & \multicolumn{3}{c}{\textsc{\small{Semafor Lexicon}}} &
    \phantom{ab} & \multicolumn{3}{c}{\textsc{\small{Full Lexicon}}}\\
    \cmidrule{3-5}
    \cmidrule{7-9}
    && Precision & Recall & $F_1$ && Precision & Recall & $F_1$\\
    \midrule
    \textsc{\small{Log-Linear Words}} && 76.97 & 63.37 & 69.51 && 77.02 & 63.55 & 69.64\\
    \textsc{\small{Wsabie Embedding}} && \textbf{78.33} & \textbf{64.51} & \textbf{70.75} && \textbf{78.33} & \textbf{64.53} & \textbf{70.76}\\
    \bottomrule
  \end{tabular}
  \caption[Full structure prediction results for FrameNet development data]{
    Full structure prediction results for FrameNet development data. This
    reports frame and argument identification performance jointly. We skip
    \textsc{\small{Log-Linear Embedding}} because it underperforms all other
    models by a large margin.}
  \label{tab:framenetfullres:dev}
\end{table}

\begin{table}[tp]
  \centering
  \begin{tabular}{@{}l@{}c@{}x{0.7in}@{}x{0.7in}@{}x{0.7in}@{}c@{}x{0.7in}@{}x{0.7in}@{}x{0.7in}@{}}
    \toprule
    Model & \phantom{ab} & \multicolumn{3}{c}{\textsc{\small{Semafor Lexicon}}} &
    \phantom{ab} & \multicolumn{3}{c}{\textsc{\small{Full Lexicon}}}\\
    \cmidrule{3-5}
    \cmidrule{7-9}
    && Precision & Recall & $F_1$ && Precision & Recall & $F_1$\\
    \midrule
    Das~et~al.~supervised\nocite{Das:2014} && 67.81 & 60.68 & 64.05 \\
    Das~et~al.~best\nocite{Das:2014} && 68.33 & 61.14 & 64.54 \\
    \midrule
    \textsc{\small{Log-Linear Words}}    && 71.21 & 63.37 & 67.06 && 73.31 & 65.20 & 69.01\\
    \textsc{\small{Wsabie Embedding}}    && \textbf{73.00} & \textbf{64.87} & \textbf{68.69} && \textbf{74.29} & \textbf{66.02} & \textbf{69.91}\\
    \bottomrule
  \end{tabular}
  \caption[Full structure prediction results for FrameNet test data]{
    Full structure prediction results for FrameNet test data. We compare to the
    prior state of the art \cite{Das:2014}.}
  \label{tab:framenetfullres}
\end{table}

Tables~\ref{tab:framenetfidres:dev} and \ref{tab:framenetfidres} present
accuracy results on frame identification. We present results on all predicates,
ambiguous predicates seen in the lexicon or the training data, and rare
ambiguous predicates that appear $\le11$ times in the training data. The
\textsc{\small{Wsabie Embedding}} model performs significantly better than the
\textsc{\small{Log-Linear Words}} baseline, while \textsc{\small{Log-Linear
    Embedding}} underperforms in every metric.

For the \textsc{\small{Semafor Lexicon}} setup, we compare with the state of the
art from \newcite{Das:2014}, who used a semi-supervised learning method to
improve upon a supervised latent-variable log-linear model. We outperform their
system on every metric, including the unseen predicates setting. When removing
the artificial restrictions on our lexicon---introduced for a fair comparison
with \newcite{Das:2014}---the absolute accuracy numbers of the
\textsc{\small{Wsabie Embedding}} model increase further to $88.73\%$.

As discussed previously, we next evaluate our model on the full frame-semantic
parsing task, by combining the frame identification model with an argument
identification model (as described in Appendix \ref{appendix:fs:argid}). We evaluate this
task adhering to the SemEval 2007 shared task evaluation setup. Again, the
\textsc{\small{Wsabie Embedding}} outperforms the previously published best
results, setting a new state of the art.
The results on the development data are in Table~\ref{tab:framenetfullres:dev},
with the actual results on the test data in Table~\ref{tab:framenetfullres}.

\begin{table}[tp]
  \centering
  \renewcommand{\arraystretch}{1.1}
  \begin{tabular}{@{}l@{}c@{}c@{}c@{ }c@{}}
    \toprule
    Model & \phantom{acdc} & All & Ambiguous & Rare \\
    \midrule
    \textsc{\small{Log-Linear Words}}    && 94.21 & 90.54 & 93.33 \\
    \textsc{\small{Log-Linear Embedding}} && 93.81 & 89.86 & \bf 93.73 \\
    \textsc{\small{Wsabie Embedding}} && \bf{94.79} & \bf{91.52} & 92.55 \\
    \bottomrule
  \end{tabular}
  \caption[Frame identification results on PropBank development data]{
    Frame identification accuracy results on the PropBank development data.
    Results are structured as in Table~\ref{tab:framenetfidres:dev}.
    }\label{tab:propbankfidres:dev}
\end{table}
\begin{table}[tp]
  \centering
  \renewcommand{\arraystretch}{1.1}
  \begin{tabular}{@{}l@{}c@{}c@{}c@{ }c@{}}
    \toprule
    Model & \phantom{acdc} & All & Ambiguous & Rare \\
    \midrule
    \textsc{\small{Log-Linear Words}}    && \bf{94.74} & \bf{92.07} & \bf{91.32} \\
    \textsc{\small{Log-Linear Embedding}} && 94.04 & 90.95 & 90.97 \\
    \textsc{\small{Wsabie Embedding}} && 94.56 & 91.82 & 90.62 \\
    \bottomrule
  \end{tabular}
  \caption[Frame identification results on PropBank test data]{
    Frame identification accuracy results on the PropBank test data. Results are
    structured as in Table~\ref{tab:framenetfidres:dev}.}\label{tab:propbankfidres}
\end{table}
\begin{table}[tp]
  \centering
  \renewcommand{\arraystretch}{1.1}
  \begin{tabular}{@{}l@{}c@{}ccc@{}}
    \toprule
    Model & \phantom{acdc} & P & R & $F_1$ \\
    \midrule
    \textsc{\small{Log-Linear Words}}    && 80.02 & 75.58 & 77.74 \\
    \textsc{\small{Wsabie Embedding}}     && \bf{80.06} & \bf{75.74} & \bf{77.84} \\
    \bottomrule
  \end{tabular}
  \caption[Full structure prediction results on PropBank development data]{
    Full frame-structure prediction results on the PropBank development
    data. This is a metric that takes into account frames and arguments
    together. See \S\ref{sec:fs:results:propbank} for more details.}
  \label{tab:propbankfullframe:dev}
\end{table}
\begin{table}[tp]
  \centering
  \renewcommand{\arraystretch}{1.1}
  \begin{tabular}{@{}l@{}c@{}ccc@{}}
    \toprule
    Model & \phantom{acdc} & P & R & $F_1$ \\
    \midrule
    \textsc{\small{Log-Linear Words}}    && \bf{81.55} & 77.83 & \bf{79.65} \\
    \textsc{\small{Wsabie Embedding}}     && 81.32 & \bf{77.97} & 79.61 \\
    \bottomrule
  \end{tabular}
  \caption[Full structure prediction results on PropBank test data]{
    Full frame-structure prediction results on the PropBank test data.}
  \label{tab:propbankfullframe}
\end{table}
\begin{table}[tp]
  \centering
  \renewcommand{\arraystretch}{1.1}
  \begin{tabular}{@{}l@{}c@{}ccc@{}}
    \toprule
    Model          & \phantom{acdc} & P & R & $F_1$ \\
    \midrule
    \textsc{\small{Log-Linear Words}}    && \bf{77.29} & \bf{71.50} & \bf{74.28} \\
    \textsc{\small{Wsabie Embedding}}     && 77.13 & 71.32 & 74.11 \\
    \bottomrule
  \end{tabular}
  \caption[CoNLL 2005 argument evaluation results on PropBank development data]{
    Argument only evaluation (semantic role labeling metrics) using the CoNLL
    2005 shared task evaluation script \protect\cite{Carreras:2005} on the
    PropBank development data.
    \label{tab:propbanksrl:dev}
    }
\end{table}
\begin{table}[tp]
  \centering
  \renewcommand{\arraystretch}{1.1}
  \begin{tabular}{@{}l@{}c@{}ccc@{}}
    \toprule
    Model          & \phantom{acdc} & P & R & $F_1$ \\
    \midrule
    Punyakanok~et~al. \em{Collins}\nocite{Punyakanok:2008} && 75.92  & 71.45 & 73.62\\
    Punyakanok~et~al. \em{Charniak}\nocite{Punyakanok:2008} && 77.09 & 75.51 & 76.29\\
    Punyakanok~et~al. \em{Combined}\nocite{Punyakanok:2008} && 80.53 & 76.94 & 78.69\\
    \midrule
    \textsc{\small{Log-Linear Words}}    && \bf{79.47} & \bf{75.11} & \bf{77.23} \\
    \textsc{\small{Wsabie Embedding}} && 79.36 & 75.04 & 77.14 \\
    \bottomrule
  \end{tabular}
  \caption[CoNLL 2005 argument evaluation results on PropBank test data]{
    Argument only evaluation (semantic role labeling metrics) using the CoNLL
    2005 shared task evaluation script \protect\cite{Carreras:2005} on the
    PropBank test data. Results from \protect\newcite{Punyakanok:2008} are taken
    from Table~11 of that paper.
    \label{tab:propbanksrl}
    }
\end{table}

\subsection{PropBank Results}\label{sec:fs:results:propbank}

Tables~\ref{tab:propbankfidres:dev} and \ref{tab:propbankfidres} show frame
identification results on the PropBank development- and test-data, respectively.
On the development set, our best model performs with the highest accuracy on all
and ambiguous predicates, but performs worse on rare ambiguous predicates. On
the test set, the \textsc{\small{Log-Linear Words}} baseline performs best by a
very narrow margin. We analyse this in the discussion section
\S\ref{sec:fs:discussion}.

Next, following the FrameNet-structure set of experiments, we provide results on
the full frame-semantic parsing task in Tables~\ref{tab:propbankfullframe:dev}
and \ref{tab:propbankfullframe}. The results follow the same trend as in the
frame identification task.

Finally, in Tables~\ref{tab:propbanksrl:dev} and \ref{tab:propbanksrl}, we
present SRL results that measure the argument performance
only, irrespective of the choice of frame. This task, for which we use the
evaluation script from CoNLL 2005 \cite{Carreras:2005}, allows us to compare the
PropBank results to the state-of-the-art results on that frame-semantic
formalism.

We note that with a better frame identification model, our performance on SRL
improves in general. Here too, the embedding model barely misses the performance
of the best baseline, but we are at par and sometimes better than the single
parser setting of a state-of-the-art SRL system \cite{Punyakanok:2008}. Note
that the \textit{Combined} results refer to a system which uses
the combination of two syntactic parsers as input.

\section{Discussion} \label{sec:fs:discussion}

With the models and experiments in this chapter we wanted to establish whether
distributed representations are a suitable mechanism for addressing semantic
tasks. The experimental evaluation of our \textsc{\small{Wsabie Embedding}}
model strongly supports this hypothesis. In this section we will
analyse the experimental evaluation in greater detail, before drawing more
general conclusions as to the role of distributed representations for semantics
in \S\ref{sec:fs:conclusion}, the concluding section of this chapter.

Let us first consider the FrameNet experiments. Here, the \textsc{\small{Wsabie
    Embedding}} model strongly outperforms all baselines as well as the prior
state of the art on all metrics. Setting a new state of the art in its own
right already strongly suggests that our distributional approach is suitable for
this task; however, in order to determine this more clearly, we further isolated
the performance of the distributed approach from all other aspects of the model.

When comparing the \wsMod model to our two baselines, the \logEmb and the
\logWord models, we discover a number of interesting results. First, we notice
that \wsMod performs better than the \logWord model, which could be seen as the
discrete counterpart to our distributed model. This result, combined with the
fact that the \logWord model still outperforms the prior state of the art
\cite{Das:2014} supports the hypothesis of this thesis, namely that distributed
representations can bring an advantage to semantically challenging tasks. We
believe this performance gain stems from the fact that the \wsMod model allows
examples with different labels and confusion sets to share information, as all
labels live in a single space and as all examples are mapped into this space
using a single projection matrix.

The second result is more interesting yet: While the \wsMod model outperforms
the \logWord model, that model in turn performs consistently better than the
\logEmb model. This could be due to a number of reasons. One particular
advantage of the \wsMod approach over the \logEmb model is that the former
learns a transformation which maps all instances and labels into a single joint
space, while the latter learns to classify for each label independently.
Therefore the \logEmb model is less able to share information across examples
and labels, and consequently more likely to suffer from sparsity-related
effects.

Why the \logWord model outperforms the \logEmb model, is a more difficult
question to answer. One likely explanation is related to the above
analysis: the failure of the log-linear approaches to share information across
multiple labels and confusion sets may be more pronounced in the embedded case,
where the input representations will have significantly higher dimensionalities
than in the \logWord case.
An important realisation here is that while distributed input representations
can be highly advantageous (as demonstrated with \wsMod), they are not
guaranteed to bring a benefit, particularly when applied in a fairly brute-force
fashion as we have with the \logEmb model.

On the PropBank data, we see that the \logWord baseline has roughly the same
performance as our model on most metrics: slightly better on the test data and
slightly worse on the development data. This can partially be explained with the
significantly larger training set size for PropBank, making features based on
words more useful.  This cannot be the only explanation, however, since the
\logEmb baseline underperforms on the FrameNet data sets as well.

Another important distinction between PropBank and FrameNet is that the latter
formalism shares frames between multiple lexical units. We see this particularly
when looking at the ``Rare'' column in Table~\ref{tab:propbankfidres}.
\textsc{\small{Wsabie Embedding}} performs poorly in this setting while
\textsc{\small{Log-Linear Embedding}} performs well.
Part of the explanation has to do with the particulars of \wsabie training.
Recall that the \textsc{\small{Wsabie Embedding}} model needs to estimate the
label location in $\R^m$ for each frame.  In other words, it must estimate 512
parameters based on at most 10 training examples.  However, since the input
representation is shared across all frames, every other training example from
all the lexical units affects the optimal estimate, since they all modify the
joint parameter matrix $M$.
By contrast, in the log-linear models each label has its own set of parameters,
and they interact only via the normalization constant.  The
\textsc{\small{Log-Linear Words}} model does not have this entanglement, but
cannot share information between words.  For PropBank, these drawbacks and
benefits balance out and we see similar performance for
\textsc{\small{Log-Linear Words}} and \textsc{\small{Log-Linear Embedding}}.  In
the FrameNet setting, estimating the label embedding is not as much of a
problem because even if a lexical unit is rare, the potential frames can be
frequent.  For example, we might have seen the \fname{Sending} frame many times,
even though \emph{telex}.\textsc{V} is a rare lexical unit.

\section{Summary}\label{sec:fs:conclusion}

In this chapter we have presented a simple model that outperforms
the previous state of the art on FrameNet-style frame-semantic parsing, and
performs on par with one of the best single-parser systems on PropBank
semantic role labelling.

Importantly, this model utilises distributed semantic representations as its
input. Unlike the prior state of the art, \newcite{Das:2014}, our model does not
rely on heuristics to construct a similarity graph and leverage WordNet; hence,
in principle it is generalisable to varying domains, and to other languages.
This again highlights the extreme usefulness of distributed representations---as
previously pointed out in Chapter \ref{chapter:distrib}---for solving a wide
variety of tasks without relying on task-specific supervised data or
annotations.

As for the question asked at the outset of this chapter, our results clearly
indicate the efficacy of distributed representations in conveying semantic
information and thus being able to solve semantic tasks.
The relative performance of the two models relying on distributed
representations---with \wsMod performing very strongly and setting the new state
of the art on FrameNet and \logEmb performing poorly on most metrics---further
lead us to the conclusion that the performance of models relying on distributed
representations strongly depends on a number of factors beyond the quality of
the input representations alone.

The analysis of sparsity-related issues allowed us to highlight some of these
factors.
In the FrameNet case, where there was less training data in absolute terms, but
more data sharing possible between representations, the \wsabie based model
performed best, whereas in the PropBank case, sparsity became more an issue for
exactly that model (which needed to learn representations in $\R^m$ for a large
  number of labels) than for the \logWord approach, as also reflected by the
results on that task.
This was highlighted especially by the experiments on rare lexical units. On the
FrameNet data---with the possibility of full information sharing---the relative
rarity of a lexical unit did not strongly affect accuracy. This result gives
further credence to the hypothesis that we can use distributed representations
to encode and share semantics, assuming that the strong performance on rare
words was caused by semantically motivated information sharing from more
frequent lexical units of the same label during training. That this is in case
so, is supported by the results on the PropBank rare data. Here, no information
is shared between labels, and consequently the \logWord model outperforms the
\wsMod setting.

In combination, these results indicate two important factors to consider when
using distributed representations as inputs in an NLP task. First, sparsity
still remains an issue and needs to be considered. Second, the key benefit of
distributed representations lies in their ability to share information. In order
to successfully make use of distributed representations, models are required
that use their inputs in a manner that maximises the possibility for information
sharing and minimises the impact of sparsity.

Another key difference between the log-linear and the \wsabie approach
is that the latter learns a transformation of its input into a smaller space.
This could be seen as a form of composition of individual representations into a
shared, dense representation of a complex object (see Figure
  \ref{fig:fs:embeddings} for that interpretation), where this smaller-space
representation benefits from increased information sharing across variables.
While the complex object in this case is a frame instance represented by
a single vector, it is feasible to consider generalizing such methods for
representing other complex linguistic structures by a composed distributed
representation. We will study such systems in the remainder of this thesis.

\part{Compositional Semantics}\label{part:comp}
\chapter{Compositional Distributed Representations}\label{chapter:compositional}

\begin{chapterabstract}
  This chapter surveys prior work on compositional semantics and gives an
  overview of key principles and methods in that field. We motivate
  compositional semantics by demonstrating the limitations of distributional
  (collocational) representations beyond the word level. Having established the
  need for composition in distributed semantics, we attempt to categorise prior
  work in this area along two axes. We introduce the distinction between
  distributional and distributed representations and finally introduce a number
  of standard techniques for representation learning in recursive and recurrent
  systems which are essential tools for all models of compositional distributed
  semantics.
\end{chapterabstract}

\section{Introduction}\label{sec:comp:intro}

So far in this thesis, we have covered distributed models of semantics at the
word level. However, for a number of important problems, semantic
representations of individual words do not suffice, but instead a semantic
representation of a larger structure---e.g. a phrase or a sentence---is
required. As elaborated in \S\ref{sec:fs:conclusion}, the \wsMod model of
Chapter \ref{chapter:frame-semantic} was a borderline case, as the model
learned to predict labels given a structured input consisting of multiple words,
with those words however being disconnected.
In other cases, for instance when wanting to establish the sentiment or the
veracity of a sentence, such an approach is likely to be insufficient and
instead a complete representation of the full sentence is more likely to lead to
success.
Therefore, during the past few years, research has shifted from using
distributional methods for modelling the semantics of words to using them for
modelling the semantics of larger linguistic units such as phrases or entire
sentences.

Self-evidently, sparsity prevents the learning of such higher representations
using the same collocational methods as applied to the word level. This can be
illustrated by considering corpus statistics. For instance, when considering the
statistics of the Europarl Corpus v7 \cite{Koehn:2005} in Table
\ref{tab:comp:stat}, the problem with learning distributed representations at
the sentence-level becomes apparent. While out of $304,786$ words, $58,552$
appear ten or more times, this is only the case for $579$ sentences, roughly
$0.0003\%$ of the total number of unique sentences found in that corpus. Hence,
in order to learn meaningful statistics at the sentence level, exponentially
more training data is required compared to the word level.

\begin{table}\centering
  \begin{tabular}{@{}lrr@{}}
    \toprule
    Statistic & \multicolumn{2}{c}{Frequency} \\ \cmidrule{2-3}
              & Sentences & Words \\ \midrule
    Token count & 1,920,209 & 47,889,787 \\
    Type count & 1,860,118 & 304,786 \\
    Types with freq. $\geq10$ & 597 & 58,552 \\
    \bottomrule
  \end{tabular}
  \caption[Word and sentence statistics for Europarl v7]{Word and sentence statistics for the English portion of the
    German-English parallel section of Europarl v7
    \protect\cite{Koehn:2005}.}
    \label{tab:comp:stat}
\end{table}

Most literature instead focuses on learning composition functions that represent
the semantics of a larger structure as a function of the representations of its
parts. Simple algebraic composition functions have been shown to suffice for
tasks such as judging bi-gram semantic similarity \cite{Mitchell:2008}. More
complex models, learning distributed representations for sentences or documents,
have proved useful in tasks such as sentiment analysis
\cite{Socher:2011,Hermann:2013:ACL}, relational similarity \cite{Turney:2012} or
dialogue analysis \cite{Kalchbrenner:2013}.

In this chapter, we first discuss the theoretical foundations of semantic
compositionality and their implications for modelling distributed compositional
models (\S\ref{sec:comp:theory}), before surveying a number of architectures for
such composition functions in \S\ref{sec:comp:architectures}. Next, we provide
an overview over commonly used objective functions and error signals employed
for learning both composition functions and semantic representations
(\S\ref{sec:comp:signals}) and finally, \S\ref{sec:comp:appl} provides a survey
of applications for such models in the literature.

\section{Theoretical Foundations}\label{sec:comp:theory}

\begin{aquote}{G. Frege, 1892}
  The meaning of an utterance is a function of the meanings of its parts and
  their composition rules.
\end{aquote}

Since Frege stated his `Principle of Semantic Compositionality' in 1892
researchers have pondered both how the meaning of a complex expression is
determined by the meanings of its parts, and how those parts are combined
\cite{Frege:1892,Pelletier:1994}. Over a hundred years on, the choice of
representational unit for this process of compositional semantics, and how these
units combine, remain open questions.

Within Natural Language Processing, work on semantic representations can roughly
be grouped into two distinct categories. On one side, Montagovian approaches are
concerned with symbolic entities composed into logical representations of
meaning. Here, \textit{meaning} is typically associated with sentences, which
are expressed as logical formulae over abstract types. On the other hand,
distributional approaches such as discussed in this thesis, learn continuous
representations of individual words. Frequently, the distributional approach to
semantics stops at precisely that point, namely the word level. However, there
has been some work on applying distributional (and by extension distributed)
approaches to semantic composition, in order to learn such distributed
representations for higher linguistic units beyond the word level.

In this thesis we focus on semantic composition on the basis of distributed
representations. With that we mean the derivation of distributed representations
of phrases or other grammatical units based on models of semantic composition
and representations for the parts which these larger structures are composed of.
Other aspects of compositionality, such as questions of compositionality versus
lexicality of compounds---`\textit{gravy} sauce' vs. `\textit{gravy train}'
\cite[\textit{inter
    alia}]{Bannard:2003,Biemann:2011,Hermann:2012:Ranking})---are not considered
    here.

Following Frege's principle, we assume that the meaning of a sentence is
composed of the meanings of the individual words or phrases it contains, and
likewise, that the meaning of a phrase can be composed of the meanings of its
words. As we assume distributed representations as the basis of our work, we can
define a general composition function, following \newcite{Mitchell:2008}, as
\begin{equation}
  \mathbf{p} = f(\mathbf{u},\mathbf{v},R,K) \label{eqn:comp:full},
\end{equation}
where \textbf{p} represents the composed meaning of inputs \textbf{u} and
\textbf{v} given some background knowledge $K$ and some syntactic relation $R$
between \textbf{u} and \textbf{v}. While \newcite{Mitchell:2008} assume
\textbf{u} and \textbf{v} to be vectors representing the semantics of the
underlying words, we remove any assumptions about the shape of the semantic
representation of words and larger structures.

This formulation of semantic composition captures ``a wide class of composition
functions,'' \cite{Mitchell:2010} sharing this underlying principle of semantic
composition. Typically, a number of constraints are placed on $f$, such as
$\mathbf{u},\mathbf{v},\mathbf{p}\in \mathbb{R}^n$, which ensures that $f$ can
be applied recursively. Conventionally the background knowledge $K$ is dropped
from most composition models. In \S\ref{sec:comp:architectures} we survey the
main classes of composition models in the literature implementing such functions
$f$.

The notion of compositionality we consider here is comparable to that found in
Montagovian formal semantics \cite{Montague:1970,Montague:1974} (in spirit, at
  least). In Montague's system, grammatical analysis rules are paired with
higher-order logical interpretations, allowing the derivation of the logical
interpretation of a sentence to be obtained deterministically from its syntactic
structure. In effect, syntax guides semantic composition.

Unlike the distributed representations discussed in this thesis, such frameworks
typically represent meaning symbolically (hence symbolic logic), which does not
make them well suited for quantitative modelling. Consider the semantic space as
represented by a symbolic logic. Here, similarity is well defined---two atoms
are similar if equal and likewise for compound types---but this definition has
no tolerance for noisy representation or vague semantics.
In fact, semantic similarity in symbolic logic collapses onto type
identity, which is undesirable in instances where semantic ambiguity needs to be
accounted for.  We expand on the discussion of these types of grammars in
Chapter \ref{chapter:syntax}, where we empirically evaluate the role of syntax
in learning models for compositional distributed semantics.

The basic composition formula (Equation \ref{eqn:comp:full}) places no
constraints on the form of its inputs. While we generally assume $u$ and $v$ to
be vectors, they could just as well be symbols (see previous paragraph) or more
complex representations such as tensors or tuples of several types of
representations.

Tensor spaces in particular have attracted a lot of attention
as a possible format for representing compositional semantics. Tensor
representations have the nice property that predicate-argument completion can be
posited as tensor contraction with tensors of suitable rank \cite[\textit{inter
    alia}]{Clark:2008,Coecke:2010,Grefenstette:2013a}. We will
describe this idea in more detail in \S\ref{sec:comp:arch:lex}.

\section{Architectures}\label{sec:comp:architectures}

\begin{figure}
  \centering
  \begin{tikzpicture}[yscale=3.6,xscale=6]
    \draw [] (0,0) rectangle (1,1);
    \draw [] (1,1) rectangle (2,2);
    \draw [] (0,0) rectangle (2,2);
    \fill[blue,path fading=south, fading angle=225, opacity=0.8] (0,0) rectangle (2,2);
    \node [align=center] at (-0.3,1.5) {Algebraic\\Composition};
    \node [align=center] at (-0.3,0.5) {Lexical\\Function\\Models};
    \node [align=center] at (0.5,-0.3) {Distributional\\Approaches};
    \node [align=center] at (1.5,-0.3) {Distributed\\Approaches};
    \draw[->,thick] (-0.06,2) -- (-0.06,0);
    \draw[->,thick] (0,-0.1) -- (2,-0.1);
    \node [align=center] at (1.5,1.5) {RAE\\\cite{Socher:2011}};
    \node [align=center] at (1.5,0.5) {CCAE\\(This thesis)};
    \node [align=center] at (0.5,1.5) {Additive BoW\\\cite{Mitchell:2008}};
    \node [align=center] at (0.5,0.5) {Adj-Noun\\\cite{Baroni:2010}};
  \end{tikzpicture}
  \caption[Main types of models for semantic composition]{Schematic overview of
  the main types of models for semantic composition. The Y-Axis differentiates
  between algebraic models and lexical models, the X-Axis between models relying
  on distributional information and more abstract models, with exemplary models
  in the boxes.}\label{fig:comp:chart}
\end{figure}
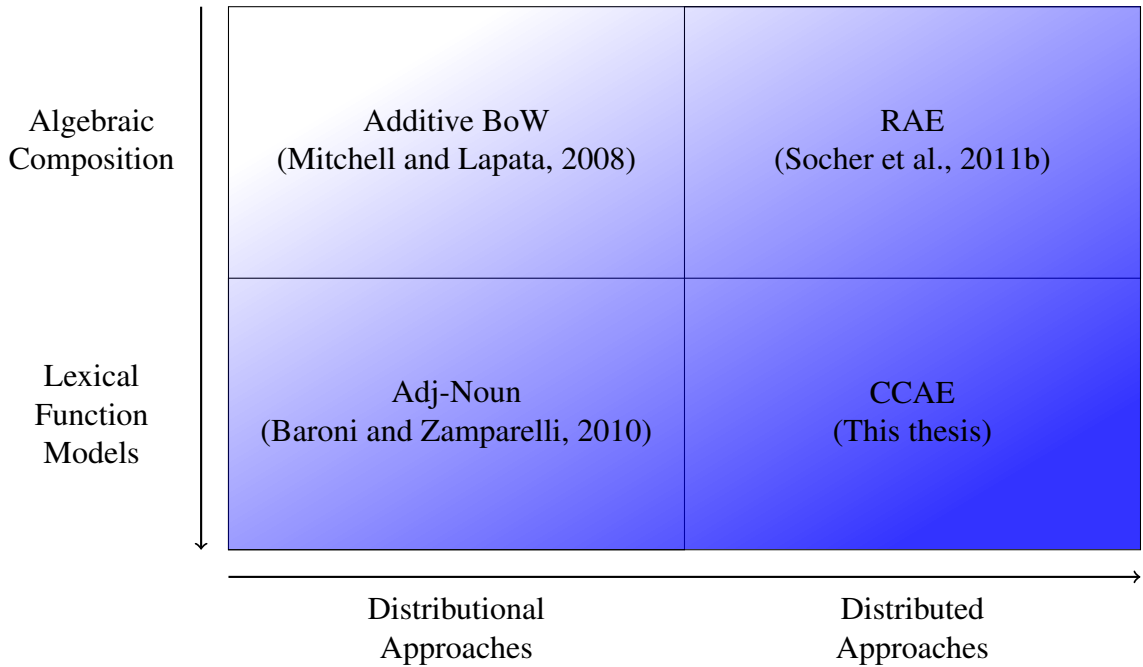

Various architectures exist for composing distributed representations. We
provide a survey over a number of relevant approaches as proposed in the
literature. We use the function defined above (Eq. \ref{eqn:comp:full}) as the
basis of this survey.

We attempt to structure prior work in this field along two dimensions (see
Figure \ref{fig:comp:chart}). First, we differentiate between the type of
composition function, which can roughly be divided between algebraic composition
models and lexical function models. On the second axis, we differentiate between
distributional and distributed approaches, where distributional approaches rely
on collocational or other distributional information for model learning and word
representations.

\subsection{Algebraic Composition}\label{sec:comp:arch:ac}

Algebraic composition has been proposed in the context of distributional
semantics as a simple mechanism for obtaining distributed representations for
composed words. Algebraic composition functions simplify Equation
\ref{eqn:comp:full} by removing the background knowledge $K$, and frequently
also the relational information $R$, effectively reducing the function signature
to
\begin{align}
  \mathbf{p} = f(\mathbf{u},\mathbf{v})
\end{align}

Addition can be seen as the simplest form of algebraic composition.
Assume representations for words \textit{red}, \textit{apple} as
$\vec{v}_{\text{red}}$, $\vec{v}_{\text{apple}}$. Then, under the additive
model, we would represent \textit{red apple} as
\begin{align*}
  \vec{v}_{\text{red apple}} = \vec{v}_{\text{red}} + \vec{v}_{\text{apple}}
\end{align*}
Addition (or averaging) has successfully been applied to some problems such as
essay grading \cite{Landauer:1997} or selectional preference
\cite{Kintsch:2001}. However, addition as a composition function makes no use of
syntactic information or for that matter word ordering. Therefore, two
sentences with matching words but different meaning would be represented by the
same vector. An example for this problem is given in \newcite{Mitchell:2010},
who compare the two sentences in Figure \ref{enum:bottles}.  While the two
sentences share the same set of words, clearly their meaning is entirely
different.

\begin{figure}
\begin{enumerate}
\item It was not the sales manager who hit the bottle that day, but the office
  worker with the serious drinking problem.
\item That day the office manager, who was drinking, hit the problem sales
  worker with the bottle, but it was not serious.
\end{enumerate}
\caption[Sentences with shared vocabulary but different meaning]{Two sentences
with a shared set of words but entirely different meaning. Example from
\protect\newcite{Mitchell:2010}.}\label{enum:bottles}
\end{figure}

Other algebraic functions that have been explored in the literature include
pointwise multiplication, weighted addition, dilation, and tensor products
(outer product or Kronecker product) --- see Table \ref{tab:comp:alg}
\cite[\textit{inter
    alia}]{Mitchell:2008,Mitchell:2009,Guevara:2010,Zanzotto:2010}.
Related research also includes holographic reduced representations, random
indexing and convolution products \cite{Widdows:2004,Widdows:2008}.
An extensive
survey of these algebraic operators applied to semantic composition can be found
in \newcite{Mitchell:2010} together with comparative results across a number of
tasks.
\begin{table}\centering
  \begin{tabular}{@{}ll@{}}
    \toprule
    Operation & Function \\ \midrule
    Additive & $\mathbf{p} = \alpha \mathbf{u} + \beta\mathbf{v}$ \\
    General Additive & $\mathbf{p} = A\mathbf{u} + B\mathbf{v}$ \\
    Multiplicative & $\mathbf{p} = \mathbf{u} \odot~\mathbf{v}$ \\
    Tensor & $\mathbf{p} = \mathbf{u} \otimes~\mathbf{v}$ \\
    Dilation & $\mathbf{p} = \left(\mathbf{u} \cdot \mathbf{u}\right) \mathbf{v}
    + \left( \lambda - 1 \right) \left( \mathbf{u} \cdot \mathbf{v}\right)
    \mathbf{u}$ \\
    \bottomrule
  \end{tabular}
  \caption[Algebraic operators frequently used for semantic
    composition]{Comparison of some algebraic operators frequently used in the
    literature for composing distributed semantic
    representations. $u$ and $v$ are inputs, $p$ the composed representation.
    $\alpha$, $\beta$ are scalar weights, $A$ and $B$ matrices, and $\lambda$ is
    a scalar stretching factor.}\label{tab:comp:alg}
\end{table}
When applied recursively, such models are frequently augmented with a
non-linearity such as a hyperbolic tangent or sigmoid function. Among other
effects, this ensures that word ordering is accounted for in the model,
resulting in different representations depending on the composition order. One
popular model that uses such a composition function are recursive autoencoders
as presented in \newcite{Socher:2011}.

\paragraph{Limitations}
While again some success could be shown for specific tasks such as similarity
ratings of adjective-noun, noun-noun and verb-object pairs, all of these
algebraic operators face similar limitations as the additive model.
Many algebraic operators ignore word ordering, thereby essentially turning these
composition models into bag of word models. Even those algebraic operators that
take word order into account, such as tensor products or dilation, ignore all
syntactic information. This limitation contradicts evidence from the literature
which suggests that such syntactic information is not only useful but necessary
for semantic composition. Chapter \ref{chapter:syntax} investigates this
question by empirically studying the role of syntax in compositional semantics.

Beyond this, some algebraic models suffer from tendencies caused by their
underlying algebraic functions. For instance, multiplicative models will tend to
zero with increasing sentence length, while additive models will tend to the
average word representation in a corpus. Chapter 2 in
\newcite{Grefenstette:2013b} provides a comprehensive analysis of these effects.

\subsection{Lexical Function Models}\label{sec:comp:arch:lex}

Lexical function models could be seen as an extension of the algebraic
composition methods described in \S\ref{sec:comp:arch:ac}. While algebraic
composition is largely parameter-free, lexical function models rely on some
parametrization in their composition function. Further, making use of syntactic
information, composition can be governed by parse trees or similar syntactic
structures.

\paragraph{Augmented Algebraic Models}
One approach to incorporate lexical information in the composition function is
to augment algebraic composition models with syntactic information. For
instance, \newcite{Zanzotto:2010} propose an enhanced additive model (BAM-SP)
for estimating selectional preference that incorporates the syntactic relation
between two words into its composition function
\begin{align}
  \mathbf{p} = \mathbf{u} \odot R_v(r)
\end{align}
with $\mathbf{u}$ and $\mathbf{v}$ reversed if $\mathbf{v}$ is the semantic head
of the two words, and $R_v(r)$ returns a selectional preference vector for word
$\mathbf{v}$ and relation $r$. Thereby, the composition function is now a
function not only of the two input words, but also of the specific relation $r$
connecting the two words. They demonstrated that this addition of syntactic
information substantially improved results on a number of tasks compared with
standard algebraic composition models. This model is essentially a parametrised
version of a simple multiplicative model, where syntactic relations govern one
of the vector choices.

Guevara proposes a partial least squares regression model to learn a similar
composition function, which extends the general additive model from Table
\ref{tab:comp:alg} \cite{Guevara:2010,Guevara:2011}. This approach, as that of
\newcite{Zanzotto:2010}, parametrizes its composition function on the syntactic
relationship between the two words that are to be composed (here limited to
  adjective-noun and verb-noun pairs).

A key difference of these models compared to the algebraic composition models
above is that the composition parameters are learned by considering the
distributional representations of both unigrams and bigrams, thereby enabling
the regression learning. The idea behind this is to directly learn bigram
representations using some distributional representation learning method. These
bigram representations are then used together with unigram representations to
learn the model parameters.

\paragraph{Syntactic Functions}

A second group of lexical function models separates the functional information
from the lexical elements more strictly. An example for this, also described in
\newcite{Zanzotto:2010}, is the general additive composition model
\begin{equation}
  \mathbf{p} = A\mathbf{u} + B\mathbf{v},
\end{equation}
where $A, B$ are square matrices ``capturing the relation $R$ and the
background knowledge $K$'' of Equation \ref{eqn:comp:full}. This means that the
word representations $\mathbf{u}$ and $\mathbf{v}$ are learnt independently of
their syntactic function.

Applied to sets of adjective-noun, noun-noun and verb-noun sequences, the
authors demonstrated that this type of composition function can be trained to
successfully separate synonymous pairs from other data using a cosine similarity
measure to determine semantic similarity.

\paragraph{Implicit Lexical Functions}

A number of variations and extensions to syntactic function models as briefly
outlined above are easily imaginable. An alternative approach would be to encode
the composition function directly in the semantic representation of a linguistic
unit.
Technically the BAM-SP model by \newcite{Zanzotto:2010} could be counted as such
an approach, with linguistic units being represented by sets of vectors, and the
composition step choosing the appropriate vector depending on the composition
context.

A more principled approach is the idea to use representations that fully
incorporate the composition function, removing the need for additional variables
such as in the fully additive approach.
Building on pregroup grammars
\cite{Clark:2008,Coecke:2010} introduced in \S\ref{sec:comp:theory},
tensor-based composition methods have been proposed as one such alternative
mechanism for capturing semantic interaction between words.

The original idea of using tensor products for semantic composition is
attributed to \newcite{Smolensky:1990}, who proposed such a model in cognitive
science\footnote{The following description is based on that in
  \newcite{Clark:2008}.}:

\begin{quote}
Meaning is represented by a set of structure roles $\{r_i\}$, which may be
occupied by fillers $\mathbf{f}_i$. Then, $s$ is a set of constituents, each a
filler/role binding $\mathbf{f}_i/r_i$.
\end{quote}

Linguistic structures such as parse trees or predicate-argument systems can be
thought of as such sets of constituents. In order to represent a set of
constituents, \newcite{Smolensky:2006} then propose a tensor product
formulation:
\begin{equation}
  \mathbf{s} = \sum_{i}\mathbf{f}_i\otimes r_i
\end{equation}

The proposal in \newcite{Clark:2008} extends this concept under the idea of
unifying distributional and symbolic models of meaning. As we described earlier,
symbolic approaches to semantics have predominately concerned themselves with
the composition of atomic types into larger logical structures, whereas
distributed approaches frequently failed to go beyond the word level. Picking up
the idea of using tensor products for composition, they propose to use
distributional representations at the word level in combination with some form
of symbolic representation at the sentence level, giving sentence
representations such as (from \newcite{Clark:2008}):
\begin{align*}
  drinks\otimes subj\otimes John\otimes obj \otimes(beer\otimes adj\otimes
    strong)\otimes adv\otimes quickly
\end{align*}

A more recent line of research has focused on tensor contraction as opposed to
tensor products for semantic composition
\cite[\textit{inter
    alia}]{Coecke:2010,Baroni:2010,Grefenstette:2011,Grefenstette:2013b}. An
important consideration here is to condition the shape of each representation
by its role, which allows the removal of the role representations $\{r_i\}$.
One example for this approach is \newcite{Baroni:2010}, which focuses on the
case of adjective-noun composition. Here, adjectives are represented as matrices
and nouns as (distributional) vectors. Thus, adjective-noun composition becomes
a simple matrix-vector product with adjectives being a linear map over nouns.
It is easy to see how this approach naturally extends to other linguistic units,
such as intransitive (matrices) and transitive verbs (order 3 tensors).
\subsection{Recursive Composition}\label{sec:comp:vsm}

Most of the work described so far is concerned with either theoretical
approaches to semantic composition or with semantic composition of small units
such as adjective-noun pairs or noun compounds, where distributional methods can
be combined with regression analysis to learn a composition
function.\footnote{Of course, the implicit lexical models are an exception
  here as these models provide a natural mechanism for composition of arbitrary
  depth given sufficiently complex implicit representations.}

In terms of the matrix of composition models in \S\ref{sec:comp:architectures}
(Figure \ref{fig:comp:chart}), these models primarily cover the left half of the
model space.
As the curse of dimensionality precludes the application of the distributional
hypothesis to linguistic structures significantly beyond the size of unigrams,
alternative approaches are required for capturing compositional semantics of
larger structures such as sentences. The implicit approaches combining
distributional and symbolic forms of composition have similar limitations: While
technically only unigram representations need to be learned, it remains unclear
how such higher-order representations (e.g. order-3 tensors for transitive
  verbs) can be learned. Further, for more complex linguistic units which can
take several arguments, the order of the tensor to be learned would become
infeasibly large.

For the sake of argument, if one used CCG types to condition the shape of
representations as has been proposed \cite{Clark:2008}, a simple word such as
\textit{for} could require an order 5 tensor with its type signature
$((S\backslash NP)\backslash(S\backslash NP))/NP$. Even with a modest 50 dimensional
embedding, this would result in a 312,500,000 dimensional representation for
this word alone, calling into question the practicability of such an approach.
There exist a number of approximate learning algorithms and other
simplifications such as low-rank tensor approximations that can help to address
this issue \egcite{Grefenstette:2013,Baroni:2010}, and hence tensor-based models
remain a very plausible space to be explored.
For the purposes of this thesis, we focus on approaches to compositional
semantics that do not rely on implicit functional representations, but instead
use composition functions in a more conventional matrix-vector setup. With
implicit functional representation we mean the idea of encoding all functional
information within the word representation. This is opposed to approaches that
use a combination of word representations and word-independent functional
features for semantic composition.

Now the key question is how to solve the learning problem without making use of
distributional representations for composed entities. Resolving this issue would
allow us to take any of the composition functions proposed so far and to
recursively apply these to learn distributed representations of sentences and
other higher order linguistic structures.
Indeed, many of the models introduced in the previous two sections have trivial
recursive extensions, meaning that they could be used to compute a
representation not only for a pair of words, but recursively for a phrase, a
sentence or a document and so on.

The generalised composition models presented in this section can be viewed as
the \textit{distributed} extension of compositional semantics. The composition
models discussed in the previous two subsections all rely on
\textit{distributional} embeddings as the basis of their composition.
Generalised vector space composition models can still be instantiated using
distributional embeddings, but no longer rely on them. Instead, word
representations and composition functions are jointly learned given some
objective function.

Here, we describe structures for producing distributed representations at the
sentence or other higher level. The remainder of this background chapter will
then outline a number of objective functions that can be used for training such
systems.

\begin{figure}[t]
\begin{center}
\includegraphics[scale=0.5]{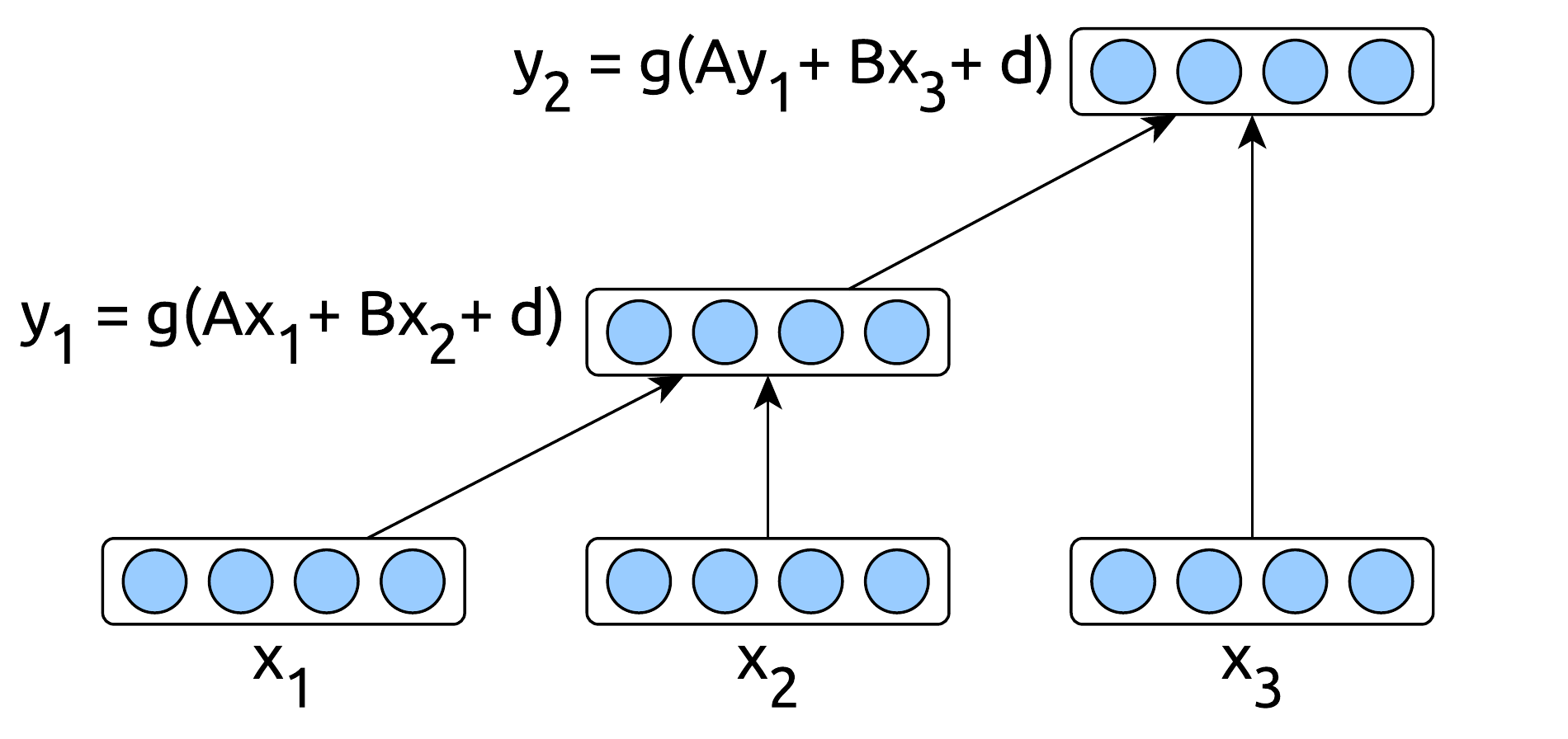}
\caption[A simple three-input recursive neural network]{A simple three-input recursive neural network. The three input words
  $x_i$ are recursively composed into vector $y_2$.}\label{fig:comp:recnn}
\end{center}
\end{figure}

\paragraph{Recursive Neural Networks} (RecNN, henceforth) are one popular
mechanism for combining fully additive models recursively. A single composition
step can be described with the following template:

\begin{align}
  p = g(Ax + By + d)\label{eqn:comp:recnn}
\end{align}
where $x, y$ are input vectors, $A, B$ weight matrices and $d$ a bias term.
$g$ represents an element-wise activation function such as a sigmoid or
hyperbolic tangent function. This non-linearity is required to turn this
composition step into a single layer neural network.
Subsequently, by enforcing $A$ and $B$ to be quadratic, we ensure that $p,x,y,d
\in \mathbb{R}^n$ for some $n$. This of course allows the composition function
to be applied recursively along some binary-branching structure such as
binarised parse trees or simple left-to-right trees (see Figure
  \ref{fig:comp:recnn}).
Recursive Neural Networks in that fashion were first applied to language in
\newcite{Socher:2011}, who used such a model for predicting sentiment (see
  \S\ref{sec:comp:appl}).

\paragraph{Convolutional Neural Networks}
\begin{figure}[t]
\begin{subfigure}[b]{0.5\linewidth}
  \centering
\includegraphics[scale=0.25]{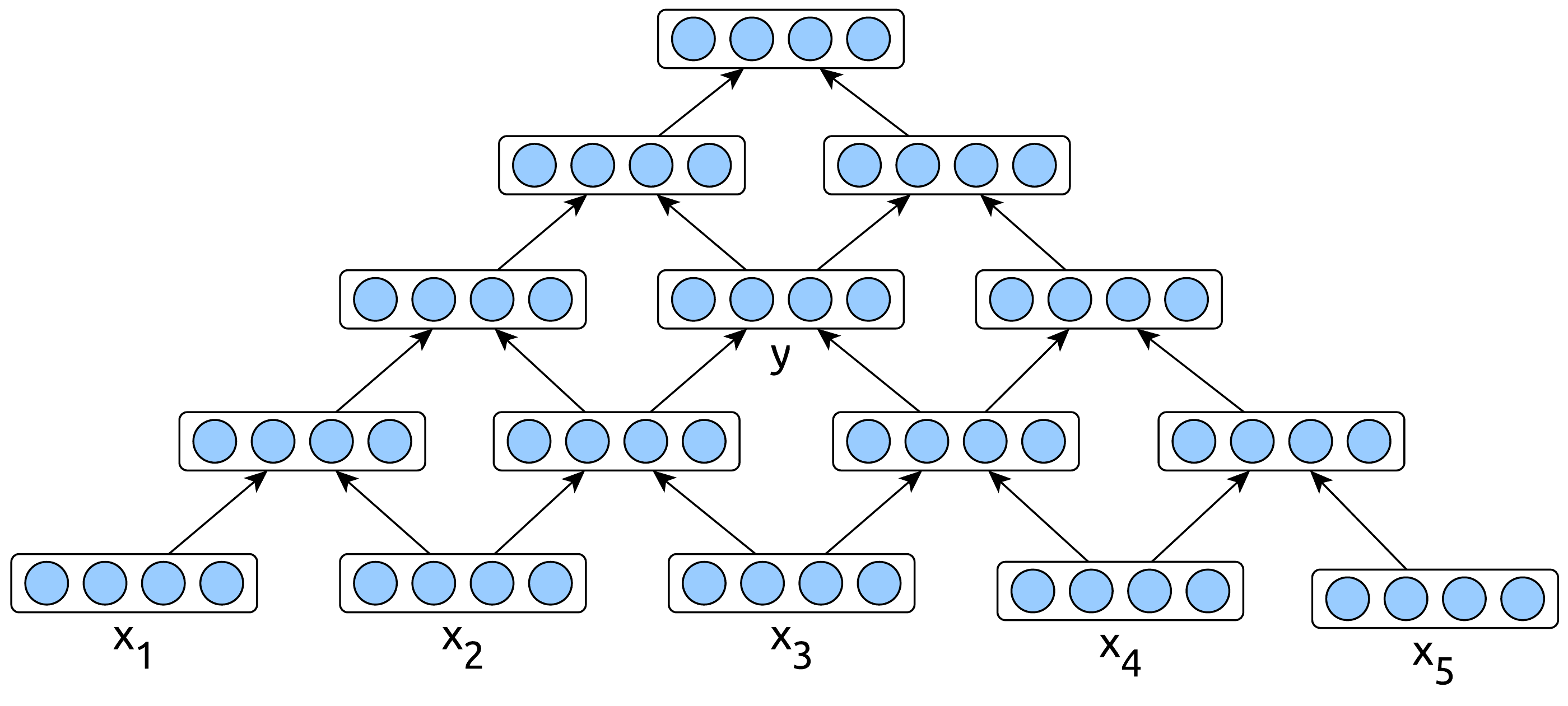}
\caption[A ConvNN with a receptive width of 3]{A ConvNN with a receptive width of 2.}
\end{subfigure}
\begin{subfigure}[b]{0.5\linewidth}
  \centering
\includegraphics[scale=0.25]{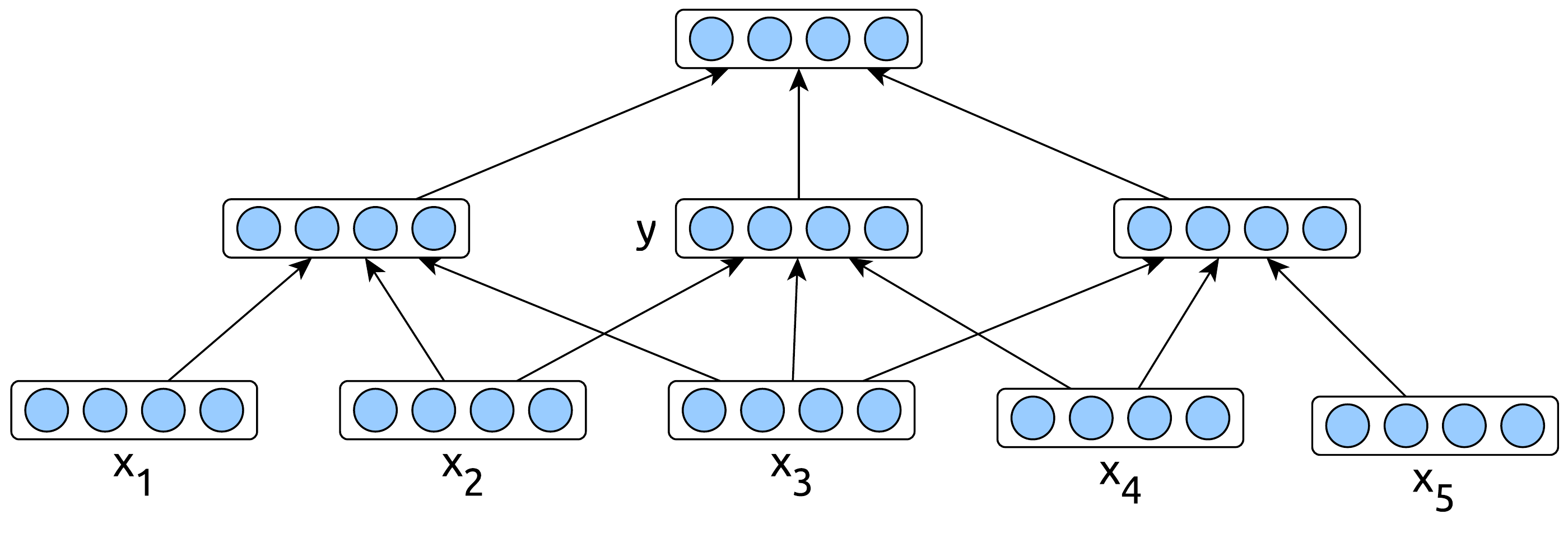}
\caption[A ConvNN with a receptive width of 3]{A ConvNN with a receptive width of 3.}
\end{subfigure}
\caption[Two ConvNN with different receptive widths]{Two ConvNN with differing receptive widths. The cell marked $y$ in each
  network captures inputs $x_2\dots x_4$. As can be seen, the connectivity of
  the network impacts the degree to which the visual field of cells in each
  layer increases.}\label{fig:comp:convnn}
\end{figure}
(ConvNN, henceforth) provide a different fashion for composing multiple vectors,
which is based on research on visual cortices in animals \cite{Hubel:1968}. The
underlying idea is to connect cells in such a fashion that while each outer cell
may cover only a small region of the visual field, cells on subsequent layers
indirectly cover increasingly large regions of the visual field. This enables
animals, humans and machines to exploit local correlations.
Convolutional Neural Networks have proved a popular tool in visual recognition
tasks such as document recognition and OCR \cite{LeCun:1998b}.

However, the underlying idea can easily be ported to natural language processing
when replacing the visual input with text (see Figure \ref{fig:comp:convnn} for
  a schematic depiction of two such ConvNN). See \egcite{Kalchbrenner:2014} for
the application of ConvNN to modelling language. Unlike a RecNN, a ConvNN does
not require any syntactic or other parse. The composition step of a ConvNN with
receptive width 2 is equivalent to that of a RecNN (Equation
  \ref{eqn:comp:recnn}), with additional weight matrices required for larger
receptive widths.

ConvNN can be structured in a number of fashions. Typically, weight matrices and
biases are either shared across all composition steps or alternatively across
all composition steps within a given layer of the network. This weight sharing
is important, as it allows the network to detect features irrespective of their
position in the input.

\paragraph{Matrix-Vector Neural Networks} (MV-RNN, henceforth) are an extension
of RecNN or ConvNN with a more complex input representation \cite{Socher:2012a}.
Here, instead of representing words by a vector, words and other units are
represented by a tuple consisting of a vector $a\in\mathbb{R}^n$ and a matrix
$A\in\mathbb{R}^{n\times n}$. Composition is achieved using the template:
\begin{align}
  p &= g(VAb + WBa + d)\\
  P &= MA + NB
\end{align}
where $V,W,M,N$ are matrices in $\mathbb{R}^{n\times n}$ and $d,g$ as before.

This extends the matrix-vector multiplication approaches presented earlier
\cite{Mitchell:2010,Zanzotto:2010} and \cite{Baroni:2010} by adding a
non-linearity as well as a mechanism for propagating weight matrices in a
recursive setup.

\section{Learning Signals}\label{sec:comp:signals}

The models discussed in \S\ref{sec:comp:vsm} provide an architecture for
producing composed representations given a lexicon of input (word) embeddings,
as well as a set of model parameters $\theta$, while being agnostic about how
these parameters and input embeddings are learned.
In this section, we survey the most popular types of learning signals used in
conjunction with compositional vector space models.

\subsection{Autoencoders}\label{sec:comp:sig:ae}

\begin{figure}[t]
  \begin{center}
    \includegraphics[scale=0.5]{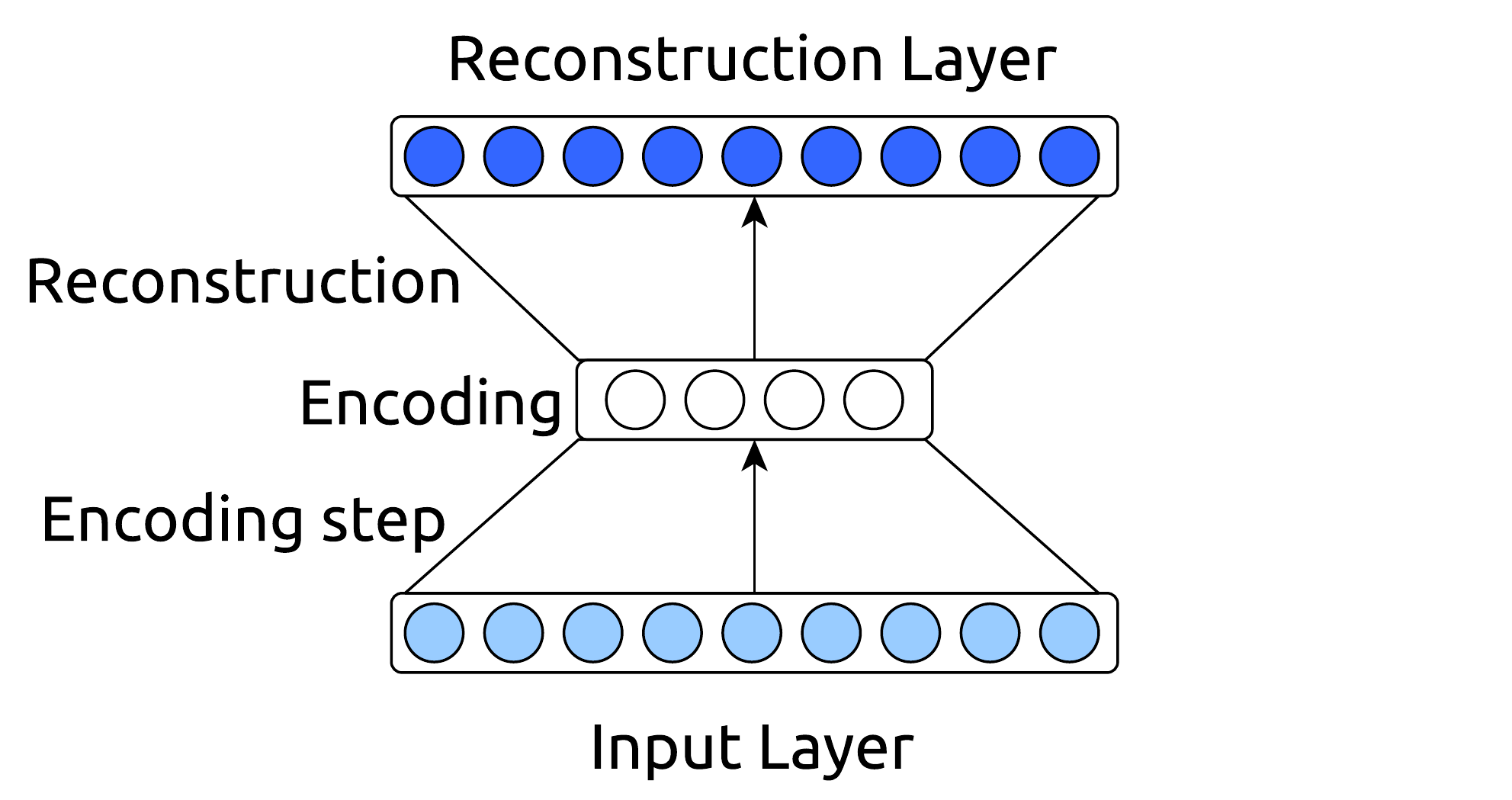}
    \caption[A simple three-layer autoencoder]{A simple three-layer autoencoder. The input represented by the vector
      at the bottom is being encoded in a smaller vector (middle), from which it is
      then reconstructed (top) into the same dimensionality as the original input
      vector.}\label{fig:comp:ae}
  \end{center}
\end{figure}

Autoencoders provide a popular method for learning embeddings from unsupervised
data or in conjunction with other, supervised signals.
They are a useful tool to compress information. One can think of an
autoencoder as a funnel through which information has to pass (see Figure
  \ref{fig:comp:ae}). By forcing the autoencoder to reconstruct an input given
only the reduced amount of information available inside the funnel it serves as
a compression tool, representing high-dimensional objects in a lower-dimensional
space.

Typically a given autoencoder, that is the functions for encoding and
reconstructing data, is used on multiple inputs.
By optimizing the two functions to minimize the difference between all inputs
and their respective reconstructions, this autoencoder will effectively discover
some hidden structures within the data that can be exploited to represent it
more efficiently.

As a simple example, assume input vectors $x_i \in \mathbb{R}^n, i \in (0..N)$,
weight matrices $W^{enc} \in \mathbb{R}^{(m\times n)} ,W^{rec} \in
\mathbb{R}^{(n\times m)}$, biases $b^{enc} \in \mathbb{R}^m$, $b^{rec} \in
\mathbb{R}^n$ and some non-linearity $g$. The encoding matrix and bias are used
to create an encoding $e_i$ from $x_i$:
\begin{align}
  e_i = g\left(f^{enc}(x_i) = W^{enc}x_i + b^{enc}\right)
\end{align}
Subsequently $e \in \mathbb{R}^m$ is used to reconstruct $x$ as $x'$ using the
reconstruction matrix and bias:
\begin{align}
  x'_i = \left(f^{rec}(e_i) = W^{rec}e_i + b^{rec}\right)
\end{align}
$\theta = (W^{enc},W^{rec},b^{enc},b^{rec})$ can then be learned by minimizing
the error function describing the difference between $x'$ and $x$:
\begin{align}
  E = \frac{1}{2} \sum_{i}^{N} \left\| x'_i - x_i \right\|^2
\end{align}
Now, if $m < n$, this will intuitively lead to $e_i$ encoding a latent structure
contained in $x_i$ and shared across all $x_j, j\in (0..N)$, with $\theta$
encoding and decoding to and from that hidden structure.

\subsubsection{Recursive Autoencoders}\label{sec:comp:sig:rae}

In \S\ref{sec:comp:sig:ae}, we introduced autoencoders as a simple mechanism to
extract latent structure by enforcing data to learn a joint compression and
reconstruction regime.

It is possible to apply multiple autoencoders on top of each other, creating a
deep autoencoder \cite{Bengio:2007,Hinton:2006}. For such a multi-layered model
to learn anything beyond what a single layer could learn, a non-linear
transformation $g$ needs to be applied at each layer. Usually, a variant of the
logistic ($\sigma$) or hyperbolic tangent ($tanh$) function is used for $g$
\cite{LeCun:1998}.
\begin{align}
  &f^{enc}(x_i) = g\left(W^{enc}x_i + b^{enc}\right) \\ \nonumber
  &f^{rec}(e_i) = g\left(W^{rec}e_i + b^{rec}\right)
\end{align}

\begin{figure}[t]
  \begin{center}
    \includegraphics[scale=0.4]{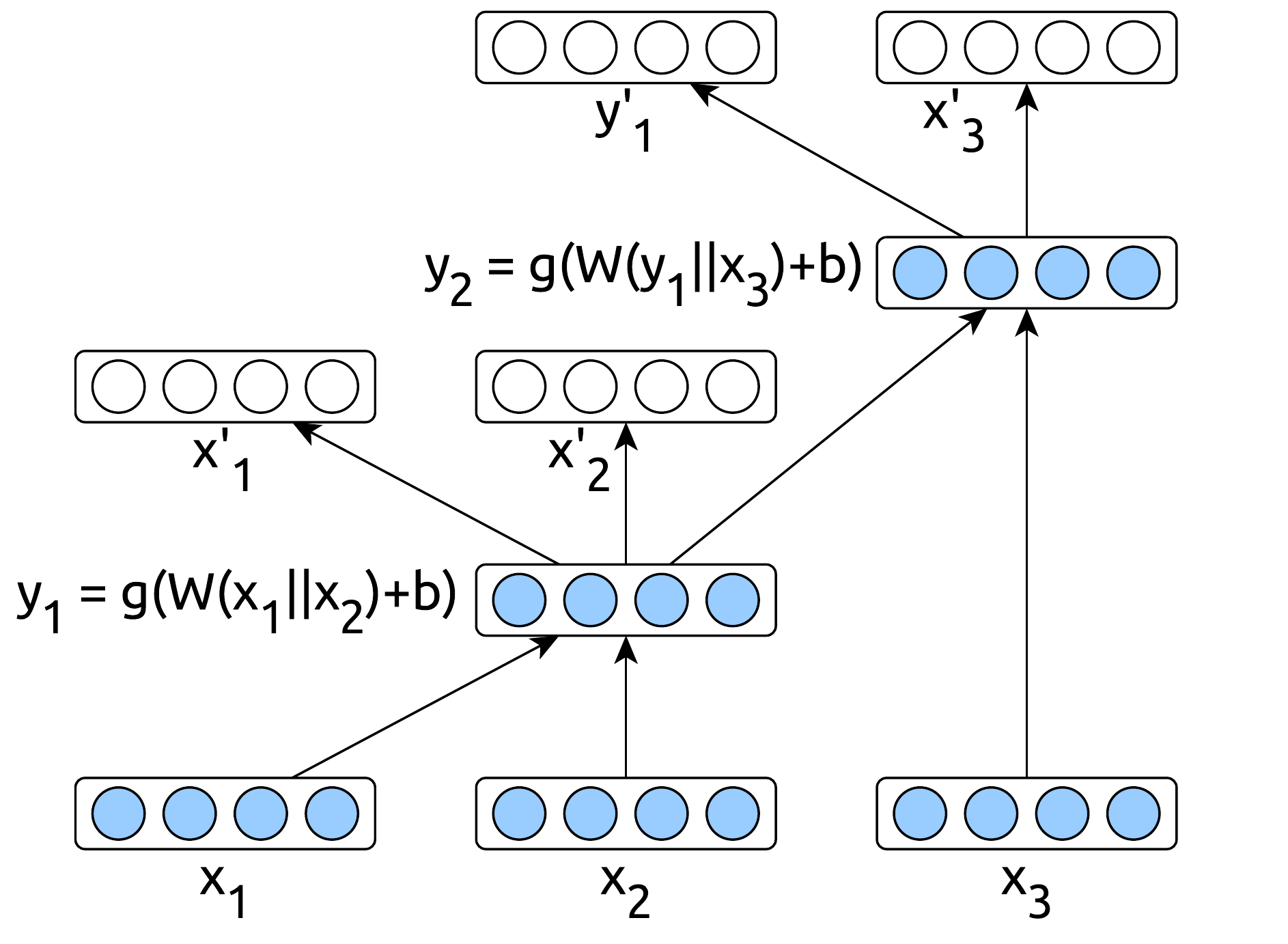}
    \caption[Recursive Autoencoder with three inputs]{RAE with three inputs. Vectors with filled (blue) circles represent
      input and hidden units; blanks (white) denote reconstruction
      layers.}\label{fig:comp:rae}
  \end{center}
\end{figure}

Furthermore, autoencoders can easily be used as a composition function by
concatenating two input vectors, such that:
\begin{align}
  e = f(x_1,x_2) &= g\left(W(x_1\|x_2)+b\right) \\
  (x'_1\|x'_2) &= g\left(W'e + b'\right) \nonumber
\end{align}
Extending this idea, recursive autoencoders (RAE) allow the modelling of data of
variable size.  By setting the $n = 2m$, it is possible to recursively combine a
structure into an autoencoder tree. See Figure \ref{fig:comp:rae} for an example,
where $x_1,x_2,x_3$ are recursively encoded into $y_2$. Thus, an RAE is
equivalent to a RecNN (\S\ref{sec:comp:vsm}) with an additional decoding
layer for each composition step.

The recursive application of autoencoders was first introduced in
\newcite{Pollack:1990}, whose recursive auto-associative memories learn vector
representations over pre-specified recursive data structures. More recently
this idea was extended and applied to dynamic structures \cite{Socher:2011}.
These types of models have become increasingly prominent since developments
within the field of Deep Learning have made the training of such hierarchical
structures more effective and tractable \cite{LeCun:1998,Hinton:2006a}.

Intuitively the top layer of an RAE will encode aspects of the information
stored in all of the input vectors.  Previously, RAEs have successfully been
applied to a number of tasks including sentiment analysis, paraphrase detection,
relation extraction and 3D object identification
\cite{Blacoe:2012,Socher:2011,Socher:2012a}.

Depending on the layout and structure of an RAE setup, one has to be careful to
prevent degeneration. If an objective function jointly optimises the error at
each layer of an autoencoder, e.g. in a convolution setup, a degenerate solution
is to minimise the magnitude of all encoding and reconstruction weights, as this would cause a
large error on the first level but minimal error on all subsequent levels. There
are various strategies for avoiding such behaviour, such as normalising
embeddings at each level in the autoencoder. Another strategy is to use
unfolding autoencoders which only consider the reconstruction error of original
input elements. We introduce unfolding autoencoders in the next subsection.

\subsubsection{Unfolding Autoencoders}

\begin{figure}[t]
  \begin{center}
    \includegraphics[scale=0.4]{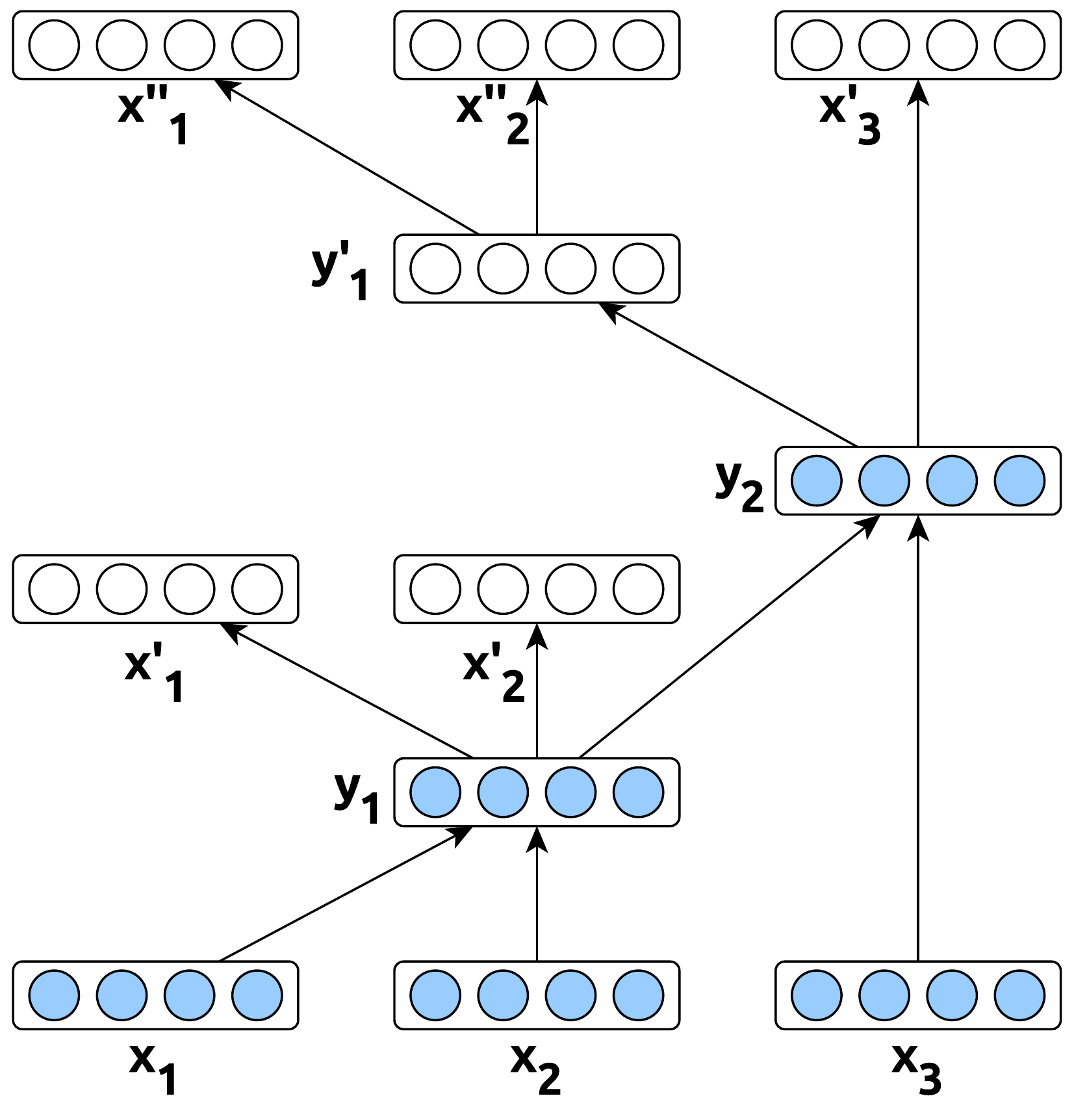}
    \caption[Unfolding autoencoder with three inputs]{Unfolding autoencoder with three inputs. Vectors with filled (blue)
      circles represent input and hidden units; blanks (white) denote intermediate
      and final reconstruction layers.}\label{fig:comp:rae:unf}
  \end{center}
\end{figure}

Unfolding autoencoders are an extension of recursive autoencoders, where each
reconstruction step is applied recursively until an original input is
reconstructed \cite{Socher:2011a}. Figure \ref{fig:comp:rae:unf} demonstrates
this: where a standard RAE would stop the reconstruction step from $y_2$ at
$y_1'$, the unfolding autoenocoder continues its recursive reconstruction by
unfolding $y_1'$ into $x_1''$ and $x_2''$, thereby reconstructing data from the
input layer.

Unfolding autoencoders have a number of nice properties. Particularly, the
unfolding prevents an RAE from degenerating, as standard recursive autoencoders
are
incentivised to learn small weights for all internal layers, thereby shrinking
the overall reconstruction error. As the unfolding autoencoder measures its
error function always by comparing with input weights, this strategy becomes
void. Of course, if input weights are updated as part of the learning process,
the issue of degenerating all weights to zero still persists, and needs to be
addressed separately.

\subsubsection{Denoising Autoencoders}

Another variant of autoencoders are denoising autoencoders. Here, the idea is to
force the hidden layer to discover structures in the data by making it
reconstruct the input data from a corrupted version of itself, with the idea
that this process will improve the robustness of the discovered representations
\cite{Vincent:2008}.

Denoising autoencoders are effectively a stochastic version of regular
autoencoders, which corrupt the input layer before feeding it into the encoding
function. A number of possibilities exist for this corruption function. In
\newcite{Vincent:2008}, it consists of randomly setting a number of inputs to
zero. Alternative corruption processes could introduce a random amount of
Gaussian noise for instance.

The denoising autoencoder thus works as follows (using the RecNN notation):
\begin{align}
  \tilde{x}_1 &\sim q(\tilde{x}_1|x_1) \\
  \tilde{x}_2 &\sim q(\tilde{x}_2|x_2) \nonumber\\
  e &= g(A\tilde{x}_1 + B\tilde{x}_2 + d) \nonumber\\
  r_1 &= g(A_1'e + d_1') \nonumber\\
  r_2 &= g(A_2'e + d_2') \nonumber\\
  E &= \|x_1 - r_1\|^2 + \|x_2 - r_2\|^2 \nonumber
\end{align}

Here, $\tilde{x}_1$ is the corrupted version of $x_1$, which is used for the
encoding $e$ and reconstruction step $r_1$. Finally, however, the error is
calculated by comparing the uncorrupted input with the reconstructed vector.

The noise added by corrupting inputs acts in a similar to a regularization term
in this setup. For a detailed account of denoising autoencoders and their
underlying mathematics, please refer to \newcite{Vincent:2008} and
\newcite{Bengio:2009}. Also of interest is \newcite{Wager:2013}, who provide an
account of how denoising acts as a regularisation mechanism in neural network.

\subsection{Classification}\label{sec:comp:sig:class}

While autoencoders simply learn to compress data efficiently by making use of
latent structures within the data, frequently we want to train models for a
specific task. For instance for a paraphrase detection task we are
interested in a robust semantic representation that allows us to identify
phrases with similar meaning, and hence autoencoders may be a suitable mechanism
for learning such a system. On the other hand, if we want to train a system to
discover sentiment in text, it might make more sense to make use of some
training data to help the composition function on the aspects of the
representation that are relevant for sentiment.

For this purpose, classification systems and errors can be used in conjunction
with a semantic composition process. A supervised classification layer
can be applied to the root node of a recursive compositional model, or indeed to
any tree nodes of the model. A simple binary classifier could be a sigmoid layer
such as this:
\begin{align}
  \text{predictor}(l{=}1|v,\theta) = \text{sigmoid}(W_{\textit{label}}v+b_{\textit{label}})
\end{align}
where $v$ is the vector to be classified, and $W_{\textit{label}}$,
$b_{\textit{label}}$ are the classifier weights and bias, respectively. $\theta$
here represents the set of all model data.

Given a label and encoding pair $(l,e)$, the classifier error can be formulated
as:
\begin{align}
  &E_{label}(l,e,\theta) = \frac{1}{2} \left\|l - o\right\|^2 \\
  &\text{with}~o = \text{sigmoid}(W_{\textit{label}}e+b_{\textit{label}})
\end{align}

Set in the context of a recursive composition model, such a classifier (and the
  corresponding classification error learning signal) can be applied to any
number of nodes in the composition tree. If we assume that only the top node in
a tree will be used for classification, we get the following objective function.
Assume a corpus of input label pairs $(\mathbf{x}, l) \in D$ where $\mathbf{x}
= \left<x_0, x_1, \dots, x_j\right>$ and $j = |\mathbf{x}|$. Further assume a recursive
composition function such that $e = \mathit{recfunc}(\mathbf{x})$ returns the
representation on top of the composition tree. Then we have
\begin{align}
  J &= \frac{1}{|D|} \sum_{(\mathbf{x},l)}^{D} E(\mathbf{x},l,\theta) +
  \frac{\lambda}{2}||\theta||^2 \\
  E(\mathbf{x},l,\theta) &= \frac{1}{2} \left\|l - \mathit{recfunc}(\mathbf{x})\right\|^2
\end{align}
where the second term of the objective function is a standard $L_2$
regularization parameter.

\subsection{Bilingual Constraints}

Beyond the autoencoding and classification signals discussed in the previous
subsections, we can also use specificities in the data for training composition
models. For instance, when training a model on data where we know that two
inputs capture the same information, we can make use of that knowledge to
force the model to assign similar representations to both inputs. This is
particularly of interest when dealing with inputs that are different on the
surface level but equivalent on a higher (i.e. composed) level. In computer
vision an example for this would be multiple photos of the same object, in
compositional semantics two sentences with the same underlying meaning.

When considering semantic representations an obvious source for such data are
paraphrases and multilingual corpora. In particular, multilingual aligned data
has nice theoretical properties that could allow models to learn representations
further removed from mono-lingual surface realisations.
As this is part of the novel work presented in this thesis, we defer this
discussion to the separate chapter on multilingual signals (Chapter
  \ref{chapter:multilingual}).

\subsection{Signal Combination}

A nice property of the space of recursive functions we are investigating here is
that the various learning signals proposed in this chapter are easily
combinable. For instance, in the following chapter we will combine recursive
autoencoders with a classification signal. Due to the type of gradient learning
employed in training such models, additional signals can be added at a
cost linear in the number of training instances.

\section{Learning}

Here we discuss strategies for efficiently learning word embeddings and model
parameters for compositional vector space models. First, we investigate how to
calculate gradients in recursive and dynamic structures. Second, we provide an
overview of gradient descent algorithms used for updating weights.

\subsection{Signal Propagation in Recursive
  Structures}\label{sec:comp:learn:prop}

Given some error signal $E$, we need to compute the gradients $\partial
E/\partial w$ for all $w\in\theta$. In neural networks and other \textit{deep},
recursive or multi-layered setups, backpropagation can be used to efficiently
calculate these gradients \cite{Goller:1996}.

Backpropagation works by calculating the partial derivatives with respect to all
temporary and internal nodes in a network, and by using the chain rule to
efficiently calculate partial derivatives of nodes lower in the network based on
those of nodes higher up (closer to the output / error signal). For a simple
example, assume a two layer neural network with input $i$, intermediate
encoding $e$ and output $o$ all in $\mathbb{R}^n$:
\begin{align}
  z &= W^a i + b^a \label{eqn:comp:learn:fwd}\\
  e &= \sigma(z) \nonumber\\
  k &= W^b e + b^b \nonumber\\
  o &= \sigma(k) \nonumber\\
  E &= \|o - y\|^2 \nonumber,
\end{align}
where $W^a \in \mathbb{R}^{m\times n}$ and $W^b \in \mathbb{R}^{m\times n}$
are encoding and reconstruction matrices and $b^a \in \mathbb{R}^m$ and $b^b \in
\mathbb{R}^n$ are bias vectors. Further, $\sigma$ represents the sigmoid
function, and $y$ the expected output of the network.
Using the intermediate values at $z$ and $k$, the derivatives with respect to
the error function $E$ can efficiently be calculated for all model weights and
inputs. For instance, the partial derivatives for the reconstruction weight
matrix can be calculated as in Table \ref{tab:comp:learn:backprop}.
\begin{table}[t]
  \centering
  \renewcommand{\arraystretch}{2.0}
  \everymath{\displaystyle}
  \begin{tabular}{ll}
  $\frac{\partial E}{\partial o} = (o - y)$ & \\
  $\frac{\partial o}{\partial k} = \sigma'(k) = \sigma(k) (1 - \sigma(k))$ \\
  $\frac{\partial k}{\partial e} = W^b$ &
  $\frac{\partial k}{\partial W^b} = e$ \\
  $\frac{\partial e}{\partial z} = \sigma'(z) = \sigma(z) (1 - \sigma(z))$ \\
  $\frac{\partial z}{\partial i} = W^a$ &
  $\frac{\partial z}{\partial W^a} = i$
\end{tabular}
\caption[Partial derivatives for backpropagation]{Partial derivatives for the backpropagation of gradients from Equation
  \protect\ref{eqn:comp:learn:fwd}.}\label{tab:comp:learn:backprop}
\end{table}
Using the chain rule, we get
\begin{align}
  &\frac{\partial E}{\partial W^b} = \frac{\partial E}{\partial o}
  \frac{\partial o}{\partial k} \frac{\partial k}{\partial W^b} \\[0.5em]
  &\frac{\partial E}{\partial i} = \frac{\partial E}{\partial o} \frac{\partial
  o}{\partial k} \frac{\partial k}{\partial e} \frac{\partial e}{\partial z}
\frac{\partial z}{\partial i} \label{eqn:comp:prop:chain}
\end{align}
and so forth. This allows us to avoid the repetitive calculation of intermediate
values such as $\frac{\partial o}{\partial k}$, which are shared by several
gradients relevant for the model update.

\subsubsection{Backpropagation Through Structure}\label{sec:comp:learn:bpts}

For deep, recursive networks, backpropagation through structure (BPTS) provides
an efficient mechanism for learning gradients \cite{Goller:1996}. BPTS is
essentially the extension of simple backpropagation to general structured models.

\begin{figure}[t]
\begin{center}
\includegraphics[scale=0.5]{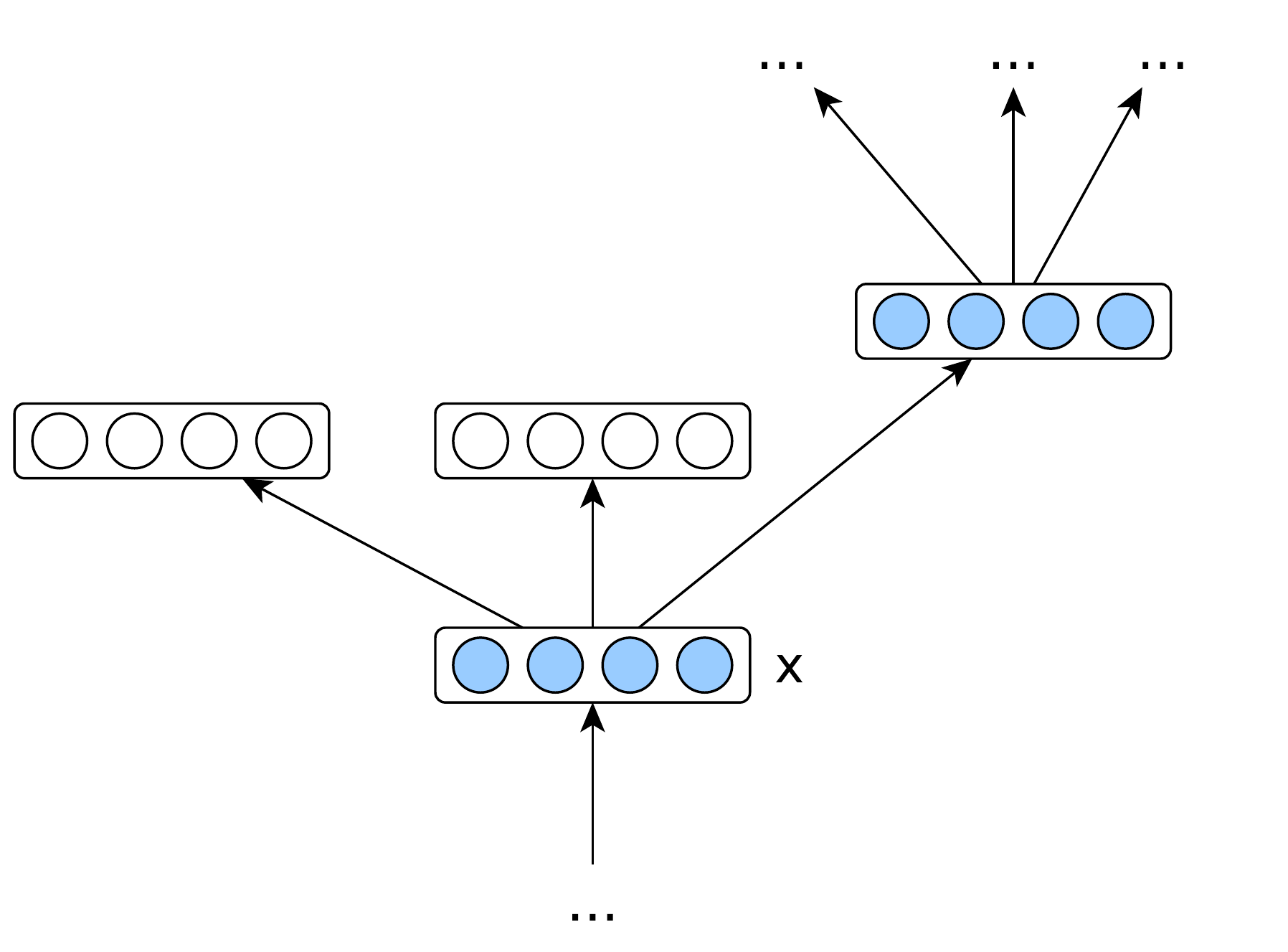}
\caption[Extract of a recursive autoencoder]{Extract of a recursive autoencoder, where $x$ represents some layer
  which is used both as an encoding for two reconstructions (left) and as an
  input for further composition (right branch).}\label{fig:comp:backprop}
\end{center}
\end{figure}

Unlike the example in \S\ref{sec:comp:learn:prop}, node $x$ receives multiple
signals from the structure, as the objective function will include terms
covering both reconstruction layers as well as propagated signals from higher up
in the model.
We know that
\begin{align}
&\frac{\partial E}{\partial x} = \sum_{\gamma\in \Gamma} \frac{\partial E}{\partial
  \gamma}\frac{\partial \gamma}{\partial x}\\
&\Gamma = \text{Successors of x} \nonumber
\end{align}
This allows us to efficiently calculate all partial derivatives with respect to
$E$.

A slight modification of the previous example (Equation
  \ref{eqn:comp:learn:fwd}) can serve to demonstrate this.  Assume an equivalent
network with a new error function
\begin{align}
  E = \|o - \hat{i}\|^2
\end{align}
where $\hat{i} = i$ (this helper variable will keep the notation clean). This is
the formulation for a simple autoencoder setup with a non-linear encoding
function, two weight matrices and bias terms.
Now, term $i$ has two successors in the network: first the
intermediate value $z$, and second the error function $E$ via $\hat{i}$. Thus,
we get
\begin{align}
  \frac{\partial E}{\partial i} &= \sum_{\gamma\in \{z,\hat{i}\}} \frac{\partial E}{\partial
  \gamma}\frac{\partial \gamma}{\partial i}\label{eqn:comp:prop:chaintwo}\\
  &= \frac{\partial E}{\partial o} \frac{\partial
  o}{\partial k} \frac{\partial k}{\partial e} \frac{\partial e}{\partial z}
\frac{\partial z}{\partial i} + \frac{\partial E}{\partial \hat{i}}
\frac{\partial \hat{i}}{\partial i} \nonumber
\end{align}

Backpropagation through structure is then the process of optimizing this
derivative calculation by computing intermediate derivatives in such an order
that no partial derivative needs to be computed repeatedly.

\subsection{Gradient Update Functions}

Using the backpropagation strategy as described above, we can represent a model
during training by a vector over its parameters $\theta_t$ and an equally sized
vector of gradients $\nabla_{\theta_t} = \nabla J(\theta_t)$ at time
$t$ given some objective function $J$. In order to minimize
$J$, a number of strategies are available to us for updating weights.

\subsubsection{Standard Gradient Descent}

A standard (or \textit{batch}) gradient descent method updates weights using a
uniformly weighted gradient subtraction
\begin{align}
  \theta_{t+1} = \theta_t - \alpha \nabla_{\theta_t}
\end{align}
where $\alpha$ is the step size or learning rate (these two terms can be used
  interchangeably).
Thus, given some differentiable objective function, all model parameters are
iteratively updated based on their gradient with respect to this error function.
For convex problems, batch gradient descent is guaranteed
to converge as long as the learning rate $\alpha$ is chosen appropriately.

For large amounts of training data, it can be prohibitively expensive and
inexpedient to perform a full batch gradient update. An alternative is to use
mini-batch or on-line stochastic gradient descent. Here, gradients are
calculated for a random subset (in on-line gradient descent a subset of size
  one) of the training data. Assume a mini-batch size of $n$ and a corpus of $m$
training examples. Then each mini-batch contains $\frac{m}{n}$ instances, and
the gradient at each step is calculated with respect to the objective function
given those instances. While this gradient is unlikely to match the
\textit{true} gradient (the objective function differentiated with respect to all
  inputs), such a strategy can lead to faster convergence as each individual
iteration will be much cheaper to calculate. The assumption for this is that
while the gradient update steps will be slightly incorrect, on average they
would still lead to the correct direction.
For convex problems, stochastic gradient descent is also almost guaranteed to
converge under the Robbins-Siegmund theorem \egcite{Bottou:2010}, provided some
mild conditions are met such as the learning rate $\alpha$ decreasing at an
appropriate rate. Unfortunately, most composition models have non-convex
objective functions, which means that typically only a local minimum will be
found.

\subsubsection{L-BFGS}

The Broyden-Fletcher-Goldfarb-Shanno algorithm (BFGS, henceforth), and
particularly the limited memory version of this algorithm (L-BFGS), provides an
alternative to the default gradient descent mechanism. BFGS is a quasi-Newtonian
method, which estimates the Hessian matrix $B_t$ (the matrix of second
  derivatives) of $J$ and uses this to determine the search direction
$\mathbf{p}_t$ of its gradient update step.
\begin{align}
  B_t \mathbf{p}_t = - \nabla J(\theta_t)
\end{align}
BFGS uses this search direction in combination with a line search method to
perform its gradient update, while simultaneously updating its estimation of the
Hessian.

While BFGS is proven to find a global optimum for convex optimization
problems, it performs well also on non-convex problems. Particularly, when used
in conjunction with Wolfe-type or Armijo-type line search, BFGS is globally
convergent for functions with Lipschitz continuous gradients \cite{Li:2000}.
This is an important result as the types of non-linearities typically employed
in the recursive frameworks described in this thesis (sigmoid and hyperbolic
  tangent functions) have Lipschitz continuous gradients, thereby guaranteeing
that BFGS will converge to some stationary point.

For more details about the properties of BFGS and various line search methods
that can be used in conjunction with BFGS or other gradient descent algorithms,
the reader is referred to Chapter 3 of \newcite{Nocedal:2006}.

\subsubsection{Adaptive Gradient Descent}

Adaptive (sub)gradient descent (AdaGrad, henceforth) is another popular gradient
descent algorithm \cite{Duchi:2011} which adjusts the learning rate on a
feature-by-feature basis. A general class of constrained optimisation update
functions for an objective function $J$ can be described by the following
function:
\begin{equation}
  \theta_{t+1} =
  \Pi_{\mathbf{\theta}}^{\mathrm{diag}(G_{\theta})^{1/2}}\left(\theta_t - \eta
    \mathrm{diag}(G_{t})^{-1/2}\nabla J(\theta_t)\right),
  \label{eqn:comp:learn:adagrad}
\end{equation}
where $G_t = \sum_{\tau=1}^t \nabla J(\theta_\tau) \nabla
J(\theta_\tau)^\top$ is the outer product of the subgradients,
$\mathrm{diag}(x)$ extracts the diagonal of a matrix $x$, and
$\Pi_\mathcal{X}^A(y) = \mathrm{argmin}_{x\in \mathcal{X}}(x{-}y)\cdot A(x{-}y)$
denotes the projection of a point $y$ onto $X$ according to $A$. $\eta$ denotes
some fixed step size.

This generalised gradient update function can be simplified. While the
projection $\Pi_X^A(y)$ is necessary for constrained problems where it is used
to project parameters onto the feasible set, we can remove this projection for
unconstrained problems. Further, if $\theta$ is represented as a vector,
equation \ref{eqn:comp:learn:adagrad} can further be reduced to
\begin{equation}
  \theta_{t+1} = \theta_t - \eta G^{-1/2}_{t}\nabla J(\theta_t),
\end{equation}
where $G_t = \sum_{\tau=1}^t \nabla J(\theta_\tau)^2$.

This algorithm is similar to second-order gradient descent, as the $G^{-1/2}_t$
term approximates the Hessian of the objective function $J$. AdaGrad has proved
a very popular algorithm for performing gradient descent. In practice it
converges much faster than L-BFGS on problems where calculating the objective
function is expensive, which makes the line search in L-BFGS costly. Further,
AdaGrad is essentially parameter-free, with the step size being controlled by
the proximal function (the Hessian approximation).  For a detailed analysis and
description of the AdaGrad algorithm, the reader is referred to
\newcite{Duchi:2011}.

\section{Applications}\label{sec:comp:appl}

The move from word representations to capturing the semantics of sentences and
other composed linguistic units has yielded models applied to a large variety of
NLP-related tasks. Here we provide a short (and by no means exhaustive) overview
of some such applications.
A number of publications targeted paraphrase detection \cite[\textit{inter
    alia}]{Mitchell:2008,Mitchell:2010,Grefenstette:2011,Blacoe:2012}, sentiment
analysis \cite{Socher:2011,Socher:2012a,Socher:2013} , and semantic relation
classification (\emph{ibid.}). Other tasks include relational similarity
\cite{Turney:2012} and discourse analysis \cite{Kalchbrenner:2013}.

Concerning the types of composition models discussed in this chapter, most
efforts so far approach the problem of modelling phrase meaning through vector
composition using linear algebraic vector operations
\cite{Mitchell:2008,Mitchell:2010,Zanzotto:2010}, matrix or tensor-based
approaches \cite[\textit{inter
    alia}]{Baroni:2010,Coecke:2010,Kartsaklis:2012,Grefenstette:2013,Clark:2013,Maillard:2014},
or through the use of recursive auto-encoding \cite{Socher:2011a} or
neural-networks \cite{Socher:2012a}.
Some alternative approaches avoid composition. For instance, \newcite{Erk:2008}
keep word vectors separate, using syntactic information from sentences to
disambiguate words in context; likewise \newcite{Turney:2012} treats the
compositional aspect of phrases and sentences as a matter of similarity measure
composition rather than vector composition.

\section{Summary}

In this chapter we have surveyed the field of compositional semantics within the
context of distributed representations. Having demonstrated in Chapter
\ref{chapter:frame-semantic} that distributed representations are useful for
solving semantically challenging tasks, we subsequently pointed out that for a
large number of such tasks representations beyond the word level would be more
beneficial.

We explored and rejected the idea of extending the distributional hypothesis
introduced in Chapter \ref{chapter:distrib} to sentence-level representations in
\S\ref{sec:comp:intro}. Following that realisation, we investigated
compositional semantics from the perspective of composing distributed
representations into higher level representations. For this we first proposed
the distinction between distributional and distributed representations and
subsequently developed a model for categorising existing and new methods for
semantic composition.

This chapter concluded with an overview of some of these models together with a
brief introduction to various important mathematical concepts and learning
algorithms that are essential for further work in this field. In the following
two chapters we will build on the analysis and background knowledge presented
here. With a view to validating the hypothesis stated in the beginning of this
thesis, Chapter \ref{chapter:syntax} attempts to evaluate the role of syntax in
compositional semantics and, given a novel model for semantic vector
composition, investigates our hypothesis that distributed representations are
also suitable for encoding semantics at the sentence-level.

Building on this, Chapter \ref{chapter:multilingual} goes one step further and
investigates representation learning with the aim to minimise the impact of
monolingual surface forms and other biases. For this, we propose a novel
algorithm for capturing semantics from multilingual corpora. Using the insights
gained in this introductory chapter as well as Chapter \ref{chapter:syntax},
this algorithm relies on semantic transfer at the sentence-level, thereby
enabling us to jointly investigate both aspects of this thesis' hypothesis.

\newcommand{\modA}{CCAE-A\xspace}
\newcommand{\modB}{CCAE-B\xspace}
\newcommand{\modC}{CCAE-C\xspace}
\newcommand{\modD}{CCAE-D\xspace}

\chapter{The Role of Syntax in Compositional Semantics}\label{chapter:syntax}

\emptyfootnote{The material in this chapter was originally presented in
  \newcite{Hermann:2013:ACL}. Aspects of the material were also presented in
  \newcite{Hermann:2013:CVSC}.}

\begin{chapterabstract}
Modelling the compositional process by which the meaning of an utterance arises
from the meaning of its parts is a fundamental task of Natural Language
Processing.  In this chapter we draw upon recent advances in the learning of
vector space representations of sentential semantics and the transparent
interface between syntax and semantics provided by Combinatory Categorial
Grammar (CCG) to introduce Combinatory Categorial Autoencoders.  This model leverages
the CCG combinatory operators to guide a non-linear transformation of meaning
within a sentence.  We use this model to learn high dimensional embeddings for
sentences and evaluate them in a range of tasks, demonstrating that the
incorporation of syntax allows a concise model to learn representations that are
both effective and general.
\end{chapterabstract}

\section{Introduction}

In this chapter we investigate compositional semantics from a syntactic
perspective. This investigation is based on Frege's principle (see
  \S\ref{sec:comp:theory}), which claims that composed meaning can be derived
from the meaning of its parts and some composition rules. Frege's principle may
be debatable from a linguistic and philosophical standpoint, but it has provided
a basis for a range of formal approaches to semantics which attempt to capture
meaning in logical models. Montague grammar \cite{Montague:1970} is a prime
example for this, building a model of composition based on lambda-calculus and
formal logic. More recent work in this field includes Combinatory Categorial
Grammar (CCG), which also places increased emphasis on syntactic coverage
\cite{Szabolcsi:1989}.

Further to Frege's principle, empirical evidence in the literature suggests that
such composition rules can benefit from syntactic information. Examples for this
include \newcite{Duffy:1989}, who investigated the effect of context for naming
target words. In particular, the authors showed that such \textit{priming}
depended on a combination of multiple preceding words rather than just the
individual preceding unigram. Subsequently, the third author of that paper
showed in \newcite{Morris:1994} that the exploitation of syntactic dependencies
for priming improved experimental results. \newcite{Morris:2002} discuss how
early report bias also conforms to syntax, further supporting this argument,
with other work reporting a loss of priming effects when scrambling words in a
sentence (see \newcite{Mitchell:2011} for a more in-depth survey of cognitive
  science research into this problem).
These early results motivate a further investigation into the role of syntax in
compositional semantics.

As discussed so far in this thesis, those searching for
the right representation for compositional semantics have recently drawn
inspiration from the success of distributional models of lexical semantics. This
approach represents single words as distributional vectors, implying that a
word's meaning is a function of the environment it appears in, be that its
syntactic role or co-occurrences with other words
\cite{Pereira:1993,Schutze:1998}.

These vectors implicitly encode semantics by
clustering words with similar environments within the vector space.  While
distributional semantics is easily applied to single words, sparsity implies
that attempts to directly extract distributional representations for larger
expressions are doomed to fail (see the discussion in \S\ref{sec:comp:intro}).
While in the past few years, attempts have been made at extending distributed
representations to semantic composition \cite[\textit{inter
    alia}]{Baroni:2010,Grefenstette:2011,Socher:2011}, these approaches make
minimal use of linguistic information beyond the word level.

Here, we attempt to bridge the gap between recent advances in machine learning
and more traditional approaches within computational linguistics.  We achieve
this goal by employing the CCG formalism to consider compositional structures at
any point in a parse tree.  CCG is attractive both for its transparent interface
between syntax and semantics, and a small but powerful set of combinatory
operators with which we can parametrise our non-linear transformations of
compositional meaning.

We present a novel class of recursive models, the Combinatory Categorial
Autoencoders (CCAE), which marry a semantic process provided by a recursive
autoencoder with the syntactic representations of the CCG formalism.  Through
this model we seek to answer two questions: Can recursive vector space models be
reconciled with a more formal notion of compositionality; and is there a role
for syntax in guiding semantics in these types of models?  CCAEs make use of CCG
combinators and types by conditioning each composition function on its
equivalent step in a CCG proof.  In terms of learning complexity and space
requirements, our models strike a balance between simpler greedy approaches
\cite{Socher:2011} and the larger recursive vector-matrix models
\cite{Socher:2012}.

We show that this combination of state of the art machine learning and an
advanced linguistic formalism translates into concise models with competitive
performance on a variety of tasks. In the experimental evaluation of our models
we show that our CCAE models match or better comparable recursive autoencoder
models.

\section{Formal Accounts of Semantic Composition}

There exist a number of formal approaches to language that provide mechanisms
for compositionality.  Generative Grammars \cite{Jackendoff:1972} treat
semantics, and thus compositionality, essentially as an extension of syntax,
with the generative (syntactic) process yielding a structure that can be
interpreted semantically.  By contrast Montague grammar achieves greater
separation between the semantic and the syntactic by using lambda calculus to
express meaning.  However, this greater separation between surface form and
meaning comes at a price in the form of reduced computability.  While this is
outside the scope of this thesis, see e.g.~\newcite{Kracht:2008} for a detailed
analysis of compositionality in these formalisms.

\subsection{Combinatory Categorial Grammar}\label{sec:syntax:ccg}

In this chapter we focus on CCG, a linguistically expressive yet computationally
efficient grammar formalism.  It uses a constituency-based structure with
complex syntactic types (categories) from which sentences can be deduced using a
small number of combinators.  CCG relies on combinatory logic (as opposed to
lambda calculus) to build its expressions.  For a detailed introduction and
analysis vis-\`{a}-vis other grammar formalisms see e.g.
\newcite{Steedman:2011}.

CCG has been described as having a transparent surface between the syntactic and
the semantic.  It is this property which makes it attractive for our purposes of
providing a conditioning structure for semantic operators.
A second benefit of the formalism is that it is designed with computational
efficiency in mind.  While one could debate the relative merits of various
linguistic formalisms, the existence of mature tools and resources, such as the
CCGBank \cite{Hockenmaier:2007}, the Groningen Meaning Bank \cite{Basile:2012}
and the C\&C Tools \cite{Curran:2007} is another big advantage for CCG.

CCG's transparent surface stems from its categorial property: Each point in a
derivation corresponds directly to an interpretable category.  These categories
(or types) associated with each term in a CCG govern how this term can be
combined with other terms in a larger structure, implicitly making them
semantically expressive.

\begin{figure}[t]
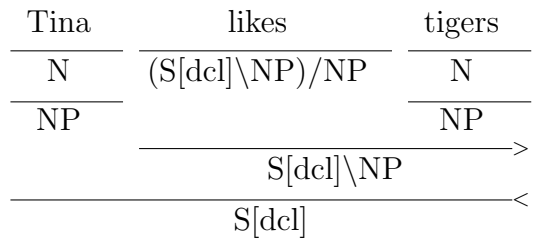
\begin{center}
\deriv{3}{
\rm ~~~Tina~~~			&	\rm ~~~likes~~~ 			& 		\rm ~~~tigers~~~		\\
\uline{1}		&	\uline{1}			&		\uline{1}	\\
\rm   N         &   \rm ~~(S[dcl]\backslash NP)/NP~~	&       \rm  N      \\
\uline{1}		&				&		\uline{1}	\\
\rm   NP 		&	 &		\rm NP			\\
				& 			\fapply{2}						\\
				&	\mc{2}{\rm 	S[dcl]\backslash NP}					\\
\bapply{3} \\
\mc{3}{\rm    S[dcl]} \\
}
\caption[CCG derivation example]{CCG derivation for \textit{Tina likes tigers} with forward ($>$) and
  backward application ($<$). Horizontal lines without a marker indicate lexical
  substitution or dictionary lookup.}\label{fig:syntax:ccg}
\end{center}\end{figure}

For instance in Figure \ref{fig:syntax:ccg}, the word \textit{likes} has type
{\it (S[dcl]\textbackslash NP)/NP}, which means that it first looks for a type
\textit{NP} to its right hand side.  Subsequently the expression \textit{likes
  tigers} (as type {\it S[dcl]\textbackslash NP}) requires a second \textit{NP}
on its left. The final type of the phrase \textit{S[dcl]} indicates a sentence
and hence a complete CCG proof.

Thus at each point in a CCG parse we can deduce the possible next steps in the
derivation by considering the available types and combinatory rules.

Considering only the types of each word in a sentence, we can derive a parse
tree consisting of a set of rules applied in a certain order in order to
generate that sentence.  Any given sentence in CCG is thus parametrised by its
categories and the combinatory rules with which these categories are composed
together to form a sentence.  For our models in the rest of this chapter, we will
exploit such parse trees and the combinatory rules used within, to govern the
structure and combination mechanisms of our RAE.

\section{Model}\label{sec:syntax:model}

\begin{table}[t]
  \centering
  \begin{tabular}{@{}ll@{}}
    \toprule
    Model & CCG Elements\\
    \midrule
    \modA & parse \\
    \modB & parse + rules \\
    \modC & parse + rules + types \\
    \modD & parse + rules + child types \\
    \bottomrule
  \end{tabular}
  \caption[Aspects of the CCG formalism used by the models]{Aspects of the CCG formalism used by the different models explored in
    this chapter.}\label{tab:syntax:models}
\end{table}

\begin{table*}
  \centering
  \begin{tabular}{@{}ll@{}l@{}}
    \toprule
    Model & \multicolumn{2}{l}{Encoding Function} \\
    \midrule
    \modA & $f(x,y)$ & $= g\left(W (x\|y) + b \right)$ \\[1.1ex]
    \modB & $f(x,y,c)$ & $= g\left(W^c (x\|y) + b^c \right)$ \\[0.9ex]
    \modC & $f(x,y,c,t)$ & $= g\left(\sum_{p\in\{c,t\}}\left(W^p (x\|y) + b^p \right)\right)$ \\[1.1ex]
    \modD & $f(x,y,c,tx,ty)$ & $= g\left(W^c\left(W^{tx}x + W^{ty}y\right) + b^c\right)$ \\[1.1ex]
    \bottomrule
  \end{tabular}
  \caption[Encoding functions of the four CCAE models]{Encoding functions of the four CCAE models discussed here.}\label{tab:syntax:encfunctions}
\end{table*}

The models in this chapter combine the power of recursive, vector-based models
with the linguistic intuition of the CCG formalism.  Their purpose is to learn
semantically meaningful vector representations for sentences and phrases  of
variable size, while the purpose of this work is to investigate the use of
syntax and linguistic formalisms in such vector-based compositional models.

We assume a CCG parse to be given.  Let $C$ denote the set of combinatory rules,
and $T$ the set of categories used, respectively.
We assume a finite set of combinatory rules $C$ (Table
  \ref{tab:syntax:ccgcombinators}) and categories $S$.

\begin{table}
  \centering
  \begin{tabular}{@{}llr@{}}
    \toprule
    Combinator & Description & Frequency \\
    \midrule
    FA (${>}$) & forward application & 60.000 \\
    BA (${<}$) & backward application & 25.000 \\
    LEX      & type-changing  & 14.000 \\
    CONJ     & coordination & 6.000 \\
    RP       & right punctuation  & 8.000 \\
    LP       & left punctuation  & 2.000 \\
    BX (${<}B_x$)      & backward cross-composition  & 1.000 \\
    TR (${>}T$)      & type-raising  & 1.000 \\
    FC (${>}B$)      & forward composition  & 1.000 \\
    BC (${<}B$)      & backward composition  & 300 \\
    FUNNY    & special cases  & 160 \\
    RTC      & right punctuation type-changing  & 110 \\
    LTC      & left punctuation type-changing  & 100 \\
    GBX      & generalised backward cross-composition  & 14  \\
  \bottomrule
\end{tabular}
\caption{CCG combinatory rules considered in our models. Combinators are based
  on those implemented in the C\&C parser \protect\cite{Curran:2007}. Frequency
  indicates the rounded number of observations on the SP dataset.
  }\protect\label{tab:syntax:ccgcombinators}
\end{table}

We use the parse tree to structure an RAE, so that each combinatory step is
represented by an autoencoder function.  We refer to these models as Combinatory
Categorial Autoencoders (CCAE). In total we describe four models, each making
increased use of the CCG formalism compared with the previous one (Table
  \ref{tab:syntax:models}).

As an internal baseline we use model \modA, which is an RAE structured along a
CCG parse tree.  \modA uses a single weight matrix each for the encoding and
reconstruction step (Table \ref{tab:syntax:encfunctions}).  This model is
similar to \newcite{Socher:2011}, except that we use a fixed structure in place
of the greedy tree building approach.  As \modA uses only minimal syntactic
guidance, this should allow us to better ascertain to what degree the use of
syntax helps our semantic models.

Our second model (\modB) uses the composition function in equation
(\ref{eqn:syntax:modB}), with $c \in C$.
\begin{align}
&f^{enc}(x,y,c) = g\left(W_{enc}^c (x\|y) + b_{enc}^c \right) \label{eqn:syntax:modB} \\
&f^{rec}(e,c) = g\left(W_{rec}^c e + b_{rec}^c \right) \nonumber
\end{align}
This means that for every combinatory rule we define an equivalent autoencoder
composition function by parametrizing both the weight matrix and bias on the
combinatory rule (e.g.~Figure \ref{fig:syntax:ccgfa}).

\begin{figure}[t]
  \centering
\includegraphics[scale=0.5]{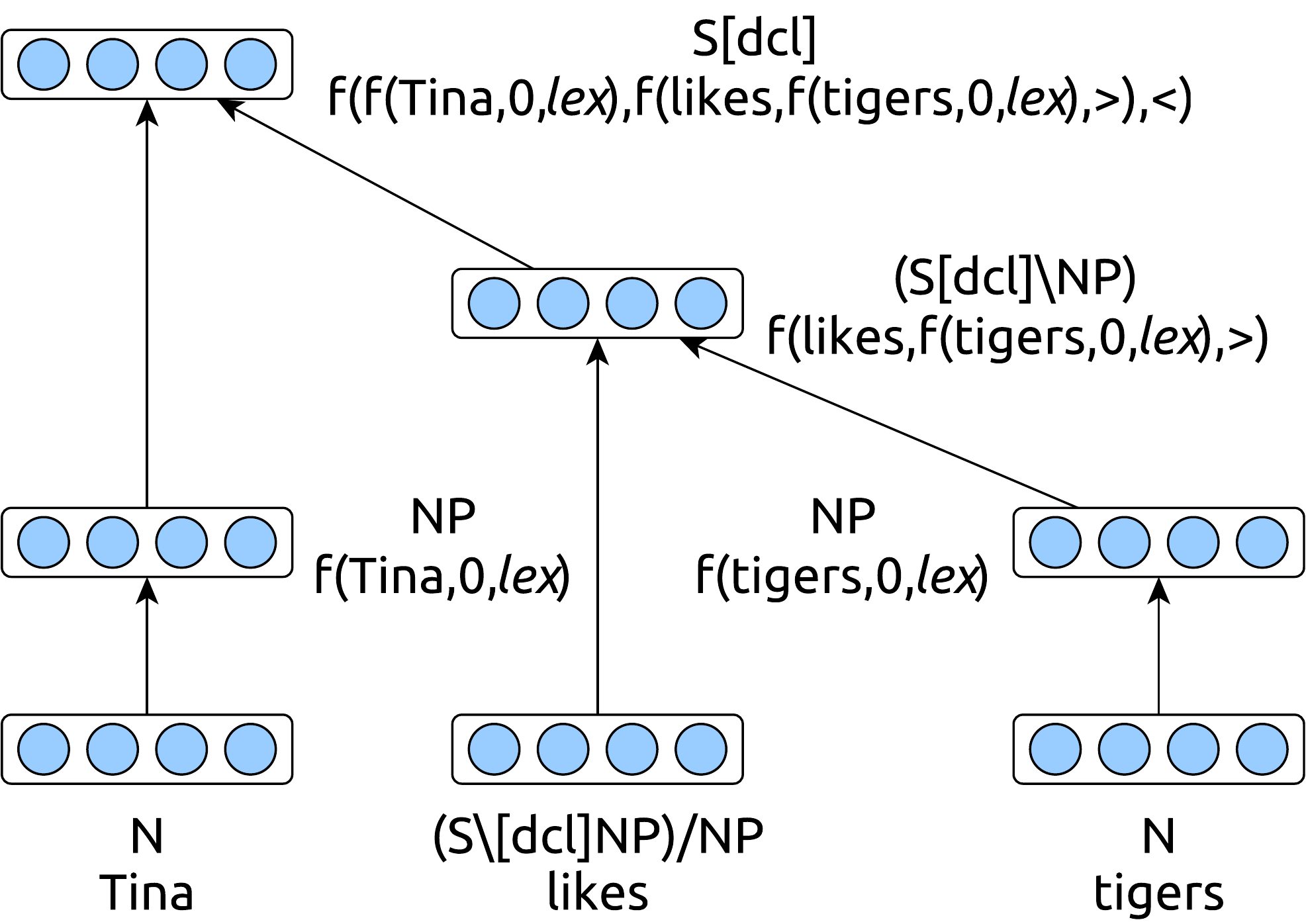}
\caption[\modB applied to \textit{Tina likes tigers}]{\modB applied to \textit{Tina likes tigers}. Next to each vector are
  the CCG category (top) and the word or function representing it (bottom).
  \textit{lex} describes the unary type-changing operation. $>$ and $<$ are
  forward and backward application.}\label{fig:syntax:ccgrae}
\end{figure}

In this model, as in the following ones, we assume a reconstruction step
symmetric with the composition step.  For the remainder of this chapter we will
focus on the composition step and drop the use of \textit{enc} and \textit{rec}
in variable names where it isn't explicitly required.  Figure
\ref{fig:syntax:ccgrae} shows model \modB applied to our previous example
sentence.

\begin{figure}[t]
\begin{subfigure}[c]{0.5\linewidth}
  \centering
\deriv{2}{
{\rm~~~\alpha :X/Y	} &	{\rm \beta :Y~~~}  \\
			\fapply{2}				\\
\mc{2}{ \alpha\beta :X}				\\
}
\end{subfigure}\begin{subfigure}[c]{0.5\linewidth}
  \centering
$g\left(W_{enc}^{{>}}(\alpha\|\beta)+b_{enc}^{{>}}\right)$
\end{subfigure}
\caption[Forward application in CCG and as an autoencoder rule]{Forward application as CCG combinator and autoencoder rule respectively.}\label{fig:syntax:ccgfa}
\end{figure}

While \modB uses only the combinatory rules, we want to make fuller use of the
linguistic information available in CCG.  For this purpose, we build another
model \modC, which parametrizes on both the combinatory rule $c \in C$ and the
CCG category $t \in T$ at every step (see Figure \ref{tab:syntax:encfunctions}).  This
model provides an additional degree of insight, as the categories $T$ are
semantically and syntactically more expressive than the CCG combinatory rules by
themselves.  Summing over weights parametrised on $c$ and $t$ respectively, adds
an additional degree of freedom and also allows for some model smoothing.

An alternative approach is encoded in model \modD.  Here we consider the
categories not of the element represented, but of the elements it is generated
from together with the combinatory rule applied to them.  The intuition is that
in the first step we transform two expressions based on their syntax.
Subsequently we combine these two conditioned on their joint combinatory rule.

An important aspect of all of these models is the use of a non-linearity in the
composition function. This non-linearity causes an interdependence between the
model inputs. As we discussed in Chapter \ref{chapter:compositional}, there are
a number of possibilities for modelling function-like behaviour in vector
composition, with matrix-vector- and tensor-composition being an obvious choice.
However, as pointed out earlier, such tensor-based models would require an
excessively large number of model parameters. The use of non-linearities can
thus be seen as a cheaper alternative (in terms of model size) to
tensor-composition for modelling functional dependence between words or larger
linguistic units. See e.g. \newcite{Bengio:2009} for a more detailed account of
this. The interdependence caused by the non-linearity can be observed by
considering the derivative of a non-linear function---here, the sigmoid
function---compared with that of a linear function:
\begin{align}
  &\text{lin}(x,y) = W_1x + W_2y \\
  &\frac{\partial \text{lin}(x,y)}{\partial x} = W_1 \\[1.4em]
  &\text{nonlin}(x,y) = \text{sigmoid}(x+y) \\
  &\frac{\partial \text{nonlin}(x,y)}{\partial x} = \text{sigmoid}(x+y)
  \left(1-\text{sigmoid}(x+y)\right)
\end{align}
The important difference here is that the gradient of the non-linear function
with respect to $x$ depends on $y$. In the linear case this interdependency does
not exist. This gradient interdependence caused by the non-linear function thus
enables a model to capture functional relationships between inputs beyond the
modelling capabilities of a linear model.

\section{Learning}

Here we briefly discuss unsupervised learning for our models.  Subsequently we
describe how these models can be extended to allow for semi-supervised training
and evaluation.

Let $\theta = \left(\mathcal{W},\mathcal{B},L\right)$ be our model parameters
and $\lambda$ a vector with regularization parameters for all model parameters.
$\mathcal{W}$ represents the set of all weight matrices, $\mathcal{B}$ the set
of all biases and $L$ the set of all word vectors.  Let $N$ be the set of
training data consisting of tree-nodes $t_n$ with inputs $x_n,y_n$ and
reconstruction $r_n$.  The error at $t_n$ given $\theta$ is:
\begin{align}
E(t_n|\theta) = \frac{1}{2} \Big\|r_n - \left(x_n\|y_n\right)\Big\|^2
\end{align}

The gradient of the regularised objective function then becomes:
\begin{align}
  \frac{\partial J}{\partial \theta} = \frac{1}{N} \sum_{n}^{N} \frac{\partial
    E(t_n|\theta)}{\partial \theta} + \lambda\theta
\end{align}

We learn the gradient using backpropagation through structure
\cite{Goller:1996}, and minimize the objective function using L-BFGS. See
\S\ref{sec:comp:learn:bpts} for details on the backpropagation algorithm.

\subsection{Supervised Learning}

The unsupervised method described so far learns a vector representation for each
sentence.  Such a representation can be useful for some tasks such as paraphrase
detection, but is not sufficient for other tasks such as sentiment
classification, which we are considering in this chapter.

In order to extract sentiment from our models, we extend them by adding a
supervised classifier on top, using the learned representations $v$ as input for
a binary classification model:
\begin{equation}
pred(l{=}1|v,\theta) = \text{sigmoid}(W_{\textit{label }}v+b_{\textit{label}})
\end{equation}
Given our corpus of CCG parses with label pairs $(N,l)$, the new objective
function becomes:
\begin{equation}
J = \frac{1}{N} \sum_{(N,l)} E(N,l,\theta) + \frac{\lambda}{2}||\theta||^2
\end{equation}
Assuming each node $n \in N$ contains children $x_n,y_n$, encoding $e_n$ and
reconstruction $r_n$, so that $n = \{x,y,e,r\}$ this breaks down into:
\begin{align}
  E(N,l,\theta) &= \sum_{n\in N} \alpha E_{rec}\left(n,\theta\right) +
  (1{-}\alpha) E_{lbl}(e_n,l,\theta) \\
  E_{rec}(n,\theta) &= \frac{1}{2} \Big\|[x_n\|y_n] - r_n\Big\|^2  \\
  E_{lbl}(e,l,\theta) &= \frac{1}{2} \left\|l - e\right\|^2
\end{align}
This is the combination of the two training signals described in
\S\ref{sec:comp:sig:rae} and \S\ref{sec:comp:sig:class}.

\section{Experiments}

Here, we describe the evaluations used to determine the performance of our
model. First, we use two corpora for sentiment analysis, which allow us to
compare model performance with a number of related approaches.
Subsequently, we perform a small qualitative analysis of the model to get a
better understanding of whether the combination of CCG parse structures and
RAE can learn semantically expressive embeddings.

In our experiments we use the hyperbolic tangent as non-linearity $g$. Unless
stated otherwise we use word-vectors of size 50, initialized using the
embeddings provided by \newcite{Turian:2010} based on the model of
\newcite{Collobert:2008}.\footnote{\url{http://www.metaoptimize.com/projects/wordreprs/}}

We use the C\&C parser \cite{Clark:2007} to generate CCG parse trees for the
data used in our experiments.  For models \modC and \modD we use the 25 most
frequent CCG categories (as extracted from the British National Corpus) with an
additional general weight matrix in order to catch all remaining types (Appendix
  \ref{appendix:syntax:cat}).

\subsection{Sentiment Analysis}\label{sec:syntax:sentiment}

We evaluate our model on the MPQA opinion corpus \cite{Wiebe:2005}, which
annotates expressions for sentiment.\footnote{\url{http://mpqa.cs.pitt.edu/}}
The corpus consists of 10,624 instances with approximately 70 percent describing
a negative sentiment.  We apply the same pre-processing as \newcite{Nakagawa:2010}
and \newcite{Socher:2011} by using an additional sentiment lexicon
\cite{Wilson:2005} during the model training for this experiment.

As a second corpus we make use of the sentence polarity (SP) dataset v1.0
\cite{Pang:2005}.\footnote{\url{http://www.cs.cornell.edu/people/pabo/movie-review-data/}}
This dataset consists of 10662 sentences extracted from movie reviews which are
manually labelled with positive or negative sentiment and equally distributed
across sentiment.

\paragraph{Experiment 1: Semi-Supervised Training}

In the first experiment, we use a semi-supervised training strategy combining an
autoencoder signal with a classification signal such as described in Chapter
\ref{chapter:compositional}. We initialise our models with the embeddings
provided by \newcite{Turian:2010}. The results of this evaluation are in Table
\ref{tab:syntax:sent-results}.  While we achieve the best performance on the
MPQA corpus, the results on the SP corpus are less convincing.  Perhaps
surprisingly, the simplest model \modA outperforms the other models on this
dataset.

When considering the two datasets, sparsity seems a likely explanation for this
difference in results: In the MPQA experiment most instances are very short with
an average length of 3 words, while the average sentence length in the SP corpus
is 21 words.  The MPQA task is further simplified through the use of an
additional sentiment lexicon.  Considering dictionary size, the SP corpus has a
dictionary of 22k words, more than three times the size of the MPQA dictionary.

This issue of sparsity is exacerbated in the more complex CCAE models, where the
training points are spread across  different CCG types and rules.  While the
initialization of the word vectors with previously learned embeddings (as was
  previously shown by \newcite{Socher:2011}) helps the models, all other model
variables such as composition weights and biases are still initialised randomly
and thus highly dependent on the amount of training data available.

\begin{table}[t]
  \centering
  \begin{tabular}{@{}l@{ }cc@{}}
    \toprule
    Method & MPQA & SP \\
    \midrule
    Voting with two lexica			& 81.7 & 63.1 \\
    MV-RNN \cite{Socher:2012}		&  -   & 79.0 \\
    RAE (rand) \cite{Socher:2011}	& 85.7 & 76.8 \\
    TCRF \cite{Nakagawa:2010} 		& 86.1 & 77.3 \\
    RAE (init) \cite{Socher:2011}	& 86.4 & 77.7 \\
    NB \cite{Wang:2012} 				& 86.7 & 79.4 \\
    \midrule
    \modA							& 86.3 & 77.8 \\
    \modB							& 87.1 & 77.1 \\
    \modC							& 87.1 & 77.3 \\
    \modD							& \textbf{87.2} & 76.7 \\
    \bottomrule
  \end{tabular}
  \caption[Accuracy on sentiment classification tasks]{Accuracy of sentiment classification on the sentiment polarity (SP)
    and MPQA datasets. For NB we only display the best result among a larger
    group of models analysed in that paper.}\label{tab:syntax:sent-results}
\end{table}

\paragraph{Experiment 2: Pretraining}

Due to our analysis of the results of the initial experiment, we ran a second
series of experiments on the SP corpus.  We follow \newcite{Scheible:2013} for
this second series of experiments, which are carried out on a random 90/10
training-testing split, with some data reserved for development.

Instead of initialising the model with external word embeddings, we first train
it on a large amount of data with the aim of overcoming the sparsity issues
encountered in the previous experiment.  Learning is thus divided into two
steps:

The first, unsupervised training phase, uses the British National Corpus
together with the SP corpus.  In this phase only the reconstruction signal is
used to learn word embeddings and transformation matrices.  Subsequently, in the
second phase, only the SP corpus is used, this time with both the reconstruction
and the label error.

By learning word embeddings and composition matrices on more data, the model is
likely to generalise better.  Particularly for the more complex models, where
the composition functions are conditioned on various CCG parameters, this should
help to overcome issues of sparsity.

If we consider the results of the pre-trained experiments in Table
\ref{tab:syntax:pre-results}, this seems to be the case.  In fact, the trend of
the previous results has been reversed, with the more complex models now
performing best, whereas in the previous experiments the simpler models
performed better.  Using the Turian embeddings instead of random initialisation
did not improve results in this setup.

\begin{table}[t]
  \centering
  \begin{tabular}{@{}l@{ }cc@{}}
    \toprule
    Model & \multicolumn{2}{c}{Training} \\
    \cmidrule{2-3}
   & Regular & Pretraining \\
    \midrule
    \modA & 77.8 & 79.5 \\
    \modB & 76.9 & 79.8 \\
    \modC & 77.1 & 81.0 \\
    \modD & 76.9 & 79.7 \\
    \bottomrule
  \end{tabular}
  \caption[Effect of pretraining on model performance on the SP dataset]{Effect
    of pretraining on model performance on the SP dataset. Results are reported
    on a random subsection of the SP corpus; thus numbers for the regular
    training method differ slightly from those in Table
    \ref{tab:syntax:sent-results}.}\label{tab:syntax:pre-results}
\end{table}

\subsection{Qualitative Analysis}

\begin{table}[t]
  \centering
  \begin{tabular}{@{}ll@{}}
    \toprule
    Expression & Most Similar \\
    \midrule
    convey the message of peace & safeguard peace and security \\
    keep alight the flame of & keep up the hope \\
    has a reason to repent & has no right \\
    a significant and successful strike & a much better position \\
    it is reassuring to believe & it is a positive development \\
    expressed their satisfaction and support & expressed their admiration and surprise \\
    is a story of success & is a staunch supporter \\
    are lining up to condemn & are going to voice their concerns \\
    more sanctions should be imposed & charges being leveled \\
    could fray the bilateral goodwill & could cause serious damage \\
    \bottomrule
  \end{tabular}
  \caption[Phrases and their semantically closes match according to
    \modD]{Phrases from the MPQA corpus and their semantically closest match
    according to \modD .}\label{tab:syntax:qual}
\end{table}

To get better insight into our models we also perform a small qualitative
analysis.  Using one of the models trained on the MPQA corpus, we generate
word-level representations of all phrases in this corpus and subsequently
identify the most related expressions by using the cosine distance measure.  We
perform this experiment on all expressions of length 5, considering all
expressions with a word length between 3 and 7 as potential matches.

As can be seen in Table \ref{tab:syntax:qual}, this works with varying success.
Linking expressions such as \textit{conveying the message of peace} and
\textit{safeguard(ing) peace and security} suggests that the model does learn
some form of semantics.

On the other hand, the connection between \textit{expressed their satisfaction
  and support} and \textit{expressed their admiration and surprise} suggests
that the pure word level content still has an impact on the model analysis.
Likewise, the expressions \textit{is a story of success} and \textit{is a
  staunch supporter} have some lexical but little semantic overlap.  Further
reducing this link between the lexical and the semantic representation is an
issue that should be addressed in future work in this area.

\section{Discussion}

Overall, our models compare favourably with the state of the art.  On the MPQA
corpus model \modD achieves the best published results we are aware of, whereas
on the SP corpus we achieve competitive results.  With an additional,
unsupervised training step we achieved results beyond the current state of the
art on this task, too.

\paragraph{Semantics}

The qualitative analysis as well as the results on the sentiment analysis task
suggest that the
CCAE models are capable of learning semantics.  An advantage of our
approach---and of autoencoders generally---is their ability to learn in an
unsupervised setting.  The pre-training step for the sentiment task was
essentially the same training step as used for the qualitative analysis.
While other models such as the MV-RNN \cite{Socher:2012} achieve good results on
a particular task, they do not allow unsupervised training.  This prevents the
possiblity of pretraining, which we showed to have a big impact on results, and
further prevents the training of general models: The CCAE models can be used for
multiple tasks without the need to re-train the main model.

\paragraph{Complexity}

Previously in this thesis we argued that our models combined the strengths of
other approaches.  By using a grammar formalism we increase the expressive power
of the model while the complexity remains low.  For the complexity analysis see
Table \ref{tab:syntax:model-complexity}.  We strike a balance between the greedy
approaches (e.g. \newcite{Socher:2011}), where learning is quadratic in the
length of each sentence and existing syntax-driven approaches such as the MV-RNN
of \newcite{Socher:2012}, where the size of the model, that is the number of
variables that needs to be learned, is quadratic in the size of the
word-embeddings.

\begin{table}[t]
  \centering
  \begin{tabular}{@{}lll@{}}
    \toprule
    & \multicolumn{2}{l}{Complexity} \\
    \cmidrule{2-3}
    Model & Size & Learning \\
    \midrule
    MV-RNN & $\mathcal{O}(nw^2)$ & $O(l)$ \\
    RAE & $\mathcal{O}(nw)$ & $O(l^2)$ \\
    CCAE-* & $\mathcal{O}(nw)$ & $O(l)$ \\
    \bottomrule
  \end{tabular}
  \caption[Comparison of model complexity]{Comparison of models. $n$ is
    dictionary size, $w$ embedding width, $l$ is sentence length. Assuming a
    training corpus of $c$ sentences, we can assume that $cl \gg n \gg w$.
    Additional factors such as CCG rules and types are treated as small
    constants for the purposes of this
    analysis.}\label{tab:syntax:model-complexity}
\end{table}

\paragraph{Sparsity}

Parametrizing on CCG types and rules increases the size of the model compared to
a greedy RAE \cite{Socher:2011}.  The effect of this was highlighted by the
sentiment analysis task, with the more complex models performing worse in
comparison with the simpler ones.
We were able to overcome this issue by using additional training data.  Beyond
this, it would also be interesting to investigate the relationships between
different types and to derive functions to incorporate this into the learning
procedure.  For instance model learning could be adjusted to enforce some
mirroring effects between the weight matrices of forward and backward
application, or to support similarities between those of forward application and
composition.

\paragraph{CCG-Vector Interface}

Exactly how the information contained in a CCG derivation is best applied to a
vector space model of compositionality remains an open question: our
investigation of this matter by exploring different model setups has proved
somewhat inconclusive. While \modD incorporated the deepest conditioning on the
CCG structure, it did not decisively outperform the simpler \modB which just
conditioned on the combinatory operators. Compared with prior work on related
models and the same experiments, however, demonstrated that any conditioning is
better than none. As pointed out above, sparsity---which we found to have a big
impact in our experiments on pre-training---favours simpler models with less
model parameters. By extension the results suggest that more complex
conditioning on the CCG parse is beneficial, but only if sufficient data is
available to appropriately train such a model.

As discussed in \S\ref{sec:syntax:model}, we approximated the functional aspect
of CCG composition using non-linearities in our model. The relative performance
of our models compared with the prior state of the art suggests that this
approximation works. However, an interesting avenue of future research would be
to improve this approximation using low-rank tensor factorisations or similar
approaches together with composition functions such as suggested by
\newcite{Clark:2013}, \newcite{Grefenstette:2013a} or \newcite{Maillard:2014}.

\section{Summary}

In this chapter we have brought a more formal notion of semantic compositionality
to vector space models based on recursive autoencoders. This was achieved
through the use of the CCG formalism to provide a conditioning structure for the
matrix vector products that define the RAE.

We have explored a number of models, each of which conditions the compositional
operations on different aspects of the CCG derivation. Our experimental
findings indicate a clear advantage for a deeper integration of syntax over
models that use only the bracketing structure of the parse tree. Further, we
demonstrated that the functional aspect of CCG composition can be approximated
in a cheap yet sensible fashion through the use of non-linearities extending a
fully-additive composition model such as the ones discussed in Chapter
\ref{chapter:compositional}.

The most effective way to condition the compositional operators on the syntax
remains unclear. Once the issue of sparsity had been addressed, the complex
models outperformed the simpler ones. Among the complex models, however, we
could not establish significant or consistent differences to convincingly argue
for a particular approach. This uncertainty could also be linked to sparsity,
considering the relatively small amounts of training data used in our
experiments.

While the connections between formal linguistics and vector space approaches to
NLP may not be immediately obvious, we believe that there is a case for the
continued investigation of ways to best combine these two schools of thought.
The approach presented here is one step towards the reconciliation of
traditional formal approaches to compositional semantics with modern machine
learning.

The system proposed in this chapter relies heavily on grammatical annotation
(here, CCG parses), both for learning distributed word representations as well
as for learning a composition model based on these word level representations.
Our results clearly indicate that this is a suitable mechanism when applied to
the English language and when considering a specific task such as sentiment
analysis where data is available for supervised training. However, two questions
remain. First, it is unclear how this approach can be extended to less resource
fortunate languages, and second, it is difficult to see how such a
semi-supervised training setup can be extended to learn more generic semantic
representations rather than representations primed for sentiment analysis or a
similar such task.

With these questions in mind, we next investigate unsupervised approaches
for representation learning. We focus on
such models with three objectives in mind. First, we want to investigate
approaches that can easily be applied to a multitude of languages, which
effectively forces us to abandon the use of syntax-driven composition and
learning functions. Second, words from these multiple languages should be
projected into joint-space embeddings, thereby enabling us to use the
representations independent of their respective source languages. Finally, we
specifically want to design a representation learning method that results in
semantically grounded representations, similar to e.g. representations learned
based on the distributional hypothesis (see \S\ref{sec:ts:distributional}) in a
monolingual scenario. The outcome of this investigation is presented in Chapter
\ref{chapter:multilingual}.

\newcommand{\addMod}{\textsc{Add}\xspace}
\newcommand{\addModplus}{\textsc{Add+}\xspace}
\newcommand{\flatMod}{\textsc{Bi}\xspace}
\newcommand{\flatModplus}{\textsc{Bi+}\xspace}
\newcommand{\docMod}{\textsc{Doc}\xspace}
\newcommand{\docModadd}{\textsc{Doc/Add}\xspace}
\newcommand{\docModflat}{\textsc{Doc/Bi}\xspace}

\newcommand{\CVM}{\textsc{CVM}\xspace}
\newcommand{\single}{\textit{single}\xspace}
\newcommand{\joint}{\textit{joint}\xspace}

\chapter{Multilingual Approaches for Learning Semantics}\label{chapter:multilingual}

\emptyfootnote{The material in this chapter was originally presented in
  \newcite{Hermann:2014:ICLR} and \newcite{Hermann:2014:ACLphil}.}

\begin{chapterabstract}
  This chapter continues our investigation into both compositional and word
  level semantic representations. Building on the conclusions drawn in Chapters
  \ref{chapter:frame-semantic} and \ref{chapter:syntax}, we devise an extension
  to the distributional hypothesis \S\ref{sec:ts:distributional} for
  multilingual data and joint-space embeddings. The models presented in this
  chapter leverage parallel data and learn to strongly align the embeddings of
  semantically equivalent sentences, while maintaining sufficient distance
  between those of dissimilar sentences. The models do not rely on word
  alignments or any syntactic information and are successfully applied to a
  number of diverse languages. Instead of employing word alignment we achieve
  semantic transfer across languages using compositional representations at the
  sentence level. We extend our approach to learn semantic representations at
  the document level, too. We evaluate these models on two cross-lingual
  document classification tasks, outperforming the prior state of the art.
  Through qualitative analysis and the study of pivoting effects we demonstrate
  that our representations are semantically plausible and can capture semantic
  relationships across languages without parallel data.
\end{chapterabstract}

\section{Introduction}

As we have established in this thesis so far, distributed representations
provide a highly suitable mechanism for encoding semantics both at the word
level and at higher levels, such as demonstrated with the sentiment analysis
task in Chapter \ref{chapter:syntax}.
That said, as pointed out both in the previous chapter as well as in the
analysis of compositional semantics learning signals in Chapter
\ref{chapter:compositional}, it is difficult to distinguish between
\textit{semantic} and \textit{task-specific} representations when considering a
task such as sentiment analysis.

In order to remove such task-specific biases from the representation learning
process, a different approach is required. Within a monolingual context, the
distributional hypothesis \cite{Firth:1957} forms the basis of most approaches
for learning word representations. This hypothesis is attractive because it
offers an approach to learn distributed representations independent of a
particular task or signal and therefore carries the promise of learning
task-independent semantic representations.

Here we extend this distributional hypothesis to multilingual data and
joint-space embeddings. We present a novel unsupervised technique for learning
semantic representations that leverages parallel corpora and employs semantic
transfer through compositional representations. Unlike most methods for learning
word representations, which are restricted to a single language, our approach
learns to represent meaning across languages in a shared multilingual semantic
space. Furthermore, by employing a multilingual variation of the distributional
hypothesis, we learn representations that are not only multilingual, but also
task-independent and semantically grounded to a degree not available to most
other approaches for learning distributed representations.

To show the efficacy of our model, we present experiments on two corpora.
First, we show that for cross-lingual document classification on the Reuters
RCV1/RCV2 corpora \cite{Lewis:2004}, we outperform the prior state of the art
\cite{Klementiev:2012}.  Second, we also present classification results on a
massively multilingual corpus which we derive from the TED corpus
\cite{Cettolo:2012}.  The results on this task, in comparison with a number of
strong baselines, further demonstrate the relevance of our approach and the
success of our method in learning multilingual semantic representations over a
wide range of languages.

\section{Overview}

Distributed representation learning describes the task of learning continuous
representations for discrete objects.  Here, we focus on learning semantic
representations and investigate how the use of multilingual data can improve
learning such representations at the word and higher level.  We present a model
that learns to represent each word in a lexicon by a continuous vector in
$\mathbb{R}^d$.  Such distributed representations allow a model to share meaning
between similar words, and have been used to capture semantic, syntactic and
morphological content \cite[\textit{inter alia}]{Collobert:2008,Turian:2010}.

We describe a multilingual objective function that uses a noise-contrastive
update between semantic representations of different languages to learn these
word embeddings. As part of this, we use a compositional vector model (\CVM,
  henceforth) to compute semantic representations of sentences and documents. A
\CVM learns semantic representations of larger syntactic units given the
semantic representations of their constituents \cite[\textit{inter
    alia}]{Clark:2007a,Mitchell:2008,Baroni:2010,Grefenstette:2011,Socher:2012,Hermann:2013:ACL}.

A key difference between our approach and those listed above is that we only
require sentence-aligned parallel data in our otherwise unsupervised learning
function. This removes a number of constraints that normally come with \CVM
models, such as the need for syntactic parse trees, word alignment or annotated
data as a training signal. At the same time, by using multiple \CVM{s} to
transfer information between languages, we enable our models to capture a
broader semantic context than would otherwise be possible.

The idea of extracting semantics from multilingual data stems from prior work in
the field of semantic grounding.  Language acquisition in humans is widely seen
as grounded in sensory-motor experience \cite{Bloom:2001,Roy:2003}.  Based on
this idea, there have been some attempts at using multi-modal data for learning
better vector representations of words (e.g. \newcite{Srivastava:2012}).  Such
methods, however, are not easily scalable across languages or to large amounts
of data for which no secondary or tertiary representation might exist.

Parallel data in multiple languages provides an alternative to such secondary
representations, as parallel texts share their semantics, and thus one language
can be used to ground the other.  Some work has exploited this idea for
transferring linguistic knowledge into low-resource languages or to learn
distributed representations at the word level \cite[\textit{inter
    alia}]{Klementiev:2012,Zou:2013,Lauly:2013}.  So far almost all of this
work has been focused on learning multilingual representations at the word
level.  As distributed representations of larger expressions have been shown to
be highly useful for a number of tasks, it seems to be a natural next step to
attempt to induce these, too, cross-lingually.

\section{Approach}

Most prior work on learning compositional semantic representations employs parse
trees on their training data to structure their composition functions
\cite[\textit{inter alia}]{Socher:2012,Hermann:2013:ACL}. Further, these
approaches typically depend on specific \textit{semantic} signals such as
sentiment- or topic-labels for their objective functions.  While these methods
have been shown to work in some cases, the need for parse trees and annotated
data limits such approaches to resource-fortunate languages. Our novel method
for learning compositional vectors removes these requirements, and as such can
more easily be applied to low-resource languages.

Specifically, we attempt to learn semantics from multilingual data. The idea is
that, given enough parallel data, a shared representation of two parallel
sentences would be forced to capture the common elements between these two
sentences. What parallel sentences share, of course, are their semantics.
Naturally, different languages express meaning in different ways. We utilise
this diversity to abstract further from mono-lingual surface realisations to
deeper semantic representations.  We exploit this semantic similarity across
languages by defining a bilingual (and trivially multilingual) energy as
follows.

\begin{figure}[t]
  \centering
  \includegraphics[scale=0.75]{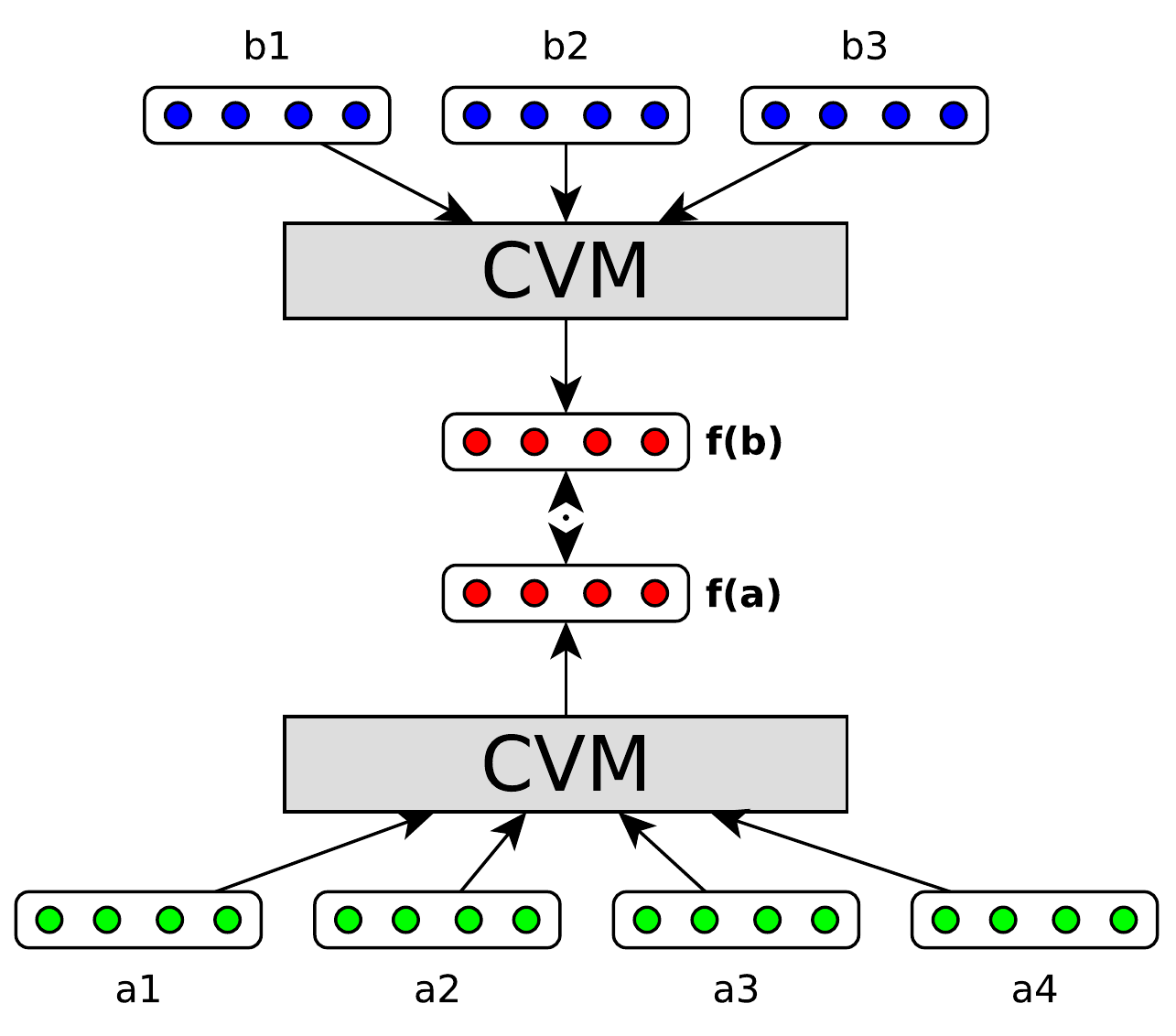}
  \caption[A parallel composition vector model]{Model with parallel input sentences $a$ and $b$. The model
    minimises the distance between the sentence level encoding of the bitext.
    Any composition functions (\CVM) can be used to generate the
    compositional sentence level representations.
    }\label{fig:bilingual}
  \vspace{-1em}
\end{figure}

Assume two functions ${f:X\rightarrow \mathbb{R}^d}$
and ${g:Y\rightarrow \mathbb{R}^d}$, which map sentences from languages $x$ and
$y$ onto distributed semantic representations in $\mathbb{R}^d$ (see Figure
  \ref{fig:bilingual}). Given a
parallel corpus $C$, we then define the energy of the model given two sentences
$(a,b) \in C$ as:
\begin{equation}
E_{bi}(a,b) = \left\| f(a) - g(b) \right\|^2\label{eqn:bi-error}
\end{equation}
We want to minimize $E_{bi}$ for all semantically equivalent sentences in the
corpus. In order to prevent the model from degenerating, we further introduce a
noise-constrastive large-margin update which ensures that the representations of
non-aligned sentences observe a certain margin from each other.
For every pair of parallel sentences $(a,b)$ we sample a number of additional
sentence pairs $(\cdot,n) \in C$, where $n$---with high probability---is not
semantically equivalent to $a$. We use these noise samples as follows:
\begin{equation}
E_{hl}(a,b,n) = \left[m + E_{bi}(a,b) - E_{bi}(a,n)\right]_{+}\nonumber
\end{equation}
where $[x]_{+} = max(x,0)$ denotes the standard hinge loss and $m$ is the margin.
This results in the following objective function:
\begin{equation}
J(\theta)=\sum_{(a,b) \in \mathcal{C}} \left( \sum_{i=1}^{k} E_{hl}(a,b,n_i) +
  \frac{\lambda}{2}\|\theta\|^2 \right)\label{eqn:objective}
\end{equation}
where $\theta$ is the set of all model variables.

\subsection{Two Composition Models}

The objective function in Equation \ref{eqn:objective} could be
coupled with any two given vector composition functions $f,g$ from the
literature. As we aim to apply our approach to a wide range of languages, we
focus on composition functions that do not require any syntactic information. We
evaluate the following two composition functions.

The first model, \addMod, represents a sentence by the sum of its word vectors.
This is a distributed bag-of-words approach as sentence ordering is not taken
into account by the model.

Second, the \flatMod model is designed to capture bigram information, using a
non-linearity over bigram pairs in its composition function:
\begin{equation}
  f(x) = \sum_{i=1}^{n} \text{tanh}\left(x_{i-1} + x_{i}\right)
\end{equation}
The use of a non-linearity enables the model to learn interesting interactions
between words in a document, which the bag-of-words approach of \addMod is not
capable of learning. We use the hyperbolic tangent as activation function. See
the discussion in Chapter \ref{chapter:syntax} (at the end of
  \S\ref{sec:syntax:model}) for more detail on the motivation for
non-linearities in such models.

\subsection{Document-level Semantics}\label{sec:docmod}

\begin{figure}[t]\centering
  \includegraphics[scale=0.45]{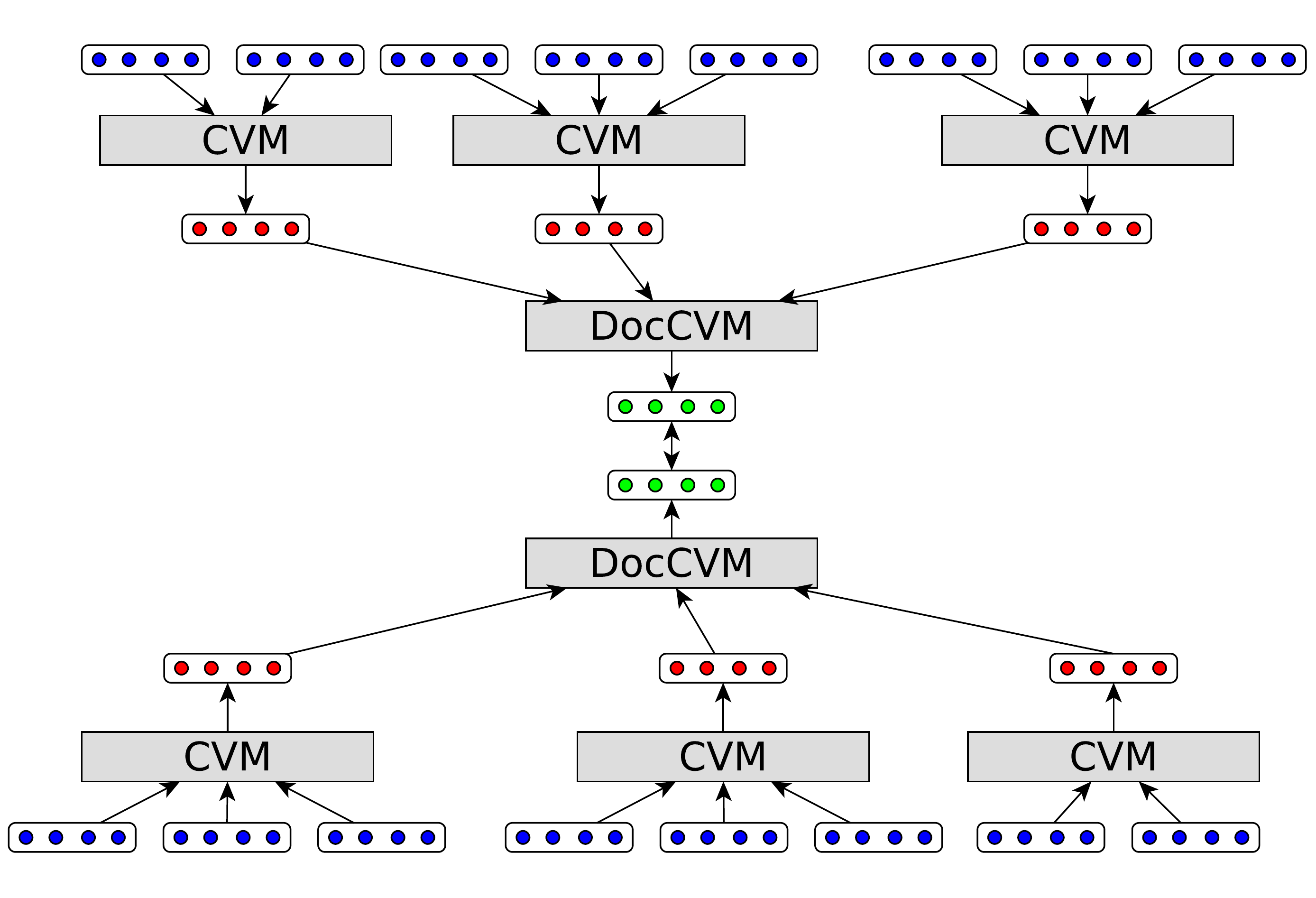}
  \caption[A parallel document-level compositional vector model]{Description of a parallel document-level compositional vector model
    (\docMod). The model recursively computes semantic representations for each
    sentence of a document and then for the document itself, treating the sentence
    vectors as inputs for a second \CVM.
    }\label{fig:docmod}
\end{figure}

For a number of tasks, such as topic modelling, representations of objects beyond
the sentence level are required. While most approaches to compositional
distributed semantics end at the sentence level, our model extends to document-level
learning quite naturally, by recursively applying the composition and objective
function (Equation \ref{eqn:objective}) to compose sentences into documents.
This is achieved by first computing semantic representations for each sentence
in a document.  Next, these representations are used as inputs in a higher-level
\CVM, computing a semantic representation of a document (Figure
  \ref{fig:docmod}).

This recursive approach integrates document-level representations into the
learning process. We can thus use corpora of parallel documents---regardless of
whether they are sentence aligned or not---to propagate a semantic signal back
to the individual words.
If sentence alignment is available, of course, the document-signal can simply be
combined with the sentence-signal, as we did with the experiments described in
\S\ref{sec:ted-cldc}.

This concept of learning compositional representations for documents contrasts
with prior work \cite[\textit{inter alia}]{Socher:2011,Klementiev:2012} who rely
on summing or averaging sentence-vectors if representations beyond the
sentence-level are required for a particular task.

We evaluate the models presented in this chapter both with and without the
document-level signal. We refer to the individual models used as \addMod and
\flatMod if used without, and as \docModadd and \docModflat is used with the
additional document composition function and error signal.

\section{Corpora}\label{sec:corpus}

We use two corpora for learning semantic representations and performing the
experiments described in the following section.

The Europarl corpus v7\footnote{\url{http://www.statmt.org/europarl/}}
\cite{Koehn:2005} was used during initial development and testing of our
approach, as well as to learn the representations used for the Cross-Lingual
Document Classification task described in \S\ref{sec:rcv-cldc}.  We considered
the English-German and English-French language pairs from this corpus.  From
each pair the final 100,000 sentences were reserved for development.

Second, we developed a massively multilingual corpus based on the TED
corpus\footnote{\url{https://wit3.fbk.eu/}} for IWSLT 2013 \cite{Cettolo:2012}.
This corpus contains English transcriptions and multilingual, sentence-aligned
translations of talks from the TED conference.  While the corpus is aimed at
machine translation tasks, we use the keywords associated with each talk to
build a subsidiary corpus for multilingual document classification as
follows.\footnote{\url{http://www.clg.ox.ac.uk/tedcldc/}}

The development sections provided with the IWSLT 2013 corpus were again reserved
for development.  We removed approximately 10 percent of the training data in
each language to create a test corpus (all talks with $id\geq1{,}400$).  The new
training corpus consists of a total of 12,078 parallel documents distributed
across 12 language pairs\footnote{English to Arabic, German, French, Spanish,
  Italian, Dutch, Polish, Brazilian Portuguese, Romanian, Russian and Turkish.
  Chinese, Farsi and Slovenian were removed due to the small size of those
  datasets.}. In total, this amounts to 1,678,219 non-English sentences (the
  number of unique English sentences is smaller as many documents are translated
  into multiple languages and thus appear repeatedly in the corpus).  Each
document (talk) contains one or several keywords.  We used the 15 most frequent
keywords for the topic classification experiments described in section
\S\ref{sec:ted-cldc}.

Both corpora were pre-processed using the set of tools provided by
cdec\footnote{\url{http://cdec-decoder.org/}} for tokenizing and
lowercasing the data.  Further, all empty sentences and their translations were
removed from the corpus.

\section{Experiments}

We report results on two experiments. First, we replicate the cross-lingual
document classification task of \newcite{Klementiev:2012}, learning distributed
representations on the Europarl corpus and evaluating on documents from the
Reuters RCV1/RCV2 corpora.  Subsequently, we design a multi-label classification
task using the TED corpus, both for training and evaluating.  The use of a wider
range of languages in the second experiments allows
us to better evaluate our models' capabilities in
learning a shared multilingual semantic representation.  We also investigate the
learned embeddings from a qualitative perspective in \S\ref{sec:qualitative}.

\subsection{Learning}

All model weights were randomly initialised using a Gaussian distribution
($\mu{=}0,\sigma^2{=}0.1$). We used the available development data to set our
model parameters. For each positive sample we used a number of noise samples ($k
  \in \{1,10,50\}$), randomly drawn from the corpus at each training epoch.  All
our embeddings have dimensionality $d{=}128$, with the margin set to
$m{=}d$.\footnote{On the RCV task we also report results for $d{=}40$ which
  matches the dimensionality of \newcite{Klementiev:2012}.} Further, we use L2
regularization with $\lambda{=}1$ and step-size in $\{0.01, 0.05\}$. We use 100
iterations for the RCV task, 500 for the TED single and 5 for the joint
corpora. We use the adaptive gradient method, AdaGrad \cite{Duchi:2011}, for
updating the weights of our models, in a mini-batch setting ($b \in \{10,50\}$).

\subsection{RCV1/RCV2 Document Classification}\label{sec:rcv-cldc}

We evaluate our models on the cross-lingual document classification (CLDC,
  henceforth) task first described in \newcite{Klementiev:2012}.  This task
involves learning language independent embeddings which are then used for
document classification across the English-German language pair.  For this, CLDC
employs a particular kind of supervision, namely using supervised training data
in one language and evaluating without further supervision in another.  Thus,
CLDC can be used to establish whether our learned representations are
semantically useful across multiple languages.

We follow the experimental setup described in \newcite{Klementiev:2012}, with
the exception that we learn our embeddings using solely the Europarl data and
use the Reuters corpora only during for classifier training and testing.  Each
document in the classification task is represented by the average of the
$d$-dimensional representations of all its sentences.  We train the multiclass
classifier using an averaged perceptron \cite{Collins:2002} with the same
settings as in \newcite{Klementiev:2012}.

We present results from four models.  The \addMod model is trained on 500k
sentence pairs of the English-German parallel section of the Europarl corpus.
The \addModplus model uses an additional 500k parallel sentences from the
English-French corpus, resulting in one million English sentences, each paired
up with either a German or a French sentence, with \flatMod and \flatModplus
trained accordingly.
The motivation behind \addModplus and \flatModplus is to investigate whether we
can learn better embeddings by introducing additional data from other languages.
A similar idea exists in machine translation where English
is frequently used to pivot between other languages \cite{Cohn:2007}.

\begin{table}[t]\centering
\begin{tabular}{@{}lrr@{}}\toprule
  \multicolumn{1}{@{}l}{Model} & en $\rightarrow$ de & de $\rightarrow$ en \\ \midrule
  \multicolumn{1}{@{}l}{Majority Class} & 46.8 & 46.8 \\
  \multicolumn{1}{@{}l}{Glossed} & 65.1 & 68.6 \\
  \multicolumn{1}{@{}l}{MT} & 68.1 & 67.4 \\
  \multicolumn{1}{@{}l}{I-Matrix} & 77.6 & 71.1 \\
  \midrule
  $dim = 40$ \\
  \addMod & 83.7 & 71.4 \\
  \addModplus & 86.2 & 76.9 \\
  \flatMod & 83.4 & 69.2 \\
  \flatModplus & 86.9 & 74.3 \\
  \midrule
  $dim = 128$ \\
  \addMod & 86.4 & 74.7 \\
  \addModplus & 87.7 & 77.5 \\
  \flatMod & 86.1 & 79.0 \\
  \flatModplus & \textbf{88.1} & \textbf{79.2} \\
  \bottomrule
\end{tabular}
\caption[Classification accuracy on the RCV corpus]{Classification accuracy for training on English and German with 1000
  labeled examples on the RCV corpus. Cross-lingual compositional representations (\addMod,
    \flatMod and their multilingual extensions), I-Matrix \cite{Klementiev:2012}
  translated (MT) and glossed (Glossed) word baselines, and the majority class
  baseline. The baseline results are from \newcite{Klementiev:2012}.}
\label{tab:results1k}
\end{table}

\pgfplotsset{every axis plot/.append style={line width=1pt}}

\begin{figure*}[t]
\begin{tikzpicture}
	\begin{axis}[
		grid=major,
		ytick={40,50,...,80},
		xlabel=Training Documents (de),
    		xtick={1,2,3,4,5,6},
	    xticklabels={$100$,$200$,$500$,$1000$,$5000$,$10$k},
		ylabel=Classification Accuracy (\%),
		legend columns=-1,
		legend entries={\addModplus, \flatModplus, I-Matrix, MT, Glossed},
		legend to name=sharedlegend,
		scale only axis,
		xmin=1,xmax=6,
		width=0.4\textwidth,
		height=0.27\textwidth,
		]
	\addplot[color=blue,mark=*] coordinates {
	(1, 68.5)
	(2, 71.2)
	(3, 76.2)
	(4, 77.5)
	(5, 78.6)
	(6, 77.9)
	};
	\addplot[color=black,mark=*] coordinates {
	(1, 66.9)
	(2, 72.0)
	(3, 78.5)
	(4, 79.2)
	(5, 79.5)
	(6, 79.5)
	};
	\addplot[color=orange,mark=diamond*] coordinates {
	(1, 67.5)
	(2, 72)
	(3, 66)
	(4, 71.1)
	(5, 73)
	(6, 73.2)
	};
	\addplot[color=red,mark=triangle*] coordinates {
	(1, 65.2)
	(2, 59.8)
	(3, 64.3)
	(4, 67)
	(5, 69)
	(6, 72.5)
	};
	\addplot[color=green!60!black,mark=x] coordinates {
	(1, 64.5)
	(2, 63.5)
	(3, 64)
	(4, 68)
	(5, 67.8)
	(6, 66.5)
	};
	\end{axis}
\end{tikzpicture}
\begin{tikzpicture}
	\begin{axis}[
		grid=major,
		xlabel=Training Documents (en),
    		xtick={1,2,3,4,5,6},
	    xticklabels={$100$,$200$,$500$,$1000$,$5000$,$10$k},
		scale only axis,
		xmin=1,xmax=6,
    ytick={40,50,60,70,80,90},
		width=0.4\textwidth,
		height=0.27\textwidth,
		]
	\addplot[color=blue,mark=*] coordinates {
	(1, 83.6)
	(2, 86.2)
	(3, 86.2)
	(4, 87.7)
	(5, 88.7)
	(6, 88.7)
	};
	\addplot[color=black,mark=*] coordinates {
	(1, 83.1)
	(2, 86.8)
	(3, 87.4)
	(4, 88.1)
	(5, 89.0)
	(6, 87.7)
	};
	\addplot[color=orange,mark=diamond*] coordinates {
	(1, 81)
	(2, 78.5)
	(3, 78.5)
	(4, 77.6)
	(5, 80)
	(6, 80.5)
	};
	\addplot[color=red,mark=triangle*] coordinates {
	(1, 49)
	(2, 74)
	(3, 70)
	(4, 68)
	(5, 77.5)
	(6, 76)
	};
	\addplot[color=green!60!black,mark=x] coordinates {
	(1, 45.5)
	(2, 69.5)
	(3, 70)
	(4, 65.5)
	(5, 73.5)
	(6, 70.7)
	};
	\end{axis}
\end{tikzpicture}
\vspace{-0.5em}
\begin{center}
\ref{sharedlegend}
\end{center}
\vspace{-0.5em}
\caption[Classification accuracies over training data]{Classification accuracy for a number of models (see Table
    \ref{tab:results1k} for model descriptions). The left chart shows results
  for these models when trained on German data and evaluated on English data,
  the right chart vice versa.}\label{fig:cldccharts}
\end{figure*}
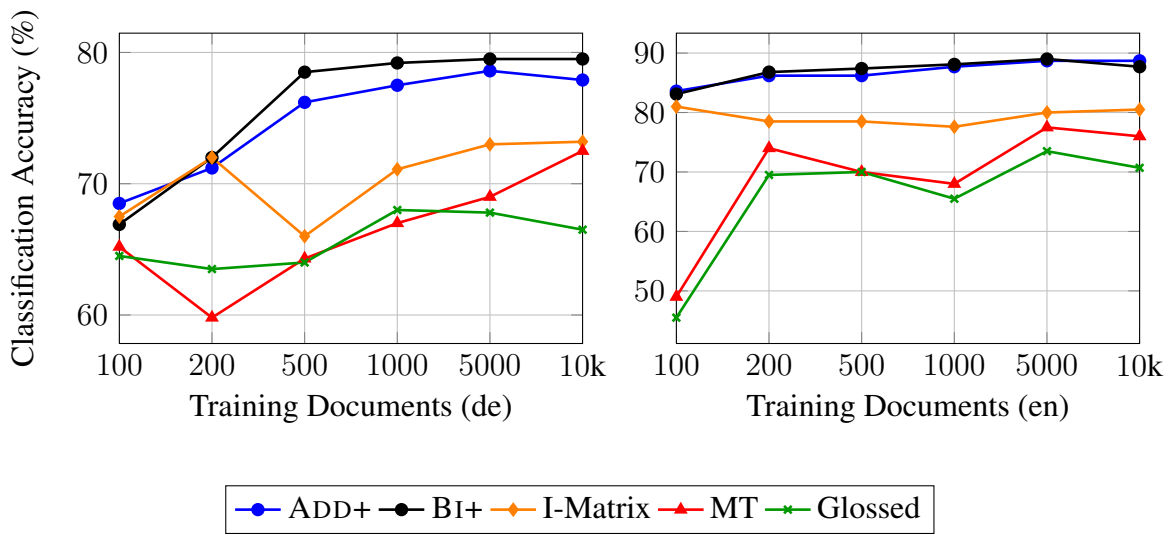

The actual CLDC experiments are performed by training on English and testing on
German documents and vice versa. Following prior work, we use varying sizes
between 100 and 10,000 documents when training the multiclass classifier. The
results of this task across training sizes are in Figure \ref{fig:cldccharts}.
Table \ref{tab:results1k} shows the results for training on 1,000 documents
compared with the results published in \newcite{Klementiev:2012}. Our models
outperform the prior state of the art, with the \flatMod models performing
slightly better than the \addMod models. As the relative results indicate, the
addition of a second language improves model performance. It it interesting to
note that results improve in both directions of the task, even though no
additional German data was used for the `+` models.

\subsection{TED Corpus Experiments}\label{sec:ted-cldc}

Here we describe our experiments on the TED corpus, which enables us to scale up
to multilingual learning. Consisting of a large number of relatively short and
parallel documents, this corpus allows us to evaluate the performance of
the \docMod model described in \S\ref{sec:docmod}.

We use the training data of the corpus to learn distributed representations
across 12 languages. Training is performed in two settings. In the \single
mode, vectors are learnt from a single language pair (en-X), while in the \joint
mode vector-learning is performed on all parallel sub-corpora simultaneously.
This setting causes words from all languages to be embedded in a single semantic
space.

\begin{table*}\centering\footnotesize
\begin{tabular}{@{}lr@{\hspace{0.4em}}r@{\hspace{0.4em}}r@{\hspace{0.4em}}r@{\hspace{0.4em}}r@{\hspace{0.4em}}r@{\hspace{0.4em}}r@{\hspace{0.4em}}r@{\hspace{0.4em}}r@{\hspace{0.4em}}r@{\hspace{0.4em}}r@{}}\toprule
Setting & \multicolumn{11}{c}{Languages}\\ \cmidrule{2-12}
& Arabic & German & Spanish & French & Italian & Dutch & Polish & Pt-Br & Rom'n & Russian & Turkish  \\
\midrule
\multicolumn{12}{@{}l}{$en\rightarrow \text{L2}$} \\
MT System &
\textbf{0.429} & \textbf{0.465} & \textbf{0.518} & \textbf{0.526} & \textbf{0.514} & \textbf{0.505} & \textbf{0.445} & \textbf{0.470} & \textbf{0.493} & 0.432 & 0.409 \\
\addMod \single &
0.328 & 0.343 & 0.401 & 0.275 & 0.282 & 0.317 & 0.141 & 0.227 & 0.282 & 0.338 & 0.241 \\
\flatMod \single &
0.375 & 0.360 & 0.379 & 0.431 & 0.465 & 0.421 & \underline{0.435} & 0.329 &
0.426 & 0.423 & \textbf{\underline{0.481}} \\
\docModadd \single &
\underline{0.410} & 0.424 & 0.383 & \underline{0.476} & \underline{0.485} & 0.264 & 0.402 & 0.354 & 0.418 & 0.448 & 0.452 \\
\docModflat \single &
0.389 & \underline{0.428} & 0.416 & 0.445 & 0.473 & 0.219 & 0.403 & 0.400 & \underline{0.467} & 0.421 & 0.457 \\
\docModadd \joint &
0.392 & 0.405 & 0.443 & 0.447 & 0.475 & \underline{0.453} & 0.394 &
\underline{0.409} & 0.446 & \textbf{\underline{0.476}} & 0.417 \\
\docModflat \joint &
0.372 & 0.369 & \underline{0.451} & 0.429 & 0.404 & 0.433 & 0.417 & 0.399 & 0.453 & 0.439 & 0.418 \\
\midrule
\multicolumn{12}{@{}l}{$\text{L2}\rightarrow en$} \\
MT System &
0.448 & 0.469 & \textbf{0.486} & 0.358 & \textbf{0.481} & 0.463 & \textbf{0.460} & 0.374 & \textbf{0.486} & 0.404 & 0.441 \\
\addMod \single &
0.380 & 0.337 & \underline{0.446} & 0.293 & 0.357 & 0.295 & 0.327 & 0.235 & 0.293 & 0.355 & 0.375 \\
\flatMod \single &
0.354 & 0.411 & 0.344 & 0.426 & 0.439 & 0.428 & \underline{0.443} &
0.357 & 0.426 & 0.442 & 0.403 \\
\docModadd \single &
\textbf{\underline{0.452}} & \textbf{\underline{0.476}} & 0.422 & 0.464 & \underline{0.461} & 0.251 & 0.400 & 0.338 & 0.407 & \underline{\textbf{0.471}} & 0.435 \\
\docModflat \single &
0.406 & 0.442 & 0.365 & \underline{\textbf{0.479}} & 0.460 & 0.235 & 0.393 & 0.380 & 0.426 & 0.467 & \underline{\textbf{0.477}} \\
\docModadd \joint &
0.396 & 0.388 & 0.399 & 0.415 & \underline{0.461} & \underline{\textbf{0.478}} &
0.352 & \textbf{\underline{0.399}} & 0.412 & 0.343 & 0.343 \\
\docModflat \joint &
0.343 & 0.375 & 0.369 & 0.419 & 0.398 & 0.438 & 0.353 & 0.391 &
\underline{0.430} & 0.375 & 0.388 \\
\bottomrule
\end{tabular}
\caption[F1-scores for TED cross-lingual document classification]{F1-scores for the TED document classification task for individual
  languages. Results are reported for both directions (training on English,
    evaluating on L2 and vice versa). Bold indicates best result, underline best
  result amongst the vector-based systems.}\label{tab:exp-beta}
\end{table*}

First, we evaluate the effect of the document-level error signal (\docMod,
  described in \S\ref{sec:docmod}), as well as whether our multilingual learning
method can extend to a larger variety of languages. We train \docMod models,
using both \addMod and \flatMod as \CVM (\docModadd, \docModflat), both in the
\single and \joint mode. For comparison, we also train \addMod and \docMod
models without the document-level error signal.  The resulting document-level
representations are used to train classifiers (system and settings as in
  \S\ref{sec:rcv-cldc}) for each language, which are then evaluated in the
paired language. In the English case we train twelve individual classifiers,
each using the training data of a single language pair only.  As described in
\S\ref{sec:corpus}, we use 15 keywords for the classification task. We report
cumulative results in the form of F1-scores, which are more insightful than
individual scores on a keyword by keyword basis.

\vspace{0.05in}
\noindent\textbf{MT System}\hspace{0.1in}
We develop a machine translation baseline as follows.  We train a
machine translation tool on the parallel training data, using the development
data of each language pair to optimize the translation system.  We use the cdec
decoder \cite{Dyer:2010} with default settings for this purpose.  With this
system we translate the test data, and then use a Na\"{i}ve Bayes
classifier\footnote{We use the implementation in Mallet \cite{McCallum:2002}}
for the actual experiments.  To exemplify, this means the $de{\to}ar$ result is
produced by training a translation system from Arabic to German. The Arabic
test set is translated into German. A classifier is then trained on the German
training data and evaluated on the translated Arabic. While we developed this
system as a baseline, it must be noted that the classifier of this system has
access to significantly more information (all words in the document) as opposed
to our models (one embedding per document), and we do not expect to necessarily
beat this system.

The results of this experiment are in Table \ref{tab:exp-beta}.  When comparing
the results between the \addMod model and the models trained using the
document-level error signal, the benefit of this additional signal becomes
clear. The \joint training mode leads to a relative improvement when training
on English data and evaluating in a second language.  This suggests that the
\joint mode improves the quality of the English embeddings more than it affects
the L2-embeddings. More surprising, perhaps, is the relative performance
between the \addMod and \flatMod composition functions, especially when compared
to the results in \S\ref{sec:rcv-cldc}, where the \flatMod models relatively
consistently performed better. We suspect that the better performance of the
additive composition function on this task is related to the smaller amount of
training data available which could cause sparsity issues for the bigram model.

As expected, the MT system slightly outperforms our models on most language
pairs. However, the overall performance of the models is comparable to that of
the MT system. Considering the relative amount of information available during
the classifier training phase, this indicates that our learned representations
are semantically useful, capturing almost the same amount of information as
available to the Na\"{i}ve Bayes classifier.

 \begin{table*}\centering\footnotesize
\begin{tabular}{@{}lr@{\hspace{0.4em}}r@{\hspace{0.4em}}r@{\hspace{0.4em}}r@{\hspace{0.4em}}r@{\hspace{0.4em}}r@{\hspace{0.4em}}r@{\hspace{0.4em}}r@{\hspace{0.4em}}r@{\hspace{0.4em}}r@{\hspace{0.4em}}r@{}}\toprule
\multirow{2}{*}{\begin{tabular}[l]{@{}l@{}}Training\\Language\end{tabular}} &
  \multicolumn{11}{c}{Test Language}\\ \cmidrule{2-12}
& Arabic & German & Spanish & French & Italian & Dutch & Polish & ~~Pt-Br & ~Rom'n & Russian & Turkish  \\
\midrule
Arabic     &       & 0.378 & 0.436 & 0.432 & 0.444 & 0.438 & 0.389 & 0.425 & 0.420 & 0.446 & 0.397 \\
German     & 0.368 &       & 0.474 & 0.460 & 0.464 & 0.440 & 0.375 & 0.417 & 0.447 & 0.458 & 0.443 \\
Spanish    & 0.353 & 0.355 &       & 0.420 & 0.439 & 0.435 & 0.415 & 0.390 & 0.424 & 0.427 & 0.382 \\
French     & 0.383 & 0.366 & 0.487 &       & 0.474 & 0.429 & 0.403 & 0.418 & 0.458 & 0.415 & 0.398 \\
Italian    & 0.398 & 0.405 & 0.461 & 0.466 &       & 0.393 & 0.339 & 0.347 & 0.376 & 0.382 & 0.352 \\
Dutch      & 0.377 & 0.354 & 0.463 & 0.464 & 0.460 &       & 0.405 & 0.386 & 0.415 & 0.407 & 0.395 \\
Polish     & 0.359 & 0.386 & 0.449 & 0.444 & 0.430 & 0.441 &       & 0.401 & 0.434 & 0.398 & 0.408 \\
Portuguese & 0.391 & 0.392 & 0.476 & 0.447 & 0.486 & 0.458 & 0.403 &       & 0.457 & 0.431 & 0.431 \\
Romanian   & 0.416 & 0.320 & 0.473 & 0.476 & 0.460 & 0.434 & 0.416 & 0.433 &       & 0.444 & 0.402 \\
Russian    & 0.372 & 0.352 & 0.492 & 0.427 & 0.438 & 0.452 & 0.430 & 0.419 & 0.441 &       & 0.447 \\
Turkish    & 0.376 & 0.352 & 0.479 & 0.433 & 0.427 & 0.423 & 0.439 & 0.367 & 0.434 & 0.411 &       \\
\bottomrule
\end{tabular}
\caption[F1-scores for pivoted TED cross-lingual document classification]{F1-scores for TED corpus document classification results when training
  and testing on two languages that do not share any parallel data. We train a
  \docModadd model on all $en$-L2 language pairs together, and then use the
  resulting embeddings to train document classifiers in each language. These
  classifiers are subsequently used to classify data from all other
  languages.}\label{tab:exp-gamma}
\end{table*}

We next investigate linguistic transfer across languages.  We re-use
the embeddings learned with the \docModadd \joint model from the previous
experiment for this purpose, and train classifiers on all non-English languages
using those embeddings. Subsequently, we evaluate their performance in
classifying documents in the remaining languages.  Results for this task are in
Table \ref{tab:exp-gamma}. While the results across language-pairs might not be
very insightful, the overall good performance compared with the results in Table
\ref{tab:exp-beta} implies that we learnt semantically meaningful vectors and in
fact a joint embedding space across thirteen languages.

\begin{table*}\centering\footnotesize
\begin{tabular}{@{}lr@{\hspace{0.5em}}r@{\hspace{0.5em}}r@{\hspace{0.5em}}r@{\hspace{0.5em}}r@{\hspace{0.5em}}r@{\hspace{0.5em}}r@{\hspace{0.5em}}r@{\hspace{0.5em}}r@{\hspace{0.5em}}r@{\hspace{0.5em}}r@{\hspace{0.5em}}r@{}}\toprule
Setting & \multicolumn{12}{c}{Languages}\\ \cmidrule{2-13}
& English & Arabic & German & Spanish & French & Italian & Dutch & Polish &
    Pt-Br & Rom. & Russ. & Turk.  \\
\midrule
Na\"{i}ve-B
& 0.481 & 0.469 & 0.471 & 0.526 & 0.532 & 0.524 & 0.522 & 0.415 & 0.465 & 0.509 & 0.465 & 0.513 \\
\midrule
Senna	& 0.400 & & & & & & & & & & & \\
Polyglot
& 0.382 & 0.416 & 0.270 & 0.418 & 0.361 & 0.332 & 0.228 & 0.323 & 0.194 & 0.300 & 0.402 & 0.295 \\
\midrule
\single Mode \\
\docModadd &
0.462 & 0.422 & 0.429 & 0.394 & 0.481 & 0.458 & 0.252 & 0.385 & 0.363 & 0.431 & 0.471 & 0.435 \\
\docModflat &
0.474 & 0.432 & 0.362 & 0.336 & 0.444 & 0.469 & 0.197 & 0.414 & 0.395 & 0.445 & 0.436 & 0.428 \\
\joint Mode \\
\docModadd &
0.475 & 0.371 & 0.386 & 0.472 & 0.451 & 0.398 & 0.439 & 0.304 & 0.394 & 0.453 & 0.402 & 0.441 \\
\docModflat &
0.378 & 0.329 & 0.358 & 0.472 & 0.454 & 0.399 & 0.409 & 0.340 & 0.431 & 0.379 & 0.395 & 0.435 \\
\bottomrule
\end{tabular}
\caption[F1-scores for TED monolingual document classification]{F1-scores on the TED corpus document classification task when training
  and evaluating on the same language. Baseline embeddings are Senna
  \protect\cite{Collobert:2011} and Polyglot \protect\cite{Al-Rfou:2013}.}\label{tab:exp-delta}
\end{table*}

In a third evaluation (Table \ref{tab:exp-delta}), we apply the embeddings
learnt with out models to a monolingual classification task, enabling us to
compare with prior work on distributed representation learning.  In this
experiment a classifier is trained in one language and then evaluated in the
same. We again use a Na\"{i}ve Bayes classifier on the raw data to establish a
reasonable upper bound.

We compare our embeddings with the SENNA embeddings, which achieve state of the
art performance on a number of tasks \cite{Collobert:2011}. Additionally, we
use the Polyglot embeddings of \newcite{Al-Rfou:2013}, who published word
embeddings across 100 languages, including all languages considered in our work.
We represent each document by the mean of its word vectors and then
apply the same classifier training and testing regime as with our models. Even
though both of these sets of embeddings were trained on much larger datasets
than ours, our models outperform these baselines on all languages---even
outperforming the Na\"{i}ve Bayes system on several languages. While this
may partly be attributed to the fact that our vectors were learned on in-domain
data, this is still a very positive outcome.

\subsection{Linguistic Analysis}\label{sec:qualitative}

\begin{figure}[t]\centering
\includegraphics[scale=0.55, frame]{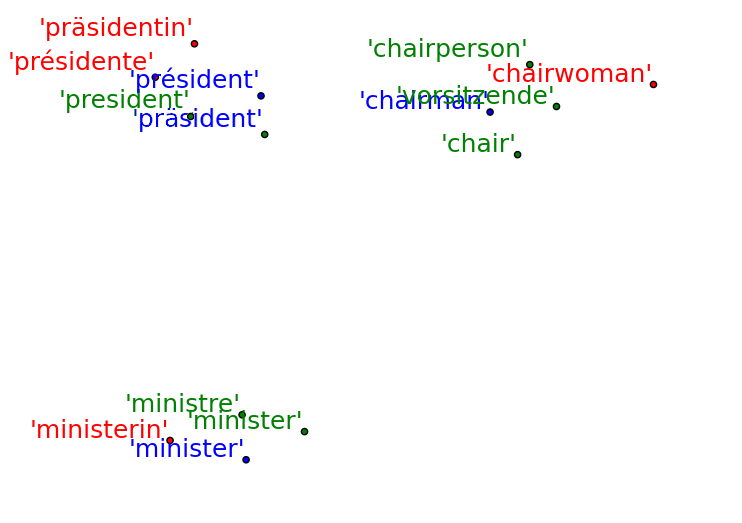}
\caption[t-SNE proejctions for \flatModplus word representations]{t-SNE projections for a number of English, French and German words as
  represented by the \flatModplus model. Even though the model did not use any
  parallel French-German data during training, it learns semantic similarity
  between these two languages using English as a pivot, and semantically
  clusters words across all languages.}\label{fig:words}
\end{figure}

\begin{figure}[t]\centering
\includegraphics[scale=0.55, frame]{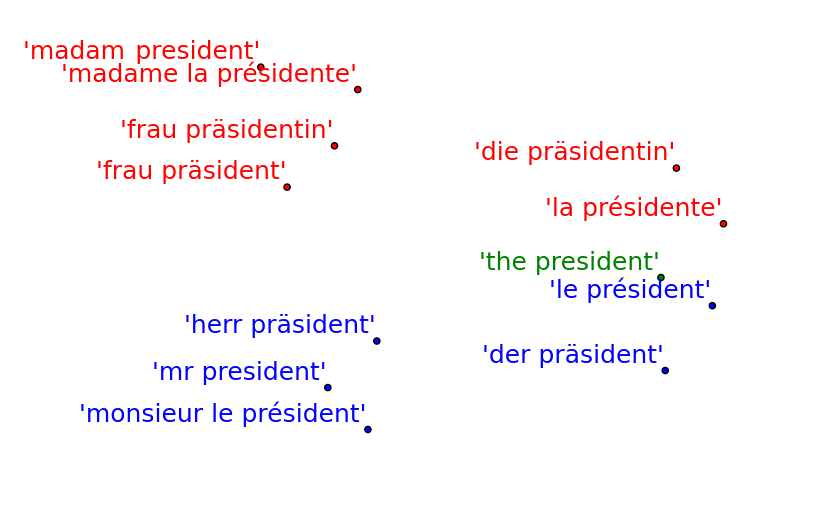}
\caption[t-SNE projections for \flatModplus phrase representations]{t-SNE projections for a number of short phrases in three languages as
  represented by the \flatModplus model. The projection demonstrates linguistic
  transfer through a pivot by. It separates phrases by gender (red for female,
    blue for male, and green for neutral) and aligns matching phrases across
  languages.}\label{fig:gender}
\end{figure}

While the classification experiments focused on establishing the semantic
content of the sentence level representations, we also want to briefly
investigate the induced word embeddings.  We use the \flatModplus model trained
on the Europarl corpus for this purpose.  Figure \ref{fig:words} shows the t-SNE
projections for a number of English, French and German words.  Even though the
model did not use any parallel French-German data during training, it still
managed to learn semantic word-word similarity across these two languages.

Going one step further, Figure \ref{fig:gender} shows t-SNE projections for a
number of short phrases in these three languages.  We use the English
\textit{the president} and gender-specific expressions \textit{Mr President} and
\textit{Madam President} as well as gender-specific equivalents in French and
German.  The projection demonstrates a number of interesting results: First, the
model correctly clusters the words into three groups, corresponding to the three
English forms and their associated translations.  Second, a separation between
genders can be observed, with male forms on the bottom half of the chart and
female forms on the top, with the neutral \textit{the president} in the vertical
middle.  Finally, if we assume a horizontal line going through \textit{the
  president}, this line could be interpreted as a ``gender divide'', with male
and female versions of one expression mirroring each other on that line.  In the
case of \textit{the president} and its translations, this effect becomes even
clearer, with the neutral English expression being projected close to the
mid-point between each other language's gender-specific versions.

These results further support our hypothesis that the bilingual contrastive
error function can learn semantically plausible embeddings and furthermore, that
it can abstract away from mono-lingual surface realisations into a shared
semantic space across languages.

\section{Related Work}

As introduced in Chapters \ref{chapter:distrib} and \ref{chapter:compositional},
most research on distributed representation induction has focused on single
languages. English, with its large number of annotated resources, has enjoyed
most attention.  However, there exists a body of prior work on learning
multilingual embeddings or on using parallel data to transfer linguistic
information across languages.  One has to differentiate between approaches such
as \newcite{Al-Rfou:2013}, that learn embeddings across a large variety of
languages, and models such as ours, that learn joint embeddings, that is a
projection into a shared semantic space across multiple languages.

Related to our work, \newcite{Yih:2011} proposed S2Nets to learn joint
embeddings of tf-idf vectors for comparable documents. Their architecture
optimises the cosine similarity of documents, using relative semantic similarity
scores during learning.  More recently, \newcite{Lauly:2013} proposed a
bag-of-words autoencoder model, where the bag-of-words representation in one
language is used to train the embeddings in another. By placing their vocabulary
in a binary branching tree, the probabilistic setup of this model is similar to
that of \newcite{Mnih:2009}. Similarly, \newcite{SarathChandar:2013} train a
cross-lingual encoder, where an autoencoder is used to recreate words in two
languages in parallel. This is effectively the linguistic extension of
\newcite{Ngiam:2011}, who used a similar method for audio and video data.

\newcite{Klementiev:2012}, our baseline in \S\ref{sec:rcv-cldc}, use a form of
multi-agent learning on word-aligned parallel data to transfer embeddings from
one language to another.
Earlier work, \newcite{Haghighi:2008}, proposed a method for inducing bilingual
lexica using monolingual feature representations and a small initial lexicon to
bootstrap with.  This approach has recently been extended by
\newcite{Mikolov:2013}, \newcite{Mikolov:2013a}, who developed a method for learning
transformation matrices to convert semantic vectors of one language into those
of another.  Is was demonstrated that this approach can be applied to improve
tasks related to machine translation.  Their CBOW model is also worth noting for
its similarities to the \addMod composition function used here.  Using a
slightly different approach, \newcite{Zou:2013}, also learned bilingual embeddings
for machine translation.

\section{Summary}

In this chapter we have presented a novel method for learning multilingual word
embeddings using parallel data in conjunction with a multilingual objective
function for compositional vector models. This approach extends the
distributional hypothesis to multilingual joint-space representations. Coupled
with very simple composition functions, vectors learned with this method
outperform the state of the art on the task of cross-lingual document
classification. Further experiments and analysis support our hypothesis that
bilingual signals are a useful tool for learning distributed representations by
enabling models to abstract away from mono-lingual surface realisations into a
deeper semantic space.

After extending our approach to include multilingual training data, we were able
to demonstrate that adding additional languages further improves the model.
Furthermore, using some qualitative experiments and visualizations, we showed
that our approach also allows us to learn semantically related embeddings across
languages without any direct training data.

Overall, the work presented in this chapter adds further support to the
hypothesis described in the introduction to this thesis. In conjunction with the
work presented in Chapter \ref{chapter:syntax}, we demonstrated that distributed
semantic representations are useful beyond the word level and have proposed
multiple approaches for learning such representations.
The multilingual models discussed in this chapter are of particular interest, as
they highlight the possibility for learning semantically plausible
representations both at the word and sentence level while at the same time
minimising the impact of language or syntax specific biases.

\part{Conclusions and Further Work}\label{part:concl}
\chapter{Further Research}\label{sec:concl:future}

There are many avenues for future work related to the research presented in this
thesis, both related to the application of distributed representations to new
tasks and to the further development of models for semantic composition. These
research opportunities include:

\begin{enumerate}
  \item In this thesis we have demonstrated the efficacy of distributed
    representations in solving semantically ambitious tasks such as
    semantic frame identification. Within the frame-semantic parsing pipeline,
    an obvious extension would be to attempt to also solve the semantic role
    labelling task using distributed information.
  \item Another extension to the experimental support of our thesis in Chapter
    \ref{chapter:frame-semantic} is to attempt to integrate this method with the
    fully compositional approaches presented in the subsequent chapters of this
    thesis---that is to use a more complex syntactic composition function to
    learn the context instance representations required for the frame
    identification stage.
  \item In Chapter \ref{chapter:syntax} we investigated a number of composition
    models that exploit the CCG framework to guide their composition steps.
    While we relied on vector-based representations in this approach and solely
    used non-linearities to enforce interdependence between arguments, there are
    alternative proposals that could be evaluated in this direction. A direction
    of particular interest here are tensor-models, where the categorial type of
    a linguistic unit would guide its shape. While there are some issues with
    the scalability and complexity of such models (see \S\ref{sec:comp:vsm}),
    recent work has made some advances in this direction by proposing
    approximations that might overcome these issues
    \cite{Grefenstette:2013a,Clark:2013,Maillard:2014}.
  \item The approach presented in Chapter \ref{chapter:multilingual} for
    learning semantically motivated multilingual distributed representations
    also invites further research. Several avenues of future work are available
    here. One possibility is to attempt to reconcile this multilingual objective
    function with more complex composition functions. While some initial
    research into this question has not borne any fruit, this deserves further
    attention. A likely explanation for the current lack of success regarding
    this problem is in the limited amount of training data available, and thus
    this problem should be revisited when we have access to larger parallel
    corpora.
  \item Another, related idea is to attempt to inverse the process described in
    most of the models presented in this thesis. This thesis focuses on learning
    representations given words. Similarly to the work in language modelling
    \cite[\textit{inter alia}]{Mikolov:2010,Mnih:2009}, it would be possible to
    imagine reversing the process and to generate words from compositional
    representations. Initial work in this direction was discussed in
    \cite{Kalchbrenner:2013,Philtalk:2013}
\end{enumerate}

\chapter{Conclusions}\label{chapter:conclusions}

Throughout this thesis we have analysed the construction, the use, capabilities
and limitations and other properties of distributed representations for
semantics.
The primary aim of this thesis was to investigate the use of distributed
representations for capturing semantics, and to evaluate their efficacy in
solving tasks in NLP for which a degree of semantic understanding could be
beneficial. At the outset in Chapter \ref{chapter:intro} we stated the
underlying hypothesis of this thesis.

\begin{aquote}{\S\ref{sec:intro:thesis}}
Our hypothesis is that distributed representations are a highly
suitable mechanism for capturing and manipulating semantics, and further, that
meaning both at the word level and beyond can be encoded distributionally.
\end{aquote}

The analysis and empirical evidence collected throughout this thesis strongly
supports our original hypothesis. Considering the first aspect on whether
distributed representations are suitable for capturing and manipulating
semantic information, we showed in Chapter \ref{chapter:frame-semantic} that
generic distributional representations can be used to outperform the state of
the art on semantically challenging tasks such as semantic frame identification.
Throughout the following chapters we repeatedly introduced simple models that
learn or exploit distributed representations and that with high consistency
outperformed more complex models reliant on symbolic reasoning or purely
frequency-based or syntactic methods.

Concerning the second aspect of our hypothesis, namely that distributed
representations are not restricted to encoding semantics at the word level, we
also evaluated and supported this claim through thorough investigation,
particularly in the second part of this thesis. In Chapter
\ref{chapter:compositional} we provided an overview of various theoretical
foundations for the composition of distributed representations into higher order
structures. Subsequently in Chapter \ref{chapter:syntax} we investigated the
role of syntax in semantic vector composition. Here we supported our hypothesis
by highlighting prior work on learning compositional semantic representations
and further by developing several new composition models that learned
semantically plausible embeddings for sentences as highlighted by their
performance on various sentiment analysis tasks.
Not content purely with the analysis of semantic representations learned through
task-specific objective functions, we investigated how to learn semantic
representations with task-independent distributions. The multilingual objective
function developed in Chapter \ref{chapter:multilingual} removes task-specific
biases and minimises the impact of monolingual surface forms on learning
distributed representations.

Combining the conclusions drawn in the earlier chapters of this thesis, the
model developed in Chapter \ref{chapter:multilingual} unifies both aspects of
our hypothesis into a single model. The models learn representations both for
words and for sentences. Furthermore, these representations are learned using a
multilingual extension of Firth's distributional hypothesis, which we argued
should result in convincing semantic representations. Put to the test on
multiple cross-lingual document classification tasks our models---again simpler
compared with a number of our baseline models---outperformed the prior state of
the art once more.

On the basis of the analysis and the experimental support provided in this
thesis, we argue that the work presented here strongly supports the hypothesis
we set out to investigate.

Looking forward, the work presented in this thesis opens up new questions.
Having established the usefulness of distributed representations for capturing
semantics, such new questions are how far this approach can be pushed, what
limitations there are to the use of distributed representations, and what
problems can or cannot be solved with this mechanism.
So far most work on distributed semantic representations focuses on learning
representations given natural language input, as well as various classification
and ranking tasks based on these representations.
One of the big challenges in the future will be to reverse this process, that is
to develop algorithms for generating natural language given composed distributed
representations. Work on neural language modelling is only a first step in that
direction. Finding efficient and effective models for this task is a key
requirement for the application of distributed representations to complex NLP
tasks such as question answering and machine translation.
When considering such tasks, additional questions come to mind that require
further research. One such question is whether distributed representations are
sufficiently expressive to encode complex meaning as required for logical or
semantic inference. A related question is whether distributed
representations will suffice on their own, or whether dual approaches that
combine distributed and symbolic representations of meaning will be more
suitable for such tasks.
There has been some initial work on these questions, but this will remain
an exciting area of research for many years to come.

\appendix

\chapter{Semantic-Frame Parsing: Argument Identification}\label{appendix:fs:argid}

 \begin{table}[t]
  \centering
  \begin{tabular}{p{2.3in}p{2.3in}}
    \toprule
    $-$ starting word of $a$ & $-$ POS of the starting word of $a$\\
    $-$ ending word of $a$ &  $-$ POS of the ending word of $a$\\
    $-$  head word of $a$ & $-$ POS of the head word of $a$\\
    $-$  bag of words in $a$ & $-$ bag of POS tags in $a$\\
    \multicolumn{2}{l}{$-$ a bias feature}\\
    \multicolumn{2}{l}{$-$ voice of the predicate use}\\
    \multicolumn{2}{l}{$-$ word cluster of $a$'s head }\\
    \multicolumn{2}{l}{$-$ word cluster of $a$'s head conjoined with word
      cluster of the predicate$^{*}$}\\
    \multicolumn{2}{l}{$-$ dependency path between $a$'s head and the
      predicate}\\
    \multicolumn{2}{p{4.6in}}{$-$ the set of dependency labels of the
      predicate's children}\\
    \multicolumn{2}{p{4.6in}}{$-$ dependency path conjoined with the POS tag
      of $a$'s head}\\
    \multicolumn{2}{p{4.6in}}{$-$ dependency path conjoined with the word
      cluster of $a$'s head}\\
    \multicolumn{2}{p{4.6in}}{$-$ position of $a$ with respect to the
      predicate (\textit{before, after, overlap or identical})}\\
    \multicolumn{2}{p{4.6in}}{$-$ whether the subject of the predicate is
      missing (\textit{missingsubj})}\\
    \multicolumn{2}{p{4.6in}}{$-$ \textit{missingsubj}, conjoined with the
      dependency path}\\
    \multicolumn{2}{p{4.6in}}{$-$ \textit{missingsubj}, conjoined with the
      dependency path from the verb dominating the predicate to $a$'s head}\\
    \bottomrule
  \end{tabular}
  \caption[SRL argument identification features]{Argument identification features. The span in consideration is termed
    $a$. Every feature in this list has two versions, one conjoined with the
    given role $r$ and the other conjoined with both $r$ and the frame $y$. The
    feature with a $^{*}$ superscript is only conjoined with the role to reduce
    its sparsity.}\label{tab:argid-features}
\end{table}

We briefly describe the argument identification model used in our frame-semantic
parsing experiments in Chapter \ref{chapter:frame-semantic}. This
model is used after the frame identification step to find arguments to fill the
semantic roles of a given frame. The model is based on existing work in this
field \cite{Xue:2004,Das:2010} and contains only a small number of
modifications, as its primary purpose is to allow us to evaluate our frame
identification model in the context of a full-scale frame-semantic parsing task.

The implementation of the argument identification system, as well as the
modifications explained here and later in \S\ref{sec:fs:exp:fn:argid} are the
work of my co-authors in \newcite{Hermann:2014:ACLgoogle}. They are included in
this thesis as they are essential for understanding the experiments discussed in
this chapter, but should not be considered part of the contribution of this
thesis.

Given $x$, the sentence with a marked predicate, the argument identification
model assumes that the predicate frame $y$ has been disambiguated. From a frame
lexicon, we look up the set of semantic roles $\mathcal{R}_y$ that associate
with $y$.  This set also contains the null role $r_{\emptyset}$. From $x$, a
rule-based candidate argument extraction algorithm then extracts a set of spans
$\mathcal{A}$ that could potentially serve as the overt arguments
$\mathcal{A}_y$ for $y$. By overtness, we mean the non-null instantiation of a
semantic role in a frame-semantic parse.
Details of the specific candidate argument extraction algorithms used vary a
little between the FrameNet and the PropBank case, and are specified in
\S\ref{sec:fs:experiments:framenet}-\S\ref{sec:fs:experiments:propbank}.

\section{Learning and Inference}

Given training data of the form $\langle \langle x^{(i)}, y^{(i)},
\mathcal{M}^{(i)}\rangle \rangle_{i=1}^{N}$, where,
\begin{equation}
  \mathcal{M} = \{(r, a) : r \in \mathcal{R}_y, a \in \mathcal{A} \cup \mathcal{A}_y\},
\end{equation}
is a set of tuples that associates each role $r$ in $\mathcal{R}_y$ with a span
$a$ according to the gold data. Note that this mapping associates spans with
the null role $r_{\emptyset}$ as well. We optimize the following log-likelihood
to train our model:
\begin{equation}
\max_{\Theta} \displaystyle \sum_{i = 1}^N \sum_{j = 1}^{\mid \mathcal{M}^{(i)}\mid} \log p_{\Theta} \big((r, a)_{j} | x, y, \mathcal{R}_y\big) - C \|\Theta\|^{2}_2\nonumber
\end{equation}
where $p_{\Theta}$ is a log-linear model normalized over the set
$\mathcal{R}_y$, with the features described in Table~\ref{tab:argid-features},
and the following resultant formulation:
\begin{equation}
\hspace{-0.1in}p_{\Theta} \big((r, a) | x, y, \mathcal{R}_y\big) =
\frac{\exp \Theta \cdot \mathbf{h}(r, a, y, x)}{\displaystyle\sum_{\bar{r} \in \mathcal{R}_y} \exp \Theta \cdot \mathbf{h}(\bar{r}, a, y, x)}.\nonumber
\end{equation}
Above, $\Theta$ are the model parameters and $\mathbf{h}$ is a feature function that uses the features
from Table~\ref{tab:argid-features}; we describe the nature of the discrete
word clusters that are used
in the feature set in \S\ref{sec:fs:experiments:common}. We train this model
using L-BFGS \cite{Liu:1989} and set $C$ to 1.0.

Although our learning mechanism uses a local log-linear model, we perform
inference globally on a per-frame basis by applying hard structural constraints.
Following \newcite{Das:2014} and \newcite{Punyakanok:2008} we use the
log-probability of the local classifiers as a score in an integer linear program
(ILP) to assign roles subject to hard constraints described in
\S\ref{sec:fs:experiments:framenet} and \S\ref{sec:fs:experiments:propbank}.  We
use an off-the-shelf ILP solver for
inference.\footnote{\url{http://scip.zib.de/}}

\chapter{FrameNet Development Data}\label{appendix:fs:dev}

Table \ref{tab:fs:appendix} features a list of the 16 randomly selected
documents from the FrameNet 1.5 corpus, which we used for development in the
frame-semantic parsing task described in Chapter \ref{chapter:frame-semantic}. The
resultant development set consists of roughly 4,500 predicates.
We used the same test set as in \newcite{Das:2014}, containing 23
documents and 4,458 predicates.

\begin{table}[h]
\begin{tabular}{l}
  \toprule
  \textbf{Development Data Filename} \\
  \midrule
  LUCorpus-v0.3\textunderscore \textunderscore 20000420\textunderscore xin\textunderscore eng-NEW.xml \\
  NTI\textunderscore \textunderscore SouthAfrica\textunderscore Introduction.xml \\
  LUCorpus-v0.3\textunderscore \textunderscore CNN\textunderscore AARONBROWN\textunderscore ENG\textunderscore 20051101\textunderscore 215800.partial-NEW.xml \\
  LUCorpus-v0.3\textunderscore \textunderscore AFGP-2002-600045-Trans.xml \\
  PropBank\textunderscore \textunderscore TicketSplitting.xml \\
  Miscellaneous\textunderscore \textunderscore Hijack.xml \\
  LUCorpus-v0.3\textunderscore \textunderscore artb\textunderscore 004\textunderscore A1\textunderscore E1\textunderscore NEW.xml \\
  NTI\textunderscore \textunderscore WMDNews\textunderscore 042106.xml \\
  C-4\textunderscore \textunderscore C-4Text.xml \\
  ANC\textunderscore \textunderscore EntrepreneurAsMadonna.xml \\
  NTI\textunderscore \textunderscore LibyaCountry1.xml \\
  NTI\textunderscore \textunderscore NorthKorea\textunderscore NuclearOverview.xml \\
  LUCorpus-v0.3\textunderscore \textunderscore 20000424\textunderscore nyt-NEW.xml \\
  NTI\textunderscore \textunderscore WMDNews\textunderscore 062606.xml \\
  ANC\textunderscore \textunderscore 110CYL070.xml \\
  LUCorpus-v0.3\textunderscore \textunderscore CNN\textunderscore ENG\textunderscore 20030614\textunderscore 173123.4-NEW-1.xml \\
  \bottomrule
\end{tabular}
\caption[Development data from the FrameNet 1.5 corpus]{List of files used as
  development set for the FrameNet 1.5 corpus.}\label{tab:fs:appendix}
\end{table}

\chapter{CCG Categories for CCAE Models}\label{appendix:syntax:cat}

\begin{table}[h]
  \centering
  \begin{tabular}[t]{@{}lr@{}}
    \toprule
    Category & Frequency \\
    \midrule
N & 516489 \\
NP & 390344 \\
S[dcl] & 209220 \\
N/N & 177776 \\
NP[nb] & 169054 \\
S[dcl]$\backslash$ NP & 162391 \\
NP[nb]/N & 161325 \\
NP$\backslash$ NP & 151296 \\
(NP$\backslash$ NP)/NP & 75891 \\
, & 70680 \\
(S[X]$\backslash$ NP)$\backslash$ (S[X]$\backslash$ NP) & 70562 \\
. & 69379 \\
S[b]$\backslash$ NP & 68185 \\
  \bottomrule
\end{tabular}
\quad\quad\quad
\begin{tabular}[t]{@{}lr@{}}
    \toprule
    Category & Frequency \\
    \midrule
conj & 45861 \\
(S[dcl]$\backslash$ NP)/NP & 45444 \\
S[pss]$\backslash$ NP & 41865 \\
((S$\backslash$ NP)$\backslash$ (S$\backslash$ NP))/NP & 37978 \\
S[adj]$\backslash$ NP & 36821 \\
(S$\backslash$ NP)$\backslash$ (S$\backslash$ NP) & 34632 \\
PP & 31215 \\
S[ng]$\backslash$ NP & 30346 \\
PP/NP & 29738 \\
(S[dcl]$\backslash$ NP)/(S[b]$\backslash$ NP) & 27309 \\
(S[b]$\backslash$ NP)/NP & 27070 \\
(S[to]$\backslash$ NP)/(S[b]$\backslash$ NP) & 22955 \\
  \bottomrule
\end{tabular}
\caption{CCG categories considered in the CCAE models. Frequency denotes the
  frequency of the labels on the British National Corpus dataset used for
  pre-training those models in \S\ref{sec:syntax:sentiment}.}\label{tab:syntax:ccgcats}
\end{table}

\addcontentsline{toc}{chapter}{References}

\bibliography{a,kmh}

\begin{thebibliography}{}

\bibitem[\protect\citename{Al-Rfou' \bgroup et al.\egroup }2013]{Al-Rfou:2013}
R.~Al-Rfou', B.~Perozzi, and S.~Skiena.
\newblock 2013.
\newblock {Polyglot: Distributed Word Representations for Multilingual {NLP}}.
\newblock In {\em Proceedings of the Seventeenth Conference on Computational
  Natural Language Learning}, pages 183--192, Sofia, Bulgaria, August.
  Association for Computational Linguistics.

\bibitem[\protect\citename{Andrews \bgroup et al.\egroup }2009]{Andrews:2009}
M.~Andrews, G.~Vigliocco, and D.~Vinson.
\newblock 2009.
\newblock {Integrating experiential and distributional data to learn semantic
  representations.}
\newblock {\em Psychological Review}, 116(3):463--498, Jul.

\bibitem[\protect\citename{Baker \bgroup et al.\egroup }1998]{Baker:1998}
C.~F. Baker, C.~J. Fillmore, and J.~B. Lowe.
\newblock 1998.
\newblock {The Berkeley FrameNet Project}.
\newblock In {\em Proceedings of the 36th Annual Meeting of the Association for
  Computational Linguistics and 17th International Conference on Computational
  Linguistics - Volume 1}, ACL '98, pages 86--90, Montreal, Quebec, Canada.
  Association for Computational Linguistics.

\bibitem[\protect\citename{Baker \bgroup et al.\egroup }2007]{Baker:2007}
C.~Baker, M.~Ellsworth, and K.~Erk.
\newblock 2007.
\newblock {{SemEval-2007} {T}ask 19: Frame Semantic Structure Extraction}.
\newblock In {\em Proceedings of the Fourth International Workshop on Semantic
  Evaluations (SemEval-2007)}, pages 99--104, Prague, Czech Republic, June.
  Association for Computational Linguistics.

\bibitem[\protect\citename{Bangalore and Joshi}1999]{Bangalore:1999}
S.~Bangalore and A.~K. Joshi.
\newblock 1999.
\newblock {Supertagging: An Approach to Almost Parsing}.
\newblock {\em Computational Linguistics}, 25(2):237--265, June.

\bibitem[\protect\citename{Bannard \bgroup et al.\egroup }2003]{Bannard:2003}
C.~Bannard, T.~Baldwin, and A.~Lascarides.
\newblock 2003.
\newblock {A Statistical Approach to the Semantics of Verb-particles}.
\newblock In {\em Proceedings of the ACL 2003 Workshop on Multiword
  Expressions: Analysis, Acquisition and Treatment - Volume 18}, MWE '03, pages
  65--72, Sapporo, Japan. Association for Computational Linguistics.

\bibitem[\protect\citename{Baroni and Zamparelli}2010]{Baroni:2010}
M.~Baroni and R.~Zamparelli.
\newblock 2010.
\newblock {Nouns Are Vectors, Adjectives Are Matrices: Representing
  Adjective-noun Constructions in Semantic Space}.
\newblock In {\em Proceedings of the 2010 Conference on Empirical Methods in
  Natural Language Processing}, EMNLP '10, pages 1183--1193, Cambridge,
  Massachusetts. Association for Computational Linguistics.

\bibitem[\protect\citename{Basile \bgroup et al.\egroup }2012]{Basile:2012}
V.~Basile, J.~Bos, K.~Evang, and N.~Venhuizen.
\newblock 2012.
\newblock {Developing a large semantically annotated corpus}.
\newblock In {\em Proceedings of the Eight International Conference on Language
  Resources and Evaluation (LREC'12)}, pages 3196--3200, Istanbul, Turkey, may.
  European Language Resources Association (ELRA).

\bibitem[\protect\citename{Bengio \bgroup et al.\egroup }2003]{Bengio:2003}
Y.~Bengio, R.~Ducharme, P.~Vincent, and C.~Jauvin.
\newblock 2003.
\newblock {A Neural Probabilistic Language Model}.
\newblock {\em Journal of Machine Learning Research}, 3:1137--1155, March.

\bibitem[\protect\citename{Bengio \bgroup et al.\egroup }2007]{Bengio:2007}
Y.~Bengio, P.~Lamblin, D.~Popovici, and H.~Larochelle.
\newblock 2007.
\newblock {Greedy Layer-Wise Training of Deep Networks}.
\newblock In B.~Sch\"{o}lkopf, J.~Platt, and T.~Hoffman, editors, {\em Advances
  in Neural Information Processing Systems 19}, pages 153--160. MIT Press,
  Cambridge, MA.

\bibitem[\protect\citename{Bengio}2009]{Bengio:2009}
Y.~Bengio.
\newblock 2009.
\newblock {Learning deep architectures for {AI}}.
\newblock {\em Foundations and Trends in Machine Learning}, 2(1):1--127,
  January.

\bibitem[\protect\citename{Biemann and Giesbrecht}2011]{Biemann:2011}
C.~Biemann and E.~Giesbrecht, editors.
\newblock 2011.
\newblock {\em {Proceedings of the Workshop on Distributional Semantics and
  Compositionality}}.
\newblock Association for Computational Linguistics, Portland, Oregon, USA,
  June.

\bibitem[\protect\citename{Blacoe and Lapata}2012]{Blacoe:2012}
W.~Blacoe and M.~Lapata.
\newblock 2012.
\newblock {A Comparison of Vector-based Representations for Semantic
  Composition}.
\newblock In {\em Proceedings of the 2012 Joint Conference on Empirical Methods
  in Natural Language Processing and Computational Natural Language Learning},
  EMNLP-CoNLL '12, pages 546--556, Jeju Island, Korea. Association for
  Computational Linguistics.

\bibitem[\protect\citename{Blei \bgroup et al.\egroup }2003]{Blei:2003}
D.~M. Blei, A.~Y. Ng, and M.~I. Jordan.
\newblock 2003.
\newblock {Latent dirichlet allocation}.
\newblock {\em Journal of Machine Learning Research}, 3:993--1022, March.

\bibitem[\protect\citename{Bloom}2001]{Bloom:2001}
P.~Bloom.
\newblock 2001.
\newblock {Precis of How Children Learn the Meanings of Words}.
\newblock {\em Behavioral and Brain Sciences}, 24:1095--1103.

\bibitem[\protect\citename{Blunsom \bgroup et al.\egroup }2013]{Philtalk:2013}
P.~Blunsom, K.~M. Hermann, and N.~Kalchbrenner.
\newblock 2013.
\newblock {Compositional Semantics, Deep Learning, and Machine Translation}.
\newblock In {\em Keynote Talk by Phil Blunsom at the 2013 MT Marathon},
  Prague, Czech Republic, September.

\bibitem[\protect\citename{Bottou}2010]{Bottou:2010}
L.~Bottou.
\newblock 2010.
\newblock {Large-Scale Machine Learning with Stochastic Gradient Descent}.
\newblock In Y.~Lechevallier and G.~Saporta, editors, {\em Proceedings of the
  19th International Conference on Computational Statistics (COMPSTAT'2010)},
  pages 177--187, Paris, France, August. Springer.

\bibitem[\protect\citename{Carreras and M\`{a}rquez}2004]{Carreras:2004}
X.~Carreras and L.~M\`{a}rquez.
\newblock 2004.
\newblock {Introduction to the {CoNLL}-2004 Shared Task: Semantic Role
  Labeling}.
\newblock In H.~T. Ng and E.~Riloff, editors, {\em HLT-NAACL 2004 Workshop:
  Eighth Conference on Computational Natural Language Learning (CoNLL-2004)},
  pages 89--97, Boston, Massachusetts, USA, May 6 - May 7. Association for
  Computational Linguistics.

\bibitem[\protect\citename{Carreras and M{\`a}rquez}2005]{Carreras:2005}
X.~Carreras and L.~M{\`a}rquez.
\newblock 2005.
\newblock {Introduction to the {CoNLL}-2005 Shared Task: Semantic Role
  Labeling}.
\newblock In {\em Proceedings of the Ninth Conference on Computational Natural
  Language Learning (CoNLL-2005)}, pages 152--164, Ann Arbor, Michigan, June.
  Association for Computational Linguistics.

\bibitem[\protect\citename{Cettolo \bgroup et al.\egroup }2012]{Cettolo:2012}
M.~Cettolo, C.~Girardi, and M.~Federico.
\newblock 2012.
\newblock {{WIT}$^3$: Web Inventory of Transcribed and Translated Talks}.
\newblock In {\em Proceedings of the 16$^{th}$ Conference of the European
  Association for Machine Translation (EAMT)}, pages 261--268, May.

\bibitem[\protect\citename{Cha}2007]{Cha:2007}
S.-H. Cha.
\newblock 2007.
\newblock {Comprehensive Survey on Distance/Similarity Measures between
  Probability Density Functions}.
\newblock {\em International Journal of Mathematical Models and Methods in
  Applied Sciences}, 1(4):300--307.

\bibitem[\protect\citename{Chiang \bgroup et al.\egroup }2013]{ISI:2013}
D.~Chiang, J.~Andreas, D.~Bauer, K.~M. Hermann, B.~Jones, and K.~Knight.
\newblock 2013.
\newblock {Parsing Graphs with Hyperedge Replacement Grammars}.
\newblock In {\em Proceedings of the 51st Annual Meeting of the Association for
  Computational Linguistics (Volume 1: Long Papers)}, Sofia, Bulgaria, August.
  Association for Computational Linguistics.

\bibitem[\protect\citename{Ciaramita and Johnson}2003]{Ciaramita:2003}
M.~Ciaramita and M.~Johnson.
\newblock 2003.
\newblock {Supersense tagging of unknown nouns in WordNet}.
\newblock In {\em Proceedings of the 2003 conference on Empirical methods in
  natural language processing}, EMNLP '03, pages 168--175, Sapporo, Japan.
  Association for Computational Linguistics.

\bibitem[\protect\citename{Clark and Curran}2007]{Clark:2007}
S.~Clark and J.~R. Curran.
\newblock 2007.
\newblock {Wide-coverage efficient statistical parsing with {CCG} and
  log-linear models}.
\newblock {\em Computational Linguistics}, 33(4):493--552, December.

\bibitem[\protect\citename{Clark and Pulman}2007]{Clark:2007a}
S.~Clark and S.~Pulman.
\newblock 2007.
\newblock {Combining Symbolic and Distributional Models of Meaning}.
\newblock In {\em Proceedings of AAAI Spring Symposium on Quantum Interaction}.
  AAAI Press.

\bibitem[\protect\citename{Clark \bgroup et al.\egroup }2008]{Clark:2008}
S.~Clark, B.~Coecke, and M.~Sadrzadeh.
\newblock 2008.
\newblock {A compositional distributional model of meaning}.
\newblock In {\em Proceedings of the Second Quantum Interaction Symposium
  (QI-2008)}, pages 133--140.

\bibitem[\protect\citename{Clark}2013]{Clark:2013}
S.~Clark.
\newblock 2013.
\newblock {Type-Driven Syntax and Semantics for Composing Meaning Vectors}.
\newblock In C.~Heunen, M.~Sadrzadeh, and E.~Grefenstette, editors, {\em
  Quantum Physics and Linguistics: A Compositional, Diagrammatic Discourse},
  pages 359--377. Oxford University Press.

\bibitem[\protect\citename{Coecke \bgroup et al.\egroup }2010]{Coecke:2010}
B.~Coecke, M.~Sadrzadeh, and S.~Clark.
\newblock 2010.
\newblock {Mathematical Foundations for a Compositional Distributional Model of
  Meaning}.
\newblock {\em Lambek Festschrift. Linguistic Analysis}, 36:345--384.

\bibitem[\protect\citename{Cohn and Lapata}2007]{Cohn:2007}
T.~Cohn and M.~Lapata.
\newblock 2007.
\newblock {Machine Translation by Triangulation: Making Effective Use of
  Multi-Parallel Corpora}.
\newblock In {\em Proceedings of the 45th Annual Meeting of the Association of
  Computational Linguistics}, pages 728--735, Prague, Czech Republic, June.
  Association for Computational Linguistics.

\bibitem[\protect\citename{Collins and Quillian}1969]{Collins:1969}
A.~Collins and M.~Quillian.
\newblock 1969.
\newblock {{Retrieval time from semantic memory}}.
\newblock {\em Journal of Verbal Learning and Verbal Behavior}, 8(2):240--247,
  April.

\bibitem[\protect\citename{Collins}2002]{Collins:2002}
M.~Collins.
\newblock 2002.
\newblock {Discriminative Training Methods for Hidden Markov Models: Theory and
  Experiments with Perceptron Algorithms}.
\newblock In {\em Proceedings of the ACL-02 Conference on Empirical Methods in
  Natural Language Processing - Volume 10}, EMNLP '02, pages 1--8,
  Philadelphia, PA. Association for Computational Linguistics.

\bibitem[\protect\citename{Collobert and Weston}2008]{Collobert:2008}
R.~Collobert and J.~Weston.
\newblock 2008.
\newblock {A Unified Architecture for Natural Language Processing: Deep Neural
  Networks with Multitask Learning}.
\newblock In {\em Proceedings of the 25th International Conference on Machine
  Learning}, ICML '08, pages 160--167, New York, NY, USA. ACM.

\bibitem[\protect\citename{Collobert \bgroup et al.\egroup
  }2011]{Collobert:2011}
R.~Collobert, J.~Weston, L.~Bottou, M.~Karlen, K.~Kavukcuoglu, and P.~Kuksa.
\newblock 2011.
\newblock {Natural Language Processing (Almost) from Scratch}.
\newblock {\em Journal of Machine Learning Research}, 12:2493--2537, November.

\bibitem[\protect\citename{Curran \bgroup et al.\egroup }2007]{Curran:2007}
J.~R. Curran, S.~Clark, and J.~Bos.
\newblock 2007.
\newblock {Linguistically Motivated Large-scale {NLP} with {C\&C} and {B}oxer}.
\newblock In {\em Proceedings of the 45th Annual Meeting of the ACL on
  Interactive Poster and Demonstration Sessions}, ACL '07, pages 33--36,
  Prague, Czech Republic. Association for Computational Linguistics.

\bibitem[\protect\citename{Curran}2004]{Curran:2004}
J.~R. Curran.
\newblock 2004.
\newblock {\em {{From distributional to semantic similarity}}}.
\newblock {Ph.D.} thesis.

\bibitem[\protect\citename{Curran}2005]{Curran:2005}
J.~R. Curran.
\newblock 2005.
\newblock {Supersense tagging of unknown nouns using semantic similarity}.
\newblock In {\em Proceedings of the 43rd Annual Meeting on Association for
  Computational Linguistics}, ACL '05, pages 26--33, Ann Arbor, Michigan.
  Association for Computational Linguistics.

\bibitem[\protect\citename{Das \bgroup et al.\egroup }2010]{Das:2010}
D.~Das, N.~Schneider, D.~Chen, and N.~A. Smith.
\newblock 2010.
\newblock {Probabilistic Frame-semantic Parsing}.
\newblock In {\em Human Language Technologies: The 2010 Annual Conference of
  the North American Chapter of the Association for Computational Linguistics},
  HLT '10, pages 948--956, Los Angeles, California. Association for
  Computational Linguistics.

\bibitem[\protect\citename{Das \bgroup et al.\egroup }2014]{Das:2014}
D.~Das, D.~Chen, A.~F.~T. Martins, N.~Schneider, and N.~A. Smith.
\newblock 2014.
\newblock {Frame-Semantic Parsing}.
\newblock {\em Computational Linguistics}, 40(1).

\bibitem[\protect\citename{de Marneffe and Manning}2013]{Marneffe:2013}
M.-C. de~Marneffe and C.~D. Manning, 2013.
\newblock {\em {Stanford typed dependencies manual}}.

\bibitem[\protect\citename{Duchi \bgroup et al.\egroup }2011]{Duchi:2011}
J.~Duchi, E.~Hazan, and Y.~Singer.
\newblock 2011.
\newblock {Adaptive Subgradient Methods for Online Learning and Stochastic
  Optimization}.
\newblock {\em Journal of Machine Learning Research}, 12:2121--2159, July.

\bibitem[\protect\citename{Duffy \bgroup et al.\egroup }1989]{Duffy:1989}
S.~A. Duffy, J.~M. Henderson, and R.~K. Morris.
\newblock 1989.
\newblock {Semantic facilitation of lexical access during sentence processing.}
\newblock {\em Journal of Experimental Psychology: Learning, Memory, and
  Cognition}, 15(5):791--801.

\bibitem[\protect\citename{Dumais \bgroup et al.\egroup }1988]{Dumais:1988}
S.~T. Dumais, G.~W. Furnas, T.~K. Landauer, S.~Deerwester, and R.~Harshman.
\newblock 1988.
\newblock {Using Latent Semantic Analysis to Improve Access to Textual
  Information}.
\newblock In {\em Proceedings of the SIGCHI Conference on Human Factors in
  Computing Systems}, CHI '88, pages 281--285, New York, NY, USA. ACM.

\bibitem[\protect\citename{Dyer \bgroup et al.\egroup }2010]{Dyer:2010}
C.~Dyer, J.~Weese, H.~Setiawan, A.~Lopez, F.~Ture, V.~Eidelman,
  J.~Ganitkevitch, P.~Blunsom, and P.~Resnik.
\newblock 2010.
\newblock {Cdec: A Decoder, Alignment, and Learning Framework for Finite-state
  and Context-free Translation Models}.
\newblock In {\em Proceedings of the ACL 2010 System Demonstrations}, ACLDemos
  '10, pages 7--12, Uppsala, Sweden. Association for Computational Linguistics.

\bibitem[\protect\citename{Erk and Pad\'{o}}2008]{Erk:2008}
K.~Erk and S.~Pad\'{o}.
\newblock 2008.
\newblock {A Structured Vector Space Model for Word Meaning in Context}.
\newblock pages 897--906.

\bibitem[\protect\citename{Fellbaum}1998]{Fellbaum:1998}
C.~Fellbaum, editor.
\newblock 1998.
\newblock {\em {{WordNet:} an electronic lexical database}}.
\newblock MIT Press.

\bibitem[\protect\citename{Fillmore \bgroup et al.\egroup }2003]{Fillmore:2003}
C.~J. Fillmore, C.~R. Johnson, and M.~R. Petruck.
\newblock 2003.
\newblock {Background to {FrameNet}}.
\newblock {\em International Journal of Lexicography}, 16(3).

\bibitem[\protect\citename{Fillmore}1982]{Fillmore:1982}
C.~J. Fillmore.
\newblock 1982.
\newblock {Frame {S}emantics}.
\newblock In {\em Linguistics in the Morning Calm}, pages 111--137. Hanshin
  Publishing Co., Seoul, South Korea.

\bibitem[\protect\citename{Firth}1957]{Firth:1957}
J.~R. Firth.
\newblock 1957.
\newblock {A synopsis of linguistic theory 1930-55.}
\newblock 1952-59:1--32.

\bibitem[\protect\citename{Foltz \bgroup et al.\egroup }1998]{Foltz:1998}
P.~W. Foltz, W.~Kintsch, and T.~K. Landauer.
\newblock 1998.
\newblock {The measurement of textual Coherence with Latent Semantic Analysis}.
\newblock {\em Discourse Processes}, 25:285--307.

\bibitem[\protect\citename{Frege}1892]{Frege:1892}
G.~Frege.
\newblock 1892.
\newblock {{\"{U}ber Sinn und Bedeutung}}.
\newblock In M.~Textor, editor, {\em Funktion - Begriff - Bedeutung}, volume~4
  of {\em Sammlung Philosophie}. Vandenhoeck \& Ruprecht, G\"{o}ttingen.

\bibitem[\protect\citename{Gildea and Jurafsky}2002]{Gildea:2002}
D.~Gildea and D.~Jurafsky.
\newblock 2002.
\newblock {Automatic Labeling of Semantic Roles}.
\newblock {\em Computational Linguistics}, 28(3):245--288.

\bibitem[\protect\citename{Goller and K\"{u}chler}1996]{Goller:1996}
C.~Goller and A.~K\"{u}chler.
\newblock 1996.
\newblock {Learning Task-Dependent Distributed Representations by
  Backpropagation Through Structure}.
\newblock In {\em Proceedings of the ICNN-96}, pages 347--352. IEEE.

\bibitem[\protect\citename{Grefenstette and Sadrzadeh}2011]{Grefenstette:2011}
E.~Grefenstette and M.~Sadrzadeh.
\newblock 2011.
\newblock {Experimental Support for a Categorical Compositional Distributional
  Model of Meaning}.
\newblock In {\em Proceedings of the Conference on Empirical Methods in Natural
  Language Processing}, EMNLP '11, pages 1394--1404, Edinburgh, United Kingdom.
  Association for Computational Linguistics.

\bibitem[\protect\citename{Grefenstette \bgroup et al.\egroup
  }2013]{Grefenstette:2013}
E.~Grefenstette, G.~Dinu, Y.-Z. Zhang, M.~Sadrzadeh, and M.~Baroni.
\newblock 2013.
\newblock {Multi-Step Regression Learning for Compositional Distributional
  Semantics}.
\newblock In {\em Proceedings of the 10th International Conference on
  Computational Semantics (IWCS 2013) -- Long Papers}, pages 131--142, Potsdam,
  Germany, March. Association for Computational Linguistics.

\bibitem[\protect\citename{Grefenstette}1994]{Grefenstette:1994}
G.~Grefenstette.
\newblock 1994.
\newblock {\em {Explorations in Automatic Thesaurus Discovery}}.
\newblock Kluwer Academic Publishers, Norwell, MA, USA.

\bibitem[\protect\citename{Grefenstette}2013a]{Grefenstette:2013b}
E.~Grefenstette.
\newblock 2013a.
\newblock {\em {Category-Theoretic Quantitative Compositional Distributional
  Models of Natural Language Semantics}}.
\newblock {Ph.D.} thesis.

\bibitem[\protect\citename{Grefenstette}2013b]{Grefenstette:2013a}
E.~Grefenstette.
\newblock 2013b.
\newblock {Towards a Formal Distributional Semantics: Simulating Logical
  Calculi with Tensors}.
\newblock In {\em Second Joint Conference on Lexical and Computational
  Semantics (*SEM), Volume 1: Proceedings of the Main Conference and the Shared
  Task: Semantic Textual Similarity}, pages 1--10, Atlanta, Georgia, USA, June.
  Association for Computational Linguistics.

\bibitem[\protect\citename{Griffiths \bgroup et al.\egroup
  }2007]{Griffiths:2007}
T.~L. Griffiths, J.~B. Tenenbaum, and M.~Steyvers.
\newblock 2007.
\newblock {Topics in semantic representation}.
\newblock {\em Psychological Review}, 114:2007.

\bibitem[\protect\citename{Guevara}2010]{Guevara:2010}
E.~Guevara.
\newblock 2010.
\newblock {A Regression Model of Adjective-Noun Compositionality in
  Distributional Semantics}.
\newblock In {\em Proceedings of the 2010 Workshop on GEometrical Models of
  Natural Language Semantics}, pages 33--37, Uppsala, Sweden, July. Association
  for Computational Linguistics.

\bibitem[\protect\citename{Guevara}2011]{Guevara:2011}
E.~Guevara.
\newblock 2011.
\newblock {Computing Semantic Compositionality in Distributional Semantics}.
\newblock In {\em Proceedings of the Ninth International Conference on
  Computational Semantics}, IWCS '11, pages 135--144, Oxford, United Kingdom.
  Association for Computational Linguistics.

\bibitem[\protect\citename{Haghighi \bgroup et al.\egroup }2008]{Haghighi:2008}
A.~Haghighi, P.~Liang, T.~Berg-Kirkpatrick, and D.~Klein.
\newblock 2008.
\newblock {Learning Bilingual Lexicons from Monolingual Corpora}.
\newblock In {\em Proceedings of ACL-08: HLT}, pages 771--779, Columbus, Ohio,
  June. Association for Computational Linguistics.

\bibitem[\protect\citename{Hermann and Blunsom}2013]{Hermann:2013:ACL}
K.~M. Hermann and P.~Blunsom.
\newblock 2013.
\newblock {The Role of Syntax in Vector Space Models of Compositional
  Semantics}.
\newblock In {\em Proceedings of the 51st Annual Meeting of the Association for
  Computational Linguistics (Volume 1: Long Papers)}, Sofia, Bulgaria, August.
  Association for Computational Linguistics.

\bibitem[\protect\citename{Hermann and Blunsom}2014a]{Hermann:2014:ICLR}
K.~M. Hermann and P.~Blunsom.
\newblock 2014a.
\newblock {Multilingual Distributed Representations without Word Alignment}.
\newblock In {\em Proceedings of the 2nd International Conference on Learning
  Representations}, Banff, Canada, April.

\bibitem[\protect\citename{Hermann and Blunsom}2014b]{Hermann:2014:ACLphil}
K.~M. Hermann and P.~Blunsom.
\newblock 2014b.
\newblock {Multilingual Models for Compositional Distributional Semantics}.
\newblock In {\em Proceedings of the 52nd Annual Meeting of the Association for
  Computational Linguistics (Volume 1: Long Papers)}, Baltimore, USA, June.
  Association for Computational Linguistics.

\bibitem[\protect\citename{Hermann \bgroup et al.\egroup
  }2012a]{Hermann:2012:Ranking}
K.~M. Hermann, P.~Blunsom, and S.~Pulman.
\newblock 2012a.
\newblock {An Unsupervised Ranking Model for Noun-Noun Compositionality}.
\newblock In {\em {*SEM 2012}: The First Joint Conference on Lexical and
  Computational Semantics -- Volume 1: Proceedings of the main conference and
  the shared task, and Volume 2: Proceedings of the Sixth International
  Workshop on Semantic Evaluation {(SemEval 2012)}}, pages 132--141,
  Montr\'{e}al, Canada, 7-8 June. Association for Computational Linguistics.

\bibitem[\protect\citename{Hermann \bgroup et al.\egroup
  }2012b]{Hermann:2012:SelPref}
K.~M. Hermann, C.~Dyer, P.~Blunsom, and S.~Pulman.
\newblock 2012b.
\newblock {Learning Semantics and Selectional Preference of Adjective-Noun
  Pairs}.
\newblock In {\em {*SEM 2012}: The First Joint Conference on Lexical and
  Computational Semantics -- Volume 1: Proceedings of the main conference and
  the shared task, and Volume 2: Proceedings of the Sixth International
  Workshop on Semantic Evaluation {(SemEval 2012)}}, pages 70--74,
  Montr\'{e}al, Canada, 7-8 June. Association for Computational Linguistics.

\bibitem[\protect\citename{Hermann \bgroup et al.\egroup
  }2013]{Hermann:2013:CVSC}
K.~M. Hermann, E.~Grefenstette, and P.~Blunsom.
\newblock 2013.
\newblock "not not bad" is not "bad": A distributional account of negation.
\newblock {\em Proceedings of the 2013 Workshop on Continuous Vector Space
  Models and their Compositionality}, August.

\bibitem[\protect\citename{Hermann \bgroup et al.\egroup
  }2014]{Hermann:2014:ACLgoogle}
K.~M. Hermann, D.~Das, J.~Weston, and K.~Ganchev.
\newblock 2014.
\newblock {Semantic Frame Identification with Distributed Word
  Representations}.
\newblock In {\em Proceedings of the 52nd Annual Meeting of the Association for
  Computational Linguistics (Volume 1: Long Papers)}, Baltimore, USA, June.
  Association for Computational Linguistics.

\bibitem[\protect\citename{Hinton and Salakhutdinov}2006]{Hinton:2006a}
G.~E. Hinton and R.~R. Salakhutdinov.
\newblock 2006.
\newblock {Reducing the Dimensionality of Data with Neural Networks}.
\newblock {\em Science}, 313(5786):504--507.

\bibitem[\protect\citename{Hinton \bgroup et al.\egroup }2006]{Hinton:2006}
G.~E. Hinton, S.~Osindero, M.~Welling, and Y.~W. Teh.
\newblock 2006.
\newblock {Unsupervised Discovery of Nonlinear Structure Using Contrastive
  Backpropagation}.
\newblock {\em Cognitive Science}, 30(4):725--731.

\bibitem[\protect\citename{Hockenmaier and Steedman}2007]{Hockenmaier:2007}
J.~Hockenmaier and M.~Steedman.
\newblock 2007.
\newblock {CCGbank: A Corpus of CCG Derivations and Dependency Structures
  Extracted from the Penn Treebank}.
\newblock {\em Computational Linguistics}, 33(3):355--396, September.

\bibitem[\protect\citename{Hovy \bgroup et al.\egroup }2006]{Hovy:2006}
E.~Hovy, M.~Marcus, M.~Palmer, L.~Ramshaw, and R.~Weischedel.
\newblock 2006.
\newblock {{O}nto{N}otes: The 90\% Solution}.
\newblock In {\em Proceedings of the Human Language Technology Conference of
  the NAACL, Companion Volume: Short Papers}, NAACL-Short '06, pages 57--60,
  New York, NY. Association for Computational Linguistics.

\bibitem[\protect\citename{Huang and Yates}2009]{Huang:2009}
F.~Huang and A.~Yates.
\newblock 2009.
\newblock {Distributional Representations for Handling Sparsity in Supervised
  Sequence-labeling}.
\newblock In {\em Proceedings of the Joint Conference of the 47th Annual
  Meeting of the ACL and the 4th International Joint Conference on Natural
  Language Processing of the AFNLP: Volume 1 - Volume 1}, ACL '09, pages
  495--503, Suntec, Singapore. Association for Computational Linguistics.

\bibitem[\protect\citename{Hubel and Wiesel}1968]{Hubel:1968}
D.~H. Hubel and T.~N. Wiesel.
\newblock 1968.
\newblock {Receptive Fields and Functional Architecture of Monkey Striate
  Cortex}.
\newblock {\em Journal of Physiology (London)}, 195:215--243.

\bibitem[\protect\citename{Jackendoff}1972]{Jackendoff:1972}
R.~Jackendoff.
\newblock 1972.
\newblock {\em {{Semantic Interpretation in Generative Grammar}}}.
\newblock MIT Press, Cambridge, MA.

\bibitem[\protect\citename{Johansson and Nugues}2007]{Johansson:2007}
R.~Johansson and P.~Nugues.
\newblock 2007.
\newblock {{LTH}: Semantic Structure Extraction using Nonprojective Dependency
  Trees}.
\newblock In {\em Proceedings of the Fourth International Workshop on Semantic
  Evaluations (SemEval-2007)}, pages 227--230, Prague, Czech Republic, June.
  Association for Computational Linguistics.

\bibitem[\protect\citename{Jones and Mewhort}2007]{Jones:2007}
M.~N. Jones and D.~J.~K. Mewhort.
\newblock 2007.
\newblock {Representing Word Meaning and Order Information in a Composite
  Holographic Lexicon}.
\newblock {\em psychological Review}, 114(1):1--37.

\bibitem[\protect\citename{Jones \bgroup et al.\egroup }2012]{ISI:2012}
B.~Jones, J.~Andreas, D.~Bauer, K.~M. Hermann, and K.~Knight.
\newblock 2012.
\newblock {Semantics-Based Machine Translation with Hyperedge Replacement
  Grammars}.
\newblock In {\em Proceedings of COLING 2012}, pages 1359--1376, Mumbai, India,
  December. The COLING 2012 Organizing Committee.
\newblock First four are joint first author in randomized order.

\bibitem[\protect\citename{Kalchbrenner and Blunsom}2013]{Kalchbrenner:2013}
N.~Kalchbrenner and P.~Blunsom.
\newblock 2013.
\newblock {Recurrent Convolutional Neural Networks for Discourse
  Compositionality}.
\newblock In {\em Proceedings of the Workshop on Continuous Vector Space Models
  and their Compositionality}, pages 119--126, Sofia, Bulgaria, August.
  Association for Computational Linguistics.

\bibitem[\protect\citename{Kalchbrenner \bgroup et al.\egroup
  }2014]{Kalchbrenner:2014}
N.~Kalchbrenner, E.~Grefenstette, and P.~Blunsom.
\newblock 2014.
\newblock {A Convolutional Neural Network for Modelling Sentences}.
\newblock In {\em Proceedings of the 52nd Annual Meeting of the Association for
  Computational Linguistics (Volume 1: Long Papers)}, Baltimore, USA, June.
  Association for Computational Linguistics.

\bibitem[\protect\citename{Kartsaklis \bgroup et al.\egroup
  }2012]{Kartsaklis:2012}
D.~Kartsaklis, M.~Sadrzadeh, and S.~Pulman.
\newblock 2012.
\newblock {A Unified Sentence Space for Categorical
  Distributional-Compositional Semantics: Theory and Experiments}.
\newblock In {\em Proceedings of 24th International Conference on Computational
  Linguistics (COLING 2012): Posters}, pages 549--558, Mumbai, India, December.

\bibitem[\protect\citename{Kintsch}2001]{Kintsch:2001}
W.~Kintsch.
\newblock 2001.
\newblock {Predication}.
\newblock {\em Cognitive Science}, 25:173--202.

\bibitem[\protect\citename{Klementiev \bgroup et al.\egroup
  }2012]{Klementiev:2012}
A.~Klementiev, I.~Titov, and B.~Bhattarai.
\newblock 2012.
\newblock {Inducing Crosslingual Distributed Representations of Words}.
\newblock In {\em Proceedings of COLING 2012}, pages 1459--1474, Mumbai, India,
  December. The COLING 2012 Organizing Committee.

\bibitem[\protect\citename{Koehn}2005]{Koehn:2005}
P.~Koehn.
\newblock 2005.
\newblock {{Europarl: A Parallel Corpus for Statistical Machine Translation}}.
\newblock In {\em {Conference Proceedings: the tenth Machine Translation
  Summit}}, pages 79--86, Phuket, Thailand. AAMT, AAMT.

\bibitem[\protect\citename{Kracht}2008]{Kracht:2008}
M.~Kracht.
\newblock 2008.
\newblock {{Compositionality in Montague Grammar}}.
\newblock In W.~Hinzen, E.~Machery, and M.~Werning, editors, {\em {Handbook of
  Compositionality}}, pages 47 -- 63. Oxford University Press.

\bibitem[\protect\citename{Lafferty \bgroup et al.\egroup }2001]{Lafferty:2001}
J.~D. Lafferty, A.~McCallum, and F.~C.~N. Pereira.
\newblock 2001.
\newblock {Conditional Random Fields: Probabilistic Models for Segmenting and
  Labeling Sequence Data}.
\newblock In {\em Proceedings of the Eighteenth International Conference on
  Machine Learning}, ICML '01, pages 282--289, San Francisco, CA, USA. Morgan
  Kaufmann Publishers Inc.

\bibitem[\protect\citename{Landauer and Dumais}1997]{Landauer:1997}
T.~K. Landauer and S.~T. Dumais.
\newblock 1997.
\newblock {A solution to {Plato's} problem: The latent semantic analysis theory
  of acquisition, induction, and representation of knowledge.}
\newblock {\em Psychological review}, 104(2):211--240.

\bibitem[\protect\citename{Lauly \bgroup et al.\egroup }2013]{Lauly:2013}
S.~Lauly, A.~Boulanger, and H.~Larochelle.
\newblock 2013.
\newblock Learning multilingual word representations using a bag-of-words
  autoencoder.
\newblock In {\em Deep Learning Workshop at NIPS}.

\bibitem[\protect\citename{LeCun \bgroup et al.\egroup }1998a]{LeCun:1998b}
Y.~LeCun, L.~Bottou, Y.~Bengio, and P.~Haffner.
\newblock 1998a.
\newblock {Gradient-Based Learning Applied to Document Recognition}.
\newblock {\em Proceedings of the IEEE}, 86(11):2278--2324, November.

\bibitem[\protect\citename{LeCun \bgroup et al.\egroup }1998b]{LeCun:1998}
Y.~LeCun, L.~Bottou, G.~Orr, and K.-R. Muller.
\newblock 1998b.
\newblock {Efficient BackProp}.
\newblock In G.~Orr and M.~K., editors, {\em Neural Networks: Tricks of the
  trade}. Springer.

\bibitem[\protect\citename{Lewis \bgroup et al.\egroup }2004]{Lewis:2004}
D.~D. Lewis, Y.~Yang, T.~G. Rose, and F.~Li.
\newblock 2004.
\newblock {{RCV1}: A New Benchmark Collection for Text Categorization
  Research}.
\newblock {\em Journal of Machine Learning Research}, 5:361--397, December.

\bibitem[\protect\citename{Li and Fukushima}2000]{Li:2000}
D.-H. Li and M.~Fukushima.
\newblock 2000.
\newblock {On the Global Convergence of the {BFGS} Method for Nonconvex
  Unconstrained Optimization Problems}.
\newblock {\em SIAM J. on Optimization}, 11(4):1054--1064, April.

\bibitem[\protect\citename{Lin}1999]{Lin:1999}
D.~Lin.
\newblock 1999.
\newblock {Automatic Identification of Non-compositional Phrases}.
\newblock In {\em Proceedings of the 37th Annual Meeting of the Association for
  Computational Linguistics on Computational Linguistics}, ACL '99, pages
  317--324, College Park, Maryland. Association for Computational Linguistics.

\bibitem[\protect\citename{Liu and Nocedal}1989]{Liu:1989}
D.~C. Liu and J.~Nocedal.
\newblock 1989.
\newblock {On the limited memory {BFGS} method for large scale optimization}.
\newblock {\em Mathematical Programming}, 45(3).

\bibitem[\protect\citename{Maillard \bgroup et al.\egroup }2014]{Maillard:2014}
J.~Maillard, S.~Clark, and E.~Grefenstette.
\newblock 2014.
\newblock {A Type-Driven Tensor-Based Semantics for {CCG}}.
\newblock In {\em EACL 2014 Type Theory and Natural Language Semantics
  Workshop}.

\bibitem[\protect\citename{M\`{a}rquez \bgroup et al.\egroup
  }2008]{Marquez:2008}
L.~M\`{a}rquez, X.~Carreras, K.~C. Litkowski, and S.~Stevenson.
\newblock 2008.
\newblock {Semantic role labeling: an introduction to the special issue}.
\newblock {\em Computational Linguistics}, 34(2).

\bibitem[\protect\citename{Masterman}2005]{Masterman:2005}
M.~Masterman.
\newblock 2005.
\newblock {Language, Cohesion and Form}.
\newblock In Y.~Wilks, editor, {\em {Studies in Natural Language Processing}}.
  Cambridge University Press, Cambridge, UK.

\bibitem[\protect\citename{Matsubayashi \bgroup et al.\egroup
  }2009]{Matsubayashi:2009}
Y.~Matsubayashi, N.~Okazaki, and J.~Tsujii.
\newblock 2009.
\newblock {A Comparative Study on Generalization of Semantic Roles in
  {FrameNet}}.
\newblock In {\em Proceedings of the Joint Conference of the 47th Annual
  Meeting of the ACL and the 4th International Joint Conference on Natural
  Language Processing of the AFNLP}, pages 19--27, Suntec, Singapore, August.
  Association for Computational Linguistics.

\bibitem[\protect\citename{McCallum}2002]{McCallum:2002}
A.~K. McCallum.
\newblock 2002.
\newblock {MALLET: A Machine Learning for Language Toolkit}.
\newblock http://mallet.cs.umass.edu.

\bibitem[\protect\citename{McDonald}2000]{McDonald:2000}
S.~A. McDonald.
\newblock 2000.
\newblock {{Environmental determinants of lexical processing effort}}.

\bibitem[\protect\citename{McRae \bgroup et al.\egroup }1997]{McRae:1997}
K.~McRae, V.~R. de~Sa, and M.~S. Seidenberg.
\newblock 1997.
\newblock {On the nature and scope of featural representations of word
  meaning.}
\newblock {\em Journal of Experimental Psychology: General}, 126(2):99--130,
  06.

\bibitem[\protect\citename{Meyers \bgroup et al.\egroup }2004]{Meyers:2004}
A.~Meyers, R.~Reeves, C.~Macleod, R.~Szekely, V.~Zielinska, B.~Young, and
  R.~Grishman.
\newblock 2004.
\newblock {The {NomBank} Project: An Interim Report}.
\newblock In A.~Meyers, editor, {\em HLT-NAACL 2004 Workshop: Frontiers in
  Corpus Annotation}, pages 24--31, Boston, Massachusetts, USA, May 2 - May 7.
  Association for Computational Linguistics.

\bibitem[\protect\citename{Mikolov \bgroup et al.\egroup }2010]{Mikolov:2010}
T.~Mikolov, M.~Karafi\'{a}t, L.~Burget, J.~\v{C}ernock\'{y}, and S.~Khudanpur.
\newblock 2010.
\newblock {Recurrent neural network based language model}.
\newblock In {\em Proceedings of the 11th Annual Conference of the
  International Speech Communication Association (INTERSPEECH 2010)}, volume
  2010, pages 1045--1048. International Speech Communication Association.

\bibitem[\protect\citename{Mikolov \bgroup et al.\egroup }2013a]{Mikolov:2013}
T.~Mikolov, K.~Chen, G.~Corrado, and J.~Dean.
\newblock 2013a.
\newblock {Efficient Estimation of Word Representations in Vector Space}.
\newblock In {\em Proceedings of the Workshop at the 1st International
  Conference on Learning Representations}, Scottsdale, Arizona, USA, May.

\bibitem[\protect\citename{Mikolov \bgroup et al.\egroup }2013b]{Mikolov:2013a}
T.~Mikolov, Q.~V. Le, and I.~Sutskever.
\newblock 2013b.
\newblock {Exploiting Similarities among Languages for Machine Translation}.
\newblock {\em CoRR}.

\bibitem[\protect\citename{Mitchell and Lapata}2008]{Mitchell:2008}
J.~Mitchell and M.~Lapata.
\newblock 2008.
\newblock {Vector-based Models of Semantic Composition}.
\newblock In {\em Proceedings of ACL-08: HLT}, pages 236--244, Columbus, Ohio,
  June. Association for Computational Linguistics.

\bibitem[\protect\citename{Mitchell and Lapata}2009]{Mitchell:2009}
J.~Mitchell and M.~Lapata.
\newblock 2009.
\newblock {{Language models based on semantic composition}}.
\newblock In {\em Proceedings of the 2009 Conference on Empirical Methods in
  Natural Language Processing: Volume 1}, pages 430--439. Association for
  Computational Linguistics.

\bibitem[\protect\citename{Mitchell and Lapata}2010]{Mitchell:2010}
J.~Mitchell and M.~Lapata.
\newblock 2010.
\newblock {Composition in Distributional Models of Semantics}.
\newblock {\em Cognitive Science}, 34(8):1388--1429.

\bibitem[\protect\citename{Mitchell}2011]{Mitchell:2011}
J.~Mitchell.
\newblock 2011.
\newblock {\em {Composition in distributional models of semantics}}.
\newblock {Ph.D.} thesis.

\bibitem[\protect\citename{Mnih and Hinton}2009]{Mnih:2009}
A.~Mnih and G.~Hinton.
\newblock 2009.
\newblock A scalable hierarchical distributed language model.
\newblock In {\em Advances in Neural Information Processing Systems},
  volume~21, pages 1081--1088.

\bibitem[\protect\citename{Montague}1970]{Montague:1970}
R.~Montague.
\newblock 1970.
\newblock {Universal Grammar}.
\newblock {\em Theoria}, 36(3):373--398.

\bibitem[\protect\citename{Montague}1974]{Montague:1974}
R.~Montague.
\newblock 1974.
\newblock {{English as a Formal Language}}.
\newblock {\em Formal Semantics: The Essential Readings}.

\bibitem[\protect\citename{Morris and Harris}2002]{Morris:2002}
A.~L. Morris and C.~L. Harris.
\newblock 2002.
\newblock {Sentence context, word recognition, and repetition blindness}.
\newblock {\em Journal of Experimental Psychology: Learning, Memory, and
  Cognition}, pages 962--982.

\bibitem[\protect\citename{Morris}1994]{Morris:1994}
R.~K. Morris.
\newblock 1994.
\newblock {Lexical and message-level sentence context effects on fixation times
  in reading.}
\newblock {\em Journal of Experimental Psychology: Learning, Memory, and
  Cognition}, 20(1):92--103.

\bibitem[\protect\citename{Nakagawa \bgroup et al.\egroup }2010]{Nakagawa:2010}
T.~Nakagawa, K.~Inui, and S.~Kurohashi.
\newblock 2010.
\newblock {Dependency Tree-based Sentiment Classification Using CRFs with
  Hidden Variables}.
\newblock In {\em Human Language Technologies: The 2010 Annual Conference of
  the North American Chapter of the Association for Computational Linguistics},
  HLT '10, pages 786--794, Los Angeles, California. Association for
  Computational Linguistics.

\bibitem[\protect\citename{Ngiam \bgroup et al.\egroup }2011]{Ngiam:2011}
J.~Ngiam, A.~Khosla, M.~Kim, J.~Nam, H.~Lee, and A.~Y. Ng.
\newblock 2011.
\newblock Multimodal deep learning.
\newblock In {\em Proceedings of the 29th International Conference on Machine
  Learning}, ICML '11.

\bibitem[\protect\citename{Nocedal and Wright}2006]{Nocedal:2006}
J.~Nocedal and S.~J. Wright.
\newblock 2006.
\newblock {\em {Numerical Optimization}}.
\newblock Springer, New York, 2nd edition.

\bibitem[\protect\citename{Pad\'{o} and Lapata}2007]{Pado:2007}
S.~Pad\'{o} and M.~Lapata.
\newblock 2007.
\newblock {Dependency-Based Construction of Semantic Space Models}.
\newblock {\em Computational Linguistics}, 33(2):161--199, June.

\bibitem[\protect\citename{Palmer \bgroup et al.\egroup }2005]{Palmer:2005}
M.~Palmer, D.~Gildea, and P.~Kingsbury.
\newblock 2005.
\newblock {The {Proposition} Bank: An Annotated Corpus of Semantic Roles}.
\newblock {\em Computational Linguistics}, 31(1):71--106.

\bibitem[\protect\citename{Pang and Lee}2005]{Pang:2005}
B.~Pang and L.~Lee.
\newblock 2005.
\newblock {Seeing Stars: Exploiting Class Relationships for Sentiment
  Categorization with Respect to Rating Scales}.
\newblock In {\em Proceedings of the 43rd Annual Meeting on Association for
  Computational Linguistics}, ACL '05, pages 115--124, Ann Arbor, Michigan.
  Association for Computational Linguistics.

\bibitem[\protect\citename{Peirce}1931]{Peirce:1931}
C.~S. Peirce.
\newblock 1931.
\newblock {\em {{Collected Papers of Charles Sanders Peirce}}}.
\newblock Harvard University Press.

\bibitem[\protect\citename{Pelletier}1994]{Pelletier:1994}
F.~J. Pelletier.
\newblock 1994.
\newblock {The Principle of Semantic Compositionality}.
\newblock {\em Topoi}, 13:11--24.

\bibitem[\protect\citename{Pereira \bgroup et al.\egroup }1993]{Pereira:1993}
F.~Pereira, N.~Tishby, and L.~Lee.
\newblock 1993.
\newblock {Distributional Clustering of {E}nglish Words}.
\newblock In {\em Proceedings of the 31st Annual Meeting on Association for
  Computational Linguistics}, ACL '93, pages 183--190, Columbus, Ohio.
  Association for Computational Linguistics.

\bibitem[\protect\citename{Pollack}1990]{Pollack:1990}
J.~B. Pollack.
\newblock 1990.
\newblock {Recursive Distributed Representations}.
\newblock {\em Artificial Intelligence}, 46:77--105.

\bibitem[\protect\citename{Punyakanok \bgroup et al.\egroup
  }2008]{Punyakanok:2008}
V.~Punyakanok, D.~Roth, and W.-T. Yih.
\newblock 2008.
\newblock {The Importance of Syntactic Parsing and Inference in Semantic Role
  Labeling}.
\newblock {\em Computational Linguistics}, 34(2).

\bibitem[\protect\citename{Richens}1956]{Richens:1956}
R.~H. Richens.
\newblock 1956.
\newblock {Preprogramming for machine translation}.
\newblock {\em Mechanical Translation}, 3(1):20--25.

\bibitem[\protect\citename{Richens}1958]{Richens:1958}
R.~H. Richens.
\newblock 1958.
\newblock {Interlingual machine translation}.
\newblock {\em The Computer Journal}, 1(3):144--147.

\bibitem[\protect\citename{Riedel \bgroup et al.\egroup }2013]{Riedel:2013}
S.~Riedel, L.~Yao, A.~McCallum, and B.~M. Marlin.
\newblock 2013.
\newblock {Relation Extraction with Matrix Factorization and Universal
  Schemas}.
\newblock In {\em Proceedings of the 2013 Conference of the North American
  Chapter of the Association for Computational Linguistics: Human Language
  Technologies}, pages 74--84, Atlanta, Georgia, June. Association for
  Computational Linguistics.

\bibitem[\protect\citename{Roy}2003]{Roy:2003}
D.~Roy.
\newblock 2003.
\newblock {Grounded Spoken Language Acquisition: Experiments in Word Learning}.
\newblock {\em IEEE Transactions on Multimedia}, 5(2):197--209, June.

\bibitem[\protect\citename{Sarath~Chandar \bgroup et al.\egroup
  }2013]{SarathChandar:2013}
A.~P. Sarath~Chandar, M.~K. Mitesh, B.~Ravindran, V.~Raykar, and A.~Saha.
\newblock 2013.
\newblock {Multilingual Deep Learning}.
\newblock In {\em Deep Learning Workshop at NIPS}.

\bibitem[\protect\citename{Scheible and Sch{\"u}tze}2013]{Scheible:2013}
C.~Scheible and H.~Sch{\"u}tze.
\newblock 2013.
\newblock {Cutting Recursive Autoencoder Trees}.
\newblock In {\em Proceedings of the 1st International Conference on Learning
  Representations}, Scottsdale, Arizona, USA, May.

\bibitem[\protect\citename{Sch{\"u}tze}1998]{Schutze:1998}
H.~Sch{\"u}tze.
\newblock 1998.
\newblock {Automatic word sense discrimination}.
\newblock {\em Computational Linguistics}, 24(1):97--123.

\bibitem[\protect\citename{Smolensky and Legendre}2006]{Smolensky:2006}
P.~Smolensky and G.~Legendre.
\newblock 2006.
\newblock {\em {{The Harmonic Mind: From Neural Computation to
  Optimality-Theoretic Grammar Volume I: Cognitive Architecture}}}.

\bibitem[\protect\citename{Smolensky}1990]{Smolensky:1990}
P.~Smolensky.
\newblock 1990.
\newblock {{Tensor product variable binding and the representation of symbolic
  structures in connectionist systems}}.
\newblock {\em Artificial intelligence}, 46(1-2):159--216.

\bibitem[\protect\citename{Socher \bgroup et al.\egroup }2011a]{Socher:2011a}
R.~Socher, E.~H. Huang, J.~Pennin, C.~D. Manning, and A.~Y. Ng.
\newblock 2011a.
\newblock Dynamic pooling and unfolding recursive autoencoders for paraphrase
  detection.
\newblock In J.~Shawe-Taylor, R.~Zemel, P.~Bartlett, F.~Pereira, and
  K.~Weinberger, editors, {\em Advances in Neural Information Processing
  Systems 24}, pages 801--809. Curran Associates, Inc.

\bibitem[\protect\citename{Socher \bgroup et al.\egroup }2011b]{Socher:2011}
R.~Socher, J.~Pennington, E.~H. Huang, A.~Y. Ng, and C.~D. Manning.
\newblock 2011b.
\newblock {Semi-supervised Recursive Autoencoders for Predicting Sentiment
  Distributions}.
\newblock In {\em Proceedings of the Conference on Empirical Methods in Natural
  Language Processing}, EMNLP '11, pages 151--161, Edinburgh, United Kingdom.
  Association for Computational Linguistics.

\bibitem[\protect\citename{Socher \bgroup et al.\egroup }2012a]{Socher:2012}
R.~Socher, B.~Huval, B.~Bath, C.~D. Manning, and A.~Y. Ng.
\newblock 2012a.
\newblock Convolutional-recursive deep learning for 3d object classification.
\newblock In F.~Pereira, C.~Burges, L.~Bottou, and K.~Weinberger, editors, {\em
  Advances in Neural Information Processing Systems 25}, pages 656--664. Curran
  Associates, Inc.

\bibitem[\protect\citename{Socher \bgroup et al.\egroup }2012b]{Socher:2012a}
R.~Socher, B.~Huval, C.~D. Manning, and A.~Y. Ng.
\newblock 2012b.
\newblock {Semantic Compositionality Through Recursive Matrix-vector Spaces}.
\newblock In {\em Proceedings of the 2012 Joint Conference on Empirical Methods
  in Natural Language Processing and Computational Natural Language Learning},
  EMNLP-CoNLL '12, pages 1201--1211, Jeju Island, Korea. Association for
  Computational Linguistics.

\bibitem[\protect\citename{Socher \bgroup et al.\egroup }2013]{Socher:2013}
R.~Socher, A.~Perelygin, J.~Wu, J.~Chuang, C.~D. Manning, A.~Y. Ng, and
  C.~Potts.
\newblock 2013.
\newblock {Recursive Deep Models for Semantic Compositionality Over a Sentiment
  Treebank}.
\newblock In {\em Proceedings of the 2013 Conference on Empirical Methods in
  Natural Language Processing}, EMNLP '13, pages 1631--1642, Seattle, WA,
  October. Association for Computational Linguistics.

\bibitem[\protect\citename{Sp\"{a}rck~Jones}1988]{SparckJones:1988}
K.~Sp\"{a}rck~Jones.
\newblock 1988.
\newblock {A Statistical Interpretation of Term Specificity and Its Application
  in Retrieval}.
\newblock In P.~Willett, editor, {\em Document Retrieval Systems}, pages
  132--142. Taylor Graham Publishing, London, UK, UK.

\bibitem[\protect\citename{Srivastava and Salakhutdinov}2012]{Srivastava:2012}
N.~Srivastava and R.~Salakhutdinov.
\newblock 2012.
\newblock Multimodal learning with deep boltzmann machines.
\newblock In F.~Pereira, C.~Burges, L.~Bottou, and K.~Weinberger, editors, {\em
  Advances in Neural Information Processing Systems 25}, pages 2222--2230.
  Curran Associates, Inc.

\bibitem[\protect\citename{Steedman and Baldridge}2011]{Steedman:2011}
M.~Steedman and J.~Baldridge.
\newblock 2011.
\newblock {Combinatory Categorial Grammar}.
\newblock In {\em Non-Transformational Syntax}, pages 181--224.
  Wiley-Blackwell.

\bibitem[\protect\citename{Steyvers and Griffiths}2005]{Steyvers:2005}
M.~Steyvers and T.~Griffiths.
\newblock 2005.
\newblock {Probabilistic topic models}.
\newblock In T.~Landauer, D.~Mcnamara, S.~Dennis, and W.~Kintsch, editors, {\em
  Latent Semantic Analysis: A Road to Meaning}. Laurence Erlbaum.

\bibitem[\protect\citename{Szabolcsi}1989]{Szabolcsi:1989}
A.~Szabolcsi.
\newblock 1989.
\newblock {{Bound Variables in Syntax: Are There Any?}}
\newblock In R.~Bartsch, J.~van Benthem, and P.~van Emde~Boas, editors, {\em
  Semantics and Contextual Expression}, pages 295--318. Foris, Dordrecht.

\bibitem[\protect\citename{Turian \bgroup et al.\egroup }2010]{Turian:2010}
J.~Turian, L.-A. Ratinov, and Y.~Bengio.
\newblock 2010.
\newblock {Word Representations: A Simple and General Method for
  Semi-Supervised Learning}.
\newblock In {\em Proceedings of the 48th Annual Meeting of the Association for
  Computational Linguistics}, pages 384--394, Uppsala, Sweden, July.
  Association for Computational Linguistics.

\bibitem[\protect\citename{Turney}2012]{Turney:2012}
P.~D. Turney.
\newblock 2012.
\newblock {Domain and Function: A Dual-Space Model of Semantic Relations and
  Compositions}.
\newblock {\em Journal of Artificial Intelligence Research}, 44:533--585.

\bibitem[\protect\citename{Usunier \bgroup et al.\egroup }2009]{Usunier:2009}
N.~Usunier, D.~Buffoni, and P.~Gallinari.
\newblock 2009.
\newblock {Ranking with Ordered Weighted Pairwise Classification}.
\newblock In {\em Proceedings of the 26th Annual International Conference on
  Machine Learning}, ICML '09, pages 1057--1064, New York, NY, USA. ACM.

\bibitem[\protect\citename{Uszkoreit and Brants}2008]{Uszkoreit:2008}
J.~Uszkoreit and T.~Brants.
\newblock 2008.
\newblock {Distributed Word Clustering for Large Scale Class-Based Language
  Modeling in Machine Translation}.
\newblock In {\em Proceedings of ACL-08: HLT}, pages 755--762, Columbus, Ohio,
  June. Association for Computational Linguistics.

\bibitem[\protect\citename{Vincent \bgroup et al.\egroup }2008]{Vincent:2008}
P.~Vincent, H.~Larochelle, Y.~Bengio, and P.-A. Manzagol.
\newblock 2008.
\newblock {Extracting and Composing Robust Features with Denoising
  Autoencoders}.
\newblock In {\em Proceedings of the 25th International Conference on Machine
  Learning}, ICML '08, pages 1096--1103, New York, NY, USA. ACM.

\bibitem[\protect\citename{Wager \bgroup et al.\egroup }2013]{Wager:2013}
S.~Wager, S.~Wang, and P.~Liang.
\newblock 2013.
\newblock {Dropout Training as Adaptive Regularization}.
\newblock In C.~J.~C. Burges, L.~Bottou, Z.~Ghahramani, and K.~Q. Weinberger,
  editors, {\em Advances in Neural Information Processing Systems 26}, pages
  351--359.

\bibitem[\protect\citename{Wang and Manning}2012]{Wang:2012}
S.~Wang and C.~D. Manning.
\newblock 2012.
\newblock {Baselines and Bigrams: Simple, Good Sentiment and Topic
  Classification}.
\newblock In {\em Proceedings of the 50th Annual Meeting of the Association for
  Computational Linguistics: Short Papers - Volume 2}, ACL '12, pages 90--94,
  Jeju Island, Korea. Association for Computational Linguistics.

\bibitem[\protect\citename{Weinberger and Saul}2009]{Weinberger:2009}
K.~Q. Weinberger and L.~K. Saul.
\newblock 2009.
\newblock {Distance Metric Learning for Large Margin Nearest Neighbor
  Classification}.
\newblock {\em Journal of Machine Learning Research}, 10:207--244.

\bibitem[\protect\citename{Weston \bgroup et al.\egroup }2011]{Weston:2011}
J.~Weston, S.~Bengio, and N.~Usunier.
\newblock 2011.
\newblock {{WSABIE}: Scaling Up to Large Vocabulary Image Annotation}.
\newblock In {\em Proceedings of the Twenty-Second International Joint
  Conference on Artificial Intelligence - Volume Three}, IJCAI'11, pages
  2764--2770. AAAI Press.

\bibitem[\protect\citename{Widdows}2004]{Widdows:2004}
D.~Widdows.
\newblock 2004.
\newblock {\em {{Geometry and Meaning}}}.
\newblock CSLI Publications, Stanford, California, USA, November.

\bibitem[\protect\citename{Widdows}2008]{Widdows:2008}
D.~Widdows.
\newblock 2008.
\newblock {Semantic Vector Products: Some Initial Investigations}.
\newblock In {\em Proceedings of the Second AAAI Symposium on Quantum
  Interaction}. College Publications.

\bibitem[\protect\citename{Wiebe \bgroup et al.\egroup }2005]{Wiebe:2005}
J.~Wiebe, T.~Wilson, and C.~Cardie.
\newblock 2005.
\newblock {Annotating Expressions of Opinions and Emotions in Language}.
\newblock {\em Language Resources and Evaluation}, 39(2-3):165--210.

\bibitem[\protect\citename{Wilson \bgroup et al.\egroup }2005]{Wilson:2005}
T.~Wilson, J.~Wiebe, and P.~Hoffmann.
\newblock 2005.
\newblock {Recognizing Contextual Polarity in Phrase-level Sentiment Analysis}.
\newblock In {\em Proceedings of the Conference on Human Language Technology
  and Empirical Methods in Natural Language Processing}, HLT '05, pages
  347--354, Vancouver, British Columbia, Canada. Association for Computational
  Linguistics.

\bibitem[\protect\citename{Xue and Palmer}2004]{Xue:2004}
N.~Xue and M.~Palmer.
\newblock 2004.
\newblock {Calibrating Features for Semantic Role Labeling }.
\newblock In D.~Lin and D.~Wu, editors, {\em Proceedings of the Conference on
  Empirical Methods in Natural Language Processing}, EMNLP '04, pages 88--94,
  Barcelona, Spain, July. Association for Computational Linguistics.

\bibitem[\protect\citename{Yih \bgroup et al.\egroup }2011]{Yih:2011}
W.-T. Yih, K.~Toutanova, J.~C. Platt, and C.~Meek.
\newblock 2011.
\newblock {Learning Discriminative Projections for Text Similarity Measures}.
\newblock In {\em Proceedings of the Fifteenth Conference on Computational
  Natural Language Learning}, CoNLL '11, pages 247--256, Portland, Oregon.
  Association for Computational Linguistics.

\bibitem[\protect\citename{Zanzotto \bgroup et al.\egroup }2010]{Zanzotto:2010}
F.~M. Zanzotto, I.~Korkontzelos, F.~Fallucchi, and S.~Manandhar.
\newblock 2010.
\newblock {Estimating Linear Models for Compositional Distributional
  Semantics}.
\newblock In {\em Proceedings of the 23rd International Conference on
  Computational Linguistics}, COLING '10, pages 1263--1271, Beijing, China.
  Association for Computational Linguistics.

\bibitem[\protect\citename{Zhang and Nivre}2011]{Zhang:2011}
Y.~Zhang and J.~Nivre.
\newblock 2011.
\newblock {Transition-based Dependency Parsing with Rich Non-local Features}.
\newblock In {\em Proceedings of the 49th Annual Meeting of the Association for
  Computational Linguistics: Human Language Technologies: Short Papers - Volume
  2}, HLT '11, pages 188--193, Portland, Oregon. Association for Computational
  Linguistics.

\bibitem[\protect\citename{Zou \bgroup et al.\egroup }2013]{Zou:2013}
W.~Y. Zou, R.~Socher, D.~Cer, and C.~D. Manning.
\newblock 2013.
\newblock {Bilingual Word Embeddings for Phrase-Based Machine Translation}.
\newblock In {\em Proceedings of the 2013 Conference on Empirical Methods in
  Natural Language Processing}, pages 1393--1398, Seattle, Washington, USA,
  October. Association for Computational Linguistics.

\end{thebibliography}
\end{document}